\documentclass[twoside,11pt]{article}
\usepackage[preprint]{jmlr2e}

\usepackage{style}
\usepackage{graphicx}
\usepackage{enumitem}

\usepackage[dvipsnames]{xcolor}

\usepackage[title]{appendix}

\usepackage{algorithm}
\usepackage{algpseudocode}

\usepackage{tikz}
\usepackage{tkz-euclide}
\usetikzlibrary{positioning}
\usetikzlibrary{arrows,calc}
\usetikzlibrary{backgrounds}
\usetikzlibrary{shapes.misc}
\usetikzlibrary{arrows.meta}
\tikzset{
>=stealth',
help lines/.style={dashed, thick},
axis/.style={<->},
important line/.style={thick},
connection/.style={thick, dotted},
}

\definecolor{light_blue}{RGB}{137, 171, 227}
\definecolor{light_pink}{RGB}{234, 115, 141}
\definecolor{my_green}{RGB}{31, 191, 184}
\definecolor{my_blue}{RGB}{25, 120, 165}

\definecolor{test_1}{RGB}{241, 241, 242} 
\definecolor{test_2}{RGB}{25, 149, 173} 
\definecolor{test_3}{RGB}{161, 214, 226} 
\definecolor{test_4}{RGB}{188, 186, 190} 

\title{Learning-to-Optimize with PAC-Bayesian Guarantees: Theoretical Considerations and Practical Implementation}

\author{\name Michael Sucker \email michael.sucker@math.uni-tuebingen.de \\
        \addr Department of Mathematics \\
        University of T\"ubingen\\
        T\"ubingen, Germany
        \AND
        \name Jalal Fadili \email jalal.fadili@ensicaen.fr \\
        \addr ENSICAEN \\ Normandie Université \\ CNRS, GREYC, France
        \AND
        \name Peter Ochs \email ochs@cs.uni-saarland.de \\
        \addr Department of Mathematics and Computer Science\\
        Saarland University\\
        Saarbr\"ucken, Germany}

\editor{...}

\usepackage{lastpage}
\jmlrheading{XX}{XXXX}{1-\pageref{LastPage}}{X/XX; Revised X/XX}{X/XX}{XX-XXXX}{Michael Sucker, Jalal Fadili and Peter Ochs}

\ShortHeadings{Sucker, Fadili and Ochs}{PAC-Bayesian Learning-to-Optimize}
\firstpageno{1}

\begin{document}

\maketitle

\begin{abstract}%
    We use the PAC-Bayesian theory for the setting of learning-to-optimize. To the best of our knowledge, we present the first framework to learn optimization algorithms with provable generalization guarantees (PAC-Bayesian bounds) and explicit trade-off between convergence guarantees and convergence speed, which contrasts with the typical worst-case analysis. Our learned optimization algorithms provably outperform related ones derived from a worst-case analysis. The results rely on PAC-Bayesian bounds for general, possibly unbounded loss-functions based on exponential families. Further, we provide a concrete algorithmic realization of the framework and new methodologies for learning-to-optimize. Finally, we conduct four practically relevant experiments to support our theory. With this, we showcase that the provided learning framework yields optimization algorithms that provably outperform the state-of-the-art by orders of magnitude.
\end{abstract}


\section{Introduction}

Typically, optimization algorithms are derived by performing a worst-case analysis on a specific class of problems. Doing so one obtains theoretical convergence guarantees for any instance inside the class. However, the abstract class of problems contains an enormous number beyond the concrete problem of interest, often including also pathological functions that do not occur in practical applications. Furthermore, since the derivation is done \say{on pen and paper}, all modeling steps have to be analytically tractable. This limits the design of algorithms. Both of these restrictions can impair the performance of the resulting method on a concrete problem instance. Nevertheless, both restrictions can be alleviated through learning: Given a concrete application and performance-measure, the algorithm is trained on examples (data) to improve its performance. This enables the automatic adaptation of sophisticated algorithms to this particular setting. However, there is no free lunch: If the algorithm is explicitly based on quantities that are not analytically tractable, one cannot expect to obtain the same theoretical guarantees as for the worst-case analysis discussed above. Since the practical usefulness of an optimization algorithm without convergence guarantees is at least questionable, this is a major problem and poses the first central question: 
\begin{center}
    \emph{What kind of theoretical guarantees can be given for a learned optimization algorithm? Are we able to ensure its usefulness?}
\end{center}
One possible alternative to the common guarantees is of statistical nature: Even if we do not know exactly what the algorithm does, we can still observe its performance during training. However, this begs the  question whether its performance on the training data is actually representative for the performance on unseen data.
Therefore, in the first part of this paper we provide a theoretical framework for learning an optimization algorithm based on its performance on a training set, that is, based on the empirical risk together with a generalization bound for the (true) risk. A popular framework that provides such generalization bounds is the PAC-Bayesian approach to learning, which we apply to the setting of learning-to-optimize. This yields the following informal version of our main theoretical result (compare Example~\ref{Exa:main_result_l2o}). It states that, with high probability we will observe a (training) data set $\rvData$ for which the given bound on the risk of the algorithm's output will hold uniformly in a so-called index $\idx \in \idxset$ and distribution $\mathbb{Q}$:
\begin{theorem}[Informal] \label{Thm_informal}
    Under mild boundedness assumptions on the optimization algorithm, the $\mathbb{Q}$-average population loss $\mathcal{R}_{\sublevelfunction}$ of the algorithm's output can be bounded by the $\mathbb{Q}$-average empirical loss $\hat{\mathcal{R}}_{\sublevelfunction}$ of the algorithm's output plus some remainder term $r_N$ that vanishes with the size $N$ of the data set, that is, for all $\varepsilon > 0$:
    \begin{align*}
    \prob_{\notation{\rvData}} \left\{\forall \idx \in \idxset, \ \forall \mathbb{Q} \in \probMeasures : 
    \mathbb{Q}[ \mathcal{R}_{\sublevelfunction}] \le \mathbb{Q}[\hat{\mathcal{R}}_{\sublevelfunction}] + r_N(\idx) \right\} \ge 1-\varepsilon  \,.
    \end{align*}
\end{theorem}
Especially, the uniformity in $\mathbb{Q}$ allows for \emph{learning} such a distribution.
This provides one possible answer to the question about theoretical guarantees for learning-to-optimize. However, while being a generalization bound, such a guarantee is a statement about \emph{relative} and not absolute values, that is, how the true risk \emph{compares to} the empirical risk. Thus, one still has to train the optimization algorithm in such a way that the empirical risk is indeed small enough to be worth the effort. This is particularly important in the area of learning-to-optimize, as there are already algorithms that can provably solve the given problems in a rather short amount of time. Hence, the second central question that arises is of a more practical nature and pertains to the actual training of such an algorithm:
\begin{center}
    \emph{
    How do we learn an optimization algorithm, such that its performance is clearly superior to the one achieved by a worst-case analysis?}
\end{center}

\begin{figure}[t!]
    \centering
    \includegraphics[width=0.95\textwidth]{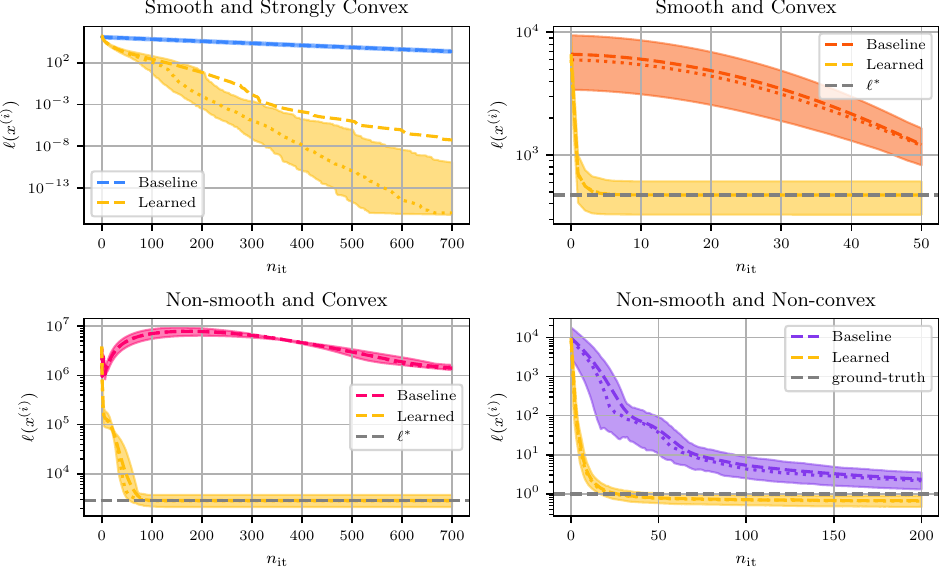}
    \caption{Some numerical results: Loss over iterations (mean as dashed and median as dotted line) of the learned algorithm compared to a standard choice.}
    \label{fig:introduction}
\end{figure}

\noindent
Therefore, in the second part of this work, we develop a concrete algorithmic realization, which allows for learning an optimization algorithm and evaluating the corresponding theoretical guarantee. This involves several key design choices that have not been used before and which are of interest in their own right. Furthermore, as empirical evaluation of our claims, we conduct four practically relevant experiments, all dealing with very different classes of functions, thereby demonstrating the wide applicability and strong practical performance of our approach. Figure~\ref{fig:introduction} provides a preview of some experimental results, and the details will be provided in Section~\ref{Sec:Experiments}. Each subplot compares the performance of the learned algorithm (yellow) to that of a standard algorithm on different problems ranging from smooth and strongly convex to non-smooth and non-convex. Since the learned algorithm is clearly superior in each case, this provides a possible answer to the question about how to train optimization algorithms. In summary, we provide a complete framework to train optimization algorithms with theoretical guarantees that are (in a certain sense) provably faster than their worst-case optimal counterparts. In particular, this work is a far reaching extension of our conference paper \citep{Sucker_Ochs_2023} by extending and clarifying the theoretical results in Sections~\ref{Sec_PAC}, and, in particular, by the algorithmic realization together with its evaluation in Sections~\ref{Sec:Learning_Procedure} and \ref{Sec:Experiments}, which additionally includes a probabilistic constraining procedure for sampling algorithms in Subsection~\ref{Sec:Sampling_under_probabilistic_constraints}. The entire code associated with this paper can be found at \url{https://github.com/MichiSucker/Learning-to-Optimize-with-PAC-Bayes}.

\subsection{Related Work}

The literature on both learning-to-optimize and the PAC-Bayes learning approach is vast. Hence, for learning-to-optimize we will mainly focus on approaches that provide some theoretical guarantees. Especially, this excludes many model-free approaches, which replace the whole update step with a learnable mapping such as a neural network. \cite{CCCHLWY2021} provide a good overview about the variety of approaches in learning-to-optimize, and good introductory references for the PAC-Bayesian approach are given by \cite{G2019}, \citet{Hellström_Durisi_Guedj_Raginsky_2023}, and \cite{Alquier_2024}. 

\subsubsection{Broader Context of Learning-to-Optimize} Optimization is an integral part of machine learning. Thus, learning-to-optimize has significant overlap with the areas of meta-learning (or \say{learning-to-learn}) and AutoML. The first one is a subset of learning-to-optimize, because it is mostly concerned with determining parameters of machine learning models \citep{Vilalta_Drissi_2002, Hospedales_Antoniou_Micaelli_Storkey_2021}. AutoML, however, more broadly refers to automating all steps necessary to create a machine learning application, which also includes the choice of an optimization algorithm and its hyperparameters \citep{Yao_Wang_Chen_Dai_Li_Tu_Yang_Yu_2018, Hutter_Kotthoff_Vanschoren_2019, He_Zhao_Chu_2021}. 

\subsubsection{Learning-to-Optimize with Guarantees} Learned optimization methods may lack theoretical guarantees for the sake of convergence speed \citep{CCCHLWY2021}. Yet, there are applications where a convergence guarantee is of highest priority: \cite{MMC2019} provide an example where a purely learning-based approach fails to reconstruct the crucial details in a medical image. Also, they prove convergence of their method by restricting the output to descent directions, for which mathematical guarantees exist. The basic idea is to trace the learned object back to, or constrain it to, a mathematical object with convergence guarantees. Similarly, \cite{SVWBDSB2016} provide sufficient conditions under which the learned mapping is a proximal mapping. Related schemes, under different assumptions and guarantees, are given by \cite{CWE2016}, \cite{TBF2017}, \cite{TG2018}, \cite{BCSB2018}, \cite{RLWCWY2019}, \cite{SWK2019}, \cite{TRPW2021} and \cite{CEM2021}. A major advantage of these methods is the fact that the number of iterations is not restricted a priori. However, a major drawback is their restriction to specific algorithms and problems. This contrasts with the approach of unrolling, pioneered by \cite{GL2010}, which limits the number of iterations, yet in principle can be applied to every iterative optimization algorithm. Here, \cite{XWGWW2016} study the convergence properties of the IHT algorithm, while \cite{CLWY2018} consider the unrolled ISTA. However, a difficulty in the theoretical analysis of unrolled algorithms is actually the notion of convergence itself, such that one rather has to consider the generalization performance. Only few works have addressed this: Either directly by means of Rademacher complexity \citep{CZRS2020}, or indirectly in form of a stability analysis \citep{KEKP2020}, as algorithmic stability is linked to generalization \citep{BE2000, BE2002, SSSS2010}. \textit{Our theoretical analysis corresponds to the approach of unrolling, that is, a fixed number of iterations. However, in the experiments we stay more closely to the iterative approach of learning an update step that can be applied for an arbitrary number of iterations.} 

\subsubsection{Design-Choices in Learning-to-Optimize} 
A major problem of many learned optimization algorithms, especially the ones based on recurrent neural networks (RNN), is their restriction to a certain number of iterations: They cannot be trained for an arbitrary number of iterations due to instabilities or memory bottlenecks. Further, often they do not generalize well to more iterations than they were trained for \citep{Andrychowicz_Denil_Gomez_Hoffman_Pfau_Schaul_Shillingford_DeFreitas_2016, Chen_Hoffman_Colmenarejo_Denil_Lillicrap_Botvinick_DeFreitas_2017, Lv_Jiang_Li_2018, CCCHLWY2021}. A typical way to mitigate this problem is to split the whole trajectory into smaller parts \citep{Andrychowicz_Denil_Gomez_Hoffman_Pfau_Schaul_Shillingford_DeFreitas_2016, Chen_Hoffman_Colmenarejo_Denil_Lillicrap_Botvinick_DeFreitas_2017, Metz_Maheswaranathan_Nixon_Freeman_SohlDickstein}. However, often this does not lead to fully satisfactory results either, such that other approaches have been proposed: To improve generalization, \citet{Lv_Jiang_Li_2018} introduce random scaling of the coordinates and the addition of a convex function to the objective. \citet{Wichrowska_Maheswaranathan_Hoffman_Colmenarejo_Denil_DeFreitas_SohlDickstein_2017} introduce a hierarchical RNN architecture, and additionally draw the number of unrollings and the unrolling length from a heavy-tailed exponential distribution. While achieving the needed generalization, this approach does not achieve the same wall-clock time as simple optimization algorithms. \citet{Metz_Maheswaranathan_Nixon_Freeman_SohlDickstein} replace the recurrent neural network with a multilayer perceptron (MLP), and they use two unbiased gradient estimators instead of one. Doing so they manage to train algorithms that are faster in wall-clock time than standard ones like Adam. 
\citet{Chen_Zhang_Jingyang_Chang_Liu_Amini_Wang_2020} consider training techniques in general, and introduce a progressive scheme that gradually increases the unrolling length, as well as an imitation learning approach to learn to mimic analytic optimizers. \\
Besides the optimizer, a crucial design choice in learning-to-optimize is that of the loss function. Typically, either the final loss or a weighted sum of the losses along the iterations is used \citep{CCCHLWY2021}.
\textit{We introduce a new loss function for training optimization algorithms, motivated by an intuitive theoretical argument. Further, we use a single learned update based on MLPs instead of an RNN, and we split the trajectory into subtrajectories and randomize its total length, however, in a new way.}

\subsubsection{PAC-Bayesian Bounds through Change-of-Measure}
PAC is an acronym for \emph{Probably Approximately Correct}, and PAC-\emph{Bayes} refers to the fact that one considers distributions instead of points \citep{Alquier_2024}. This framework allows for giving high probability bounds on the risk, either as an oracle or as an empirical bound. The key ingredient is a change-of-measure inequality, the choice of which strongly influences the corresponding bound. The one used most often is based on a variational representation of the Kullback--Leibler divergence due to \cite{DV1975}, employed, for example, by \cite{C2004, C2007}. Yet, not all bounds are based on a variational representation, that is, holding uniformly over all posterior distributions  \citep{RKSS2020}. While many bounds involve the Kullback--Leibler divergence as measure of proximity \citep{Mc2003_1, Mc2003_2, S2002, LS2002, GLL2009}, other divergences have been used: \cite{HJ2014} prove an inequality for the $\chi^2$-divergence, which is also used by \cite{L2017}. \cite{BGLR2016} and \cite{AG2018} use the Renyi-divergence ($\alpha$-divergence). \cite{OH2021} propose PAC-bounds based on f-divergences, which include the Kullback--Leibler-, $\alpha$- and $\chi^2$-divergences. More recently, \cite{Amit_Epstein_Moran_Meir_2022} propose to replace the Kullback-Leibler divergence by so-called \say{integral probability metrics}, which encompass, for example, the Wasserstein distance that obeys many favorable properties and also captures the geometry of the underlying space \citep[see][]{Villani_2009}. Motivated by this, \cite{Haddouche_Guedj_2023} also investigate PAC-Bayesian generalization bounds for the Wasserstein distance and their interplay with the output of optimization algorithms. A major advantage of using the Wasserstein distances instead of the Kullback-Leibler divergence is the fact that it does not constrain the support of the distribution a-priori through the choice of the prior. On the other hand, it demands assumptions on the loss function, which are not necessarily satisfied in learning-to-optimize. \textit{We give a general PAC-Bayesian theorem based on exponential families. Here, prior, posterior, divergence and data dependence are given naturally. Further, it allows for implementing an abstract learning framework that can be applied to a wide variety of algorithms.}

\subsubsection{Boundedness of the Loss Function}
A major drawback of many of the existing PAC-Bayes bounds is the assumption of a bounded loss-function. This assumption is mainly used to apply some exponential-moment inequality like the Hoeffding- or Bernstein-inequality \citep{RKSS2020, Alquier_2024} and several ways have been developed to circumvent this problem: \cite{GLL2009} explicitly include the exponential-moment in the bound, \cite{ARC2016} use so-called Hoeffding- and Bernstein-assumptions, \cite{C2004} restricts to the sub-Gaussian or sub-Gamma case. Another possibility to ensure the generalization or exponential-moment bounds is to use properties of the algorithm: \cite{L2017} uses algorithmic stability to provide PAC-Bayes bounds for SGD. \textit{We consider suitable properties of optimization algorithms aside from algorithmic stability to ensure the exponential-moment bounds.}

\subsubsection{Minimization of the PAC-Bound}
PAC-bounds relate the true risk to other terms such as the empirical risk. Yet, they do not directly say anything about the absolute numbers. Thus, learning procedures based on the PAC-Bayesian theory typically aim to minimize this bound: \cite{LC2001} compute non-vacuous generalization bounds through a combination of PAC-bounds with a sensitivity analysis. \cite{Dziugaite_Roy_2017} extend this by minimizing the PAC-bound directly. \cite{PRSS2021} also consider learning as minimization of the PAC-Bayes bound and provide tight generalization bounds. \cite{TIWS2017} are able to solve the  minimization problem resulting from their PAC-bound by alternating minimization. \textit{We follow this approach and consider learning as minimization of the PAC-Bayesian bound.}

\subsubsection{Choice of the Prior}
A common difficulty in learning with PAC-Bayesian bounds is the choice of the prior distribution, as it heavily influences the performance of the learned models and the generalization bound \citep{C2004, Dziugaite_Hsu_Gharbieh_Arpino_Roy_2021, PRSS2021}. In part, and especially for the Kullback-Leibler divergence, this is due to the fact that the divergence term can dominate the bound, keeping the posterior close to the prior. This leads to the idea of choosing a data- or distribution-dependent prior \citep{S2002, ParradoHernandez_Ambroladze_ShaweTaylor_Sun_2012, Lever_Laviolette_ShaweTaylor_2013, Dziugaite_Roy_2018, PRSS2021}, which, by using an independent subset of the data set, gets optimized to yield a good performance. \textit{The prior distribution strongly influences the performance of our learned algorithms. Thus, we use a data-dependent prior. Further, we show how the prior can be used for preserving essential properties during learning: It is key to control the trade-off between convergence guarantee and convergence speed.}

\subsubsection{More Generalization Bounds}
There are more areas of machine learning research that study generalization bounds. Importantly, the field of \say{stochastic optimization} (SO) provides generalization bounds for specific algorithms. The main differences to our setting are the learning approach and the assumptions made:
\begin{itemize}[leftmargin=*]
    \item Instead of a distribution over hpyerparameters, the algorithms in SO generate a point estimate, and one studies the properties of this point in terms of the stationarity measure of the true risk functional \citep{BCN2018, DavisDrusvyatskiy2022, Bianchi_Hachem_Schechtman_2022}.
    \item Instead of an abstract algorithm, the setting in SO is more explicit. Thus, more assumptions have to be made. Typical assumptions are (weak) convexity \citep{SSSS2009_SCO, DavisDrusvyatskiy2019}, bounded gradients \citep{DefossezBottouBachUsunier2022}, bounded noise \citep{DavisDrusvyatskiy2022}, or smoothness \citep{KavisLevyCevher2022}.
\end{itemize}
\emph{We provide a principled way to learn a distribution over general hyperparameters of an abstract algorithm under weak assumptions and go explicitly beyond analytically tractable quantities. Therefore, the methodology is independent of the chosen implementation. \\ } 

\noindent
The rest of the paper is structured as follows: In Section~\ref{Sec:Problem_Setup} we introduce the notation and provide a formal description of the setting. In Section~\ref{Sec_PAC}, we derive the general PAC-Bayesian theorem and relate it to other existing bounds. In Section~\ref{Subsec:learning_with_guarantees}, we identify properties of optimization algorithms that allow to apply the PAC-Bayesian theorem. As this strongly relies on assumptions on the prior distribution, we provide a probabilistic constraining procedure that allows to enforce such constraints in Section~\ref{SubSec_GuaranteeConvProb}. Then, in Section~\ref{Sec:Learning_Procedure} we describe the learning procedure and our design choices for learning-to-optimize, and in Section~\ref{Sec:Experiments} we conduct the experiments.

\section{Problem Setup \& Assumptions}\label{Sec:Problem_Setup}
In this section we establish the notation, formalize the setting, and state the main assumptions that are used throughout the remainder of the text.

\subsection{Notation} We will endow every topological space $\genSpaceOne$ with the corresponding Borel-$\sigma$-algebra $\algebra{B}(\genSpaceOne)$, and, given a product space $\genSpaceOne \times \genSpaceTwo$ of two measurable spaces $(\genSpaceOne, \algebraGenSpaceOne)$ and $(\genSpaceTwo, \algebraGenSpaceTwo)$, we endow it with the product-$\sigma$-algebra $\algebraGenSpaceOne \otimes \algebraGenSpaceTwo$. We will denote the product space of a generic number of spaces $\genSpaceOne_1, ..., \genSpaceOne_n$ by $\prod_{i=1}^n \genSpaceOne_i$, and the product-$\sigma$-algebra by $\bigotimes_{i=1}^n \algebra{B}(\genSpaceOne_i)$. If all spaces are equal, this is abbreviated as $\genSpaceOne^n$. For a function $f: \genSpaceOne \times \genSpaceTwo \to \genSpaceThree$, $f(\genRealOne,\cdot): \genSpaceTwo \to \genSpaceThree$ denotes the map $\genRealTwo \mapsto f(\genRealOne, \genRealTwo)$ with fixed element $\genRealOne \in \genSpaceOne$.
Similarly, for a set $\set{C} \subset \genSpaceOne \times \genSpaceTwo$, the section of $\set{C}$ for fixed $\genRealOne \in \genSpaceOne$ is denoted by $\set{C}_{\genRealOne} := \{\genRealTwo \in \genSpaceTwo \ : \ (\genRealOne, \genRealTwo) \in \set{C}\}$. In general, generic sets are denoted in typewriter font, for example $\set{A}$, and $\mathds{1}_\set{A}$ denotes the function that is equal to one for $\genRealOne \in \set{A}$ and zero else, while $\iota_\set{A}$ denotes the function that is equal to zero for $\genRealOne \in \set{A}$ and $+\infty$ else.\footnote{We omit the name here, as both $\mathds{1}_\set{A}$ and $\iota_\set{A}$ are called \say{indicator function}. The former in probability theory, the latter in optimization.} 
Given a measurable space $(\genSpaceOne, \algebraGenSpaceOne)$, a measure $\mu$ and a measurable function $f \ge 0$,  $\mu [f]$ denotes the integral of $f$ w.r.t. $\mu$, while $f \cdot \mu$ denotes the measure given by $(f \cdot \mu) [\set{A}] = \int_\set{A} f(\genRealOne) \ \mu(d\genRealOne)$, that is, $(f \cdot \mu) [\genSpaceOne] = \mu [f]$ and $(f \cdot \mu) [\set{A}] = \mu [f\cdot \mathds{1}_\set{A}]$. Hence, $f \cdot \mu$ is absolutely continuous w.r.t. $\mu$, written as $f \cdot \mu \ll \mu$, with $f$ being the corresponding density. Here, the set of all measures on U will be denoted by $\measures(\genSpaceOne) := \{ \mu: \algebraGenSpaceOne \to [0, \infty] \ : \ \text{$\mu$ is a measure}\}$, and the set of all probability measures that are absolutely continuous w.r.t. $\mu \in \measures(\genSpaceOne)$ are denoted by $\probMeasures(\mu) := \{\nu \in \measures(\genSpaceOne) \ : \ \text{$\nu [\genSpaceOne] = 1$ and $\nu \ll \mu$}\}$. In this context, the Kullback-Leibler divergence between two measures $\nu$ and $\mu$ is defined as 
$$
\divergence{\rm{KL}}{\mu}{\nu} = \begin{cases}
    \mu[\log(f)] = \int_{\genSpaceOne} \log(f(\genRealOne)) \ \mu(d\genRealOne), &\text{$\mu \ll \nu$ with density $f$}\,, \\ +\infty, &\text{otherwise}\,.
\end{cases}
$$ 
For the rest of the manuscript, we will fix a probability space $\pspace$, and if $\mu = \prob$ is the probability measure, the corresponding expectation is denoted by $\expectation [f] := \prob [f] = \int_\Omega f(\omega) \ \prob(d\omega)$. 
Here, we will write random variables in upper-case and corresponding realizations in lower-case with the same symbol, for example $\notation{\genRvOne = \genRealOne}$. Given two random variables $\notation{\genRvOne}: \pspace \to \genSpaceOne$ and $\notation{\genRvTwo}: \pspace \to \genSpaceTwo$, integration of a measurable function $f$ on $\genSpaceOne \times \genSpaceTwo$ w.r.t. the induced probability measure $\prob_{(\notation{\genRvOne, \genRvTwo})}$ is specified by the subscript $(\notation{\genRvOne, \genRvTwo})$, that is: 
$$
\expectation \left[ f(\notation{\genRvOne, \genRvTwo}) \right] = \int_\Omega f\left(\notation{\genRvOne, \genRvTwo}\right)(\omega) \ \prob(d\omega) = \int_{\genSpaceOne \times \genSpaceTwo} f(\notation{\genRealOne},\notation{\genRealTwo}) \ \prob_{(\notation{\genRvOne, \genRvTwo})}(d\notation{\genRealOne},d\notation{\genRealTwo}) 
= \expectation_{(\notation{\genRvOne, \genRvTwo})} \left[ f \right]  \,.
$$ 
If we have a regular version of the conditional distribution of $\notation{\genRvTwo}$, given $\notation{\genRvOne}$, denoted by $\prob_{\notation{\genRvTwo \vert \genRvOne}}$, the joint distribution $\prob_{(\notation{\genRvOne, \genRvTwo})}$ can be disintegrated into the product $\prob_{\notation{\genRvOne}} \otimes \prob_{\notation{\genRvTwo \vert \genRvOne}}$ of the marginal $\prob_{\notation{\genRvOne}}$ and the probability kernel $(x, \set{B}) \mapsto \prob_{\notation{\genRvTwo \vert \genRvOne = \genRealOne}} [\set{B}]$, which allows us to use the notation:
\begin{align*}
    \expectation[f(\notation{\genRvOne, \genRvTwo})] 
    = \int_{\genSpaceOne} \int_{\genSpaceTwo} f(\notation{\genRealOne}, \notation{\genRealTwo}) \ \prob_{\notation{\genRvTwo \vert \genRvOne = \genRealOne}} (d\notation{\genRealTwo}) \ \prob_{\notation{\genRvOne}}(d\notation{\genRealOne}) = \expectation_{\notation{\genRvOne}} \left[ \expectation_{\notation{\genRvTwo \vert \genRvOne = \genRealOne}} \left[ f(\notation{\genRealOne}, \cdot) \right] \right]\,.
\end{align*}
Note that changing the order of integration is not allowed in this case. However, if $\notation{\genRvOne}$ and $\notation{\genRvTwo}$ are independent, their joint distribution is given by the product $\prob_{\notation{\genRvOne}} \otimes \prob_{\notation{\genRvTwo}}$ for which Fubini's theorem is applicable, and the iterated integration is clarified by the subscripts $\notation{\genRvOne, \genRvTwo}$:
\begin{align*}
    \expectation \left[ f(\notation{\genRvOne, \genRvTwo}) \right] = \int_{\genSpaceOne} \int_{\genSpaceTwo} f(\notation{\genRealOne}, \notation{\genRealTwo}) \ \prob_{\notation{\genRvTwo}}(d\notation{\genRealTwo}) \ \prob_{\notation{\genRvOne}} (d\notation{\genRealOne}) 
    = \expectation_{\notation{\genRvOne}} \left[ \expectation_{\notation{\genRvTwo}} \left[ f(\notation{\genRealOne}, \cdot) \right]\vert_{\notation{\genRealOne} = \notation{\genRvOne}} \right]\,.
\end{align*}
Finally, our theoretical results rely on the notions of probability kernels and exponential families, whose definitions are recalled in Appendix~\ref{App:Missing_definitions}.

\subsection{Main Assumptions and Definitions for Learning Optimization Algorithms}\label{subsec:main_assumptions}
We assume that we are given a distribution over loss-functions with a specific structure, which is modelled by a random variable:
\begin{Ass}\label{Assumption_random_parametric_problem}
    We are given a Polish space $\notation{\parSpace}$ (separable and complete metrizable topological space) and a non-negative and measurable loss-function $\ell: \R^n \times \notation{\parSpace} \to [0, +\infty]$. Further, for some $N \in \N$, we are given i.i.d. random variables $\notation{\rvPar, \rvPar_1, ..., \rvPar_N}: \pspace \to \notation{\parSpace}$.
\end{Ass}
\noindent
Then, ideally, we would like to find a solution to each realization of the random objective:
\begin{equation}\label{Eq_random_parametric_opt_prob}
    \text{Find $x^{\notation{*}}: \notation{\parSpace} \to \R^n$, s.t. $x^{\notation{*}}(\notation{\realPar}) \in \argmin_{x \in \R^n} \ell(x, \notation{\realPar}) \quad \prob_{\rvPar}-a.s.$}
\end{equation}
However, we will only solve a relaxed version of \eqref{Eq_random_parametric_opt_prob} and provide \emph{generalization bounds} for the \emph{average performance} after training on a data set.
\begin{definition}
    The measurable function 
    $
        \notation{\rvData}: \pspace \longrightarrow \notation{\parSpace^N}, \quad   
        \omega \mapsto \left(\notation{\rvPar_1, ..., \rvPar_N} \right)(\omega)
    $
    is called a \emph{data set}, and if the random variables $\rvPar_1, ..., \rvPar_N$ are i.i.d., that is, $\mathbb{P}_{\notation{\rvData}} = \bigotimes_{i=1}^N \mathbb{P}_{\notation{\rvPar}_i} = \bigotimes_{i=1}^N \mathbb{P}_{\notation{\rvPar}}$, it is called an \emph{i.i.d. data set}. Further, $\notation{\parSpace^N}$ is called the \emph{data-space}.
\end{definition}
\noindent
PAC-Bayesian generalization bounds involve a so-called posterior distribution, which usually is a \say{data-dependent distribution}. As also pointed out by \cite{RKSS2020}, this is an instance of a probability kernel (also called a \say{stochastic-} or \say{Markov kernel}):

\begin{definition}\label{Def_Data_Dependent_Distribution}
    Let $\notation{\rvData}$ be a data set with data-space $\notation{\parSpace^N}$, and let $\genSpaceOne$ be a measurable space. A probability kernel from $\notation{\parSpace^N}$ to $\genSpaceOne$ is called a \emph{data-dependent distribution} on $\genSpaceOne$.
\end{definition}
\noindent
For solving problem \eqref{Eq_random_parametric_opt_prob}, for every realization $\notation{\realPar}$ of $\notation{\rvPar}$, we apply an optimization algorithm $\mathcal{A}$ to $\ell(\cdot, \notation{\realPar})$. For this, we consider a similar setting as \cite{L2017}, that is, randomized algorithms are considered as deterministic algorithms with randomized hyperparameters: 
\begin{definition}
    Let $\notation{\hypSpace}$ be a Polish space and $n \in \N$. A measurable function 
    $$
        \mathcal{A}: \notation{\hypSpace} \times \R^n \times \notation{\parSpace} \longrightarrow \mathbb{R}^n, \quad (\notation{\realHyp}, \iter{x}{0}, \notation{\realPar}) \mapsto \mathcal{A}(\notation{\realHyp}, \iter{x}{0}, \notation{\realPar})\,,
    $$
    is called a \emph{parametric algorithm}. $\R^n$ is the space of the optimization variable, $\notation{\parSpace}$ the space of the parameters of the loss function, and $\notation{\hypSpace}$ the space of the hyperparameters of the algorithm.
\end{definition}

\noindent
Please note that $\mathcal{A}$ corresponds to the \emph{whole} algorithm, that is, for an iterative algorithm its output is the final iterate. In the PAC-Bayesian approach, learning $\mathcal{A}$ refers to finding a distribution $\mathbb{Q}$ on $\notation{\hypSpace}$ based on its performance on a data set $\notation{\rvData}$. For this, one needs a reference distribution, called the \emph{prior}, which can (and should) encode prior knowledge about suitable choices of hyperparameters:
\begin{Ass}\label{Assumption_parametric_optimization_algorithm}
    We are given a parametric algorithm $\mathcal{A}$ with Polish hyperparameter space $\notation{\hypSpace}$, and a (prior) distribution $\mathbb{P}_{\notation{\rvHyp}}$ on $\notation{\hypSpace}$ that is induced by a random variable $\notation{\rvHyp}: \pspace \longrightarrow \notation{\hypSpace}$, which is independent of $\notation{\rvData}$ and $\notation{\rvPar}$. Further, the initialization $x^{(0)} \in \R^n$ is given and fixed.
\end{Ass}

\begin{Not}
    To simplify the notation, we use the short-hand $\ell(\notation{\realHyp, \realPar}) := \ell (\mathcal{A}(\notation{\realHyp, \realPar}), \notation{\realPar})$. Furthermore, if not needed explicitly, $\iter{x}{0}$ and $\pspace$ will not be mentioned in the following.
\end{Not}

\begin{Def}
    Suppose $\notation{\rvPar}$ and $\ell$ satisfy Assumption~\ref{Assumption_random_parametric_problem}, and $\mathcal{A}$ satisfies Assumption~\ref{Assumption_parametric_optimization_algorithm}. The \emph{risk} of $\mathcal{A}$ is defined as the measurable function:
    \begin{align*}
        \mathcal{R}: \notation{\hypSpace} \longrightarrow [0, +\infty], \
        \notation{\realHyp} \mapsto \mathbb{E} \left[ \ell \left(\mathcal{A}(\notation{\realHyp, \rvPar}), \notation{\rvPar}\right) 
        \right] 
        = \expectation[\ell(\notation{\realHyp, \rvPar})] 
        = \expectation_{\notation{\rvPar}} [\ell(\notation{\realHyp}, \cdot)]\,.
    \end{align*}
    Similarly, for an i.i.d. dataset $\notation{\rvData} = (\notation{\rvPar_1, ..., \rvPar_N})$ the \emph{empirical risk} is defined as:
    $$
    \hat{\mathcal{R}}: \notation{\hypSpace} \times \notation{\parSpace^N} \longrightarrow [0, +\infty], \quad 
    (\notation{\realHyp, \rvData}) \mapsto \hat{\mathcal{R}}(\notation{\realHyp, \rvData}) = \frac{1}{N} \sum_{i=1}^N \ell \left(\notation{\realHyp, \rvPar_i} \right) \,.
    $$
\end{Def}

\noindent
The following theory is based on exponential families, which is a very flexible class of distributions. We highlight the data-dependency in the following adjusted definition:

\begin{Def}
    Let $\emptyset \neq\idxset$ be an index set, $\notation{\rvData}$ a data set with data-space $\notation{\parSpace^N}$, and let $\genSpaceOne$ be a measurable space. A family of probability kernels $(\mathbb{Q}_{\idx})_{\idx \in \idxset}$ from $\notation{\parSpace^N}$ to $\genSpaceOne$ is called a \emph{data-dependent exponential family} (in $\natPar$ and $\suffStat$), if there is a probability measure $\mu$ on $\genSpaceOne$, functions $\natPar: \idxset \to \R^k$, $\logPartition: \idxset \times \parSpace^N \to (0, +\infty)$, and measurable functions $\suffStat: \genSpaceOne \times \notation{\parSpace^N} \to \R^k$, $\expFamFunc: \genSpaceOne \to (0, +\infty)$, such that
    $\mathbb{Q}_{\idx}(\notation{\realData}) = \expFamFunc \logPartition(\idx, \realData) \exp\left(\sprod{\natPar(\idx)}{\suffStat(\cdot, \notation{\realData})}\right) \cdot \mu$ for every $\idx \in \idxset$, $\notation{\realData \in \parSpace^N}$, that is, $\mathbb{Q}_{\idx}(\notation{\realData}, \set{B}) = \int_\set{B} \expFamFunc(\genRealOne) \logPartition(\idx, \realData) \exp(\sprod{\natPar(\idx)}{\suffStat(\genRealOne, \notation{\realData})}) \ \mu(d\genRealOne)$, $\set{B} \in \algebra{B}(\genSpaceOne)$.
\end{Def}

\noindent
We introduce data-dependency through $\suffStat$, since it strongly affects the shape of the distribution and, contrary to $\natPar$, is defined on the underlying space $\genSpaceOne$. Since we want to learn a distribution over hyperparameters $\notation{\realHyp} \in \notation{\hypSpace}$, we make the following assumption:

\begin{Ass}\label{Ass:exponential_family}
    On the hyperparameter space $\notation{\hypSpace}$, we are given a data-dependent exponential family $(\mathbb{Q}_{\idx})_{\idx \in \idxset}$ in $\natPar$ and $\suffStat$ with dominating probability measure $\mu = \prob_{\notation{\rvHyp}}$, such that the map $\notation{\realHyp} \mapsto \expFamFunc(\notation{\realHyp}) \exp(\sprod{\natPar(\idx)}{\suffStat(\notation{\realHyp, \realData})})$ is non-trivial and integrable w.r.t. $\prob_{\notation{\rvHyp}}$ for every $\idx \in \idxset$, $\notation{\realData \in \parSpace^N}$, that is, $\expectation_{\notation{\rvHyp}} \left[ \expFamFunc \exp(\sprod{\natPar(\idx)}{\suffStat(\cdot, \notation{\realData})}) \right] \in (0, \infty)$.
\end{Ass}
Then, as shown in Lemma~\ref{Lem:Q_data_dependent} in Appendix~\ref{App:Missing_Lemmas}, every member of the data-dependent exponential family is indeed a data-dependent distribution on $\notation{\hypSpace}$. In the following, the last integral in Assumption~\ref{Ass:exponential_family} will be of great interest. Here, we will use a similar notation as in \cite{BN2014} and denote
\begin{equation} \label{Eq_DefKappa}
    \begin{split}
        &c(\idx, \notation{\realData}) := \int_{\notation{\hypSpace}} \expFamFunc(\notation{\realHyp}) \exp(\sprod{\natPar(\idx)}{\suffStat(\notation{\realHyp, \realData})}) \ \prob_{\notation{\rvHyp}}(d\notation{\realHyp}) 
        = \expectation_{\notation{\rvHyp}} \left[\expFamFunc \exp(\sprod{\natPar(\idx)}{\suffStat(\cdot, \notation{\realData})})\right]\,,\\
        &\kappa(\idx, \notation{\realData}) := \log\left(c(\idx, \notation{\realData})\right) = \log \left( \expectation_{\notation{\rvHyp}} \left[\expFamFunc \exp\left(\sprod{\natPar(\idx)}{\suffStat(\cdot, \notation{\realData})} \right)\right] \right) \,.
    \end{split}
\end{equation}
With this notation, it holds that $\logPartition(\idx, \realData) = c(\idx, \realData)^{-1}$.

\begin{Rem}
    \begin{itemize}
        \item[(i)] If $\natPar$ describes a lower-dimensional manifold in $\R^k$, $(\mathbb{Q}_{\idx})_{\idx \in \idxset}$ is called a \emph{curved} exponential family \citep{E1975}, whose properties might differ from the ones for \emph{linear} exponential families, for example, convexity of the map $\idx \mapsto \logPartition(\idx, \realData)$.
        \item[(ii)] In PAC-Bayes, the dominating measure $\prob_{\notation{\rvHyp}}$ is usually referred to as \emph{prior} and every distribution $\mathbb{Q} \in \probMeasures(\prob_{\notation{\rvHyp}})$ is referred to as a \emph{posterior}. This deviates from the standard definitions of prior and posterior in Bayesian statistics.  
        \item[(iii)] In general, the integrability assumption is restrictive, as it affects the choice of $\expFamFunc, \natPar$ and $\suffStat$. However, in Section~\ref{Subsec:learning_with_guarantees} we will \emph{construct} $\natPar$ and $\suffStat$ such that this holds anyway.
        \item[(iv)] In the special case $\expFamFunc \equiv 1$ and $\natPar(\idx) \equiv \idx$, the map $\idx \mapsto c(\idx, \realData)$ is the moment-generating function of the random variable $\suffStat(\notation{\rvHyp, \realData})$. Similarly, in this case $\idx \mapsto \kappa(\idx, \realData)$ is the corresponding cumulant-generating function.
    \end{itemize}
\end{Rem}
Finally, we will restrict $\idxset$ to a compact set. This is needed to get a uniform bound in $\idx$ \citep[see][]{LC2001, C2007, Alquier_2024}.

\begin{Ass}\label{Ass:Lambda_compact}
    $\idxset$ is a compact set with finite covering $\mathcal{O} := \{ \set{O}_1, ..., \set{O}_{\coveringNumber}\}$, that is, $\idxset \subset \bigcup_{i=1}^{\coveringNumber} \set{O}_i$, such that there is a constant $\coveringConstant$, which, for every $\notation{\realData \in \parSpace^N}$, allows for the bound $\max_{i=1,..., \coveringNumber} \ \sup_{\idx, \idx' \in \set{O}_i} \kappa(\idx, \notation{\realData}) - \kappa(\idx', \notation{\realData}) \le \coveringConstant$.
\end{Ass}

\begin{Rem}
    The non-trivial part of this assumption is the existence of the constant $\coveringConstant$ for the given finite covering. It does hold, for example, if $\idxset$ is a finite set ($\coveringNumber = \abs{\idxset}$, $\coveringConstant = 0$), or, if $\left(\idxset, \rho \right)$ is a compact metric space and $\kappa$ is Lipschitz-continuous in $\idx$ (uniformly in $\notation{\realData}$) with Lipschitz constant $L$, such that $\coveringConstant = L \cdot \max_{i = 1,...,\coveringNumber} \mathrm{diam} \ \set{O}_i$, where the diameter of a set $\set{A}$ is given by $\mathrm{diam} \ \set{A} = \sup_{x,y \in \set{A}} \rho (x,y)$.
\end{Rem}

\section{General PAC-Bayesian Theorem}\label{Sec_PAC}

In this section we prove the general PAC-Bayesian bound for data-dependent exponential families, which then can be specialized into a generalization bound of the learned parametric optimization algorithm $\mathcal{A}$. It is based on the following two lemmas, whose proofs can be found in Appendix~\ref{App:Proof_Donsker_Varadhan} and \ref{App:Proof_Compact_Lambda}, respectively. The first lemma is a form of the Donsker--Varadhan variational formulation and yields uniformity in the distributions $\mathbb{Q}$, while the second lemma yields uniformity in $\idx \in \idxset$ by controlling $\idx \mapsto \kappa(\idx, \notation{\realData})$ for every $\notation{\realData \in \parSpace^N}$.
\begin{Lem}\label{Lem_DonskerVaradhanForExponentialFamilies}
    Suppose that Assumption~\ref{Ass:exponential_family} holds and define $\kappa$ as in \eqref{Eq_DefKappa}. Then for every $\idx \in \idxset$ and $\notation{\realData \in \parSpace^N}$ it holds that 
    $\kappa(\idx, \notation{\realData}) = \sup_{\mathbb{Q} \in \probMeasures(\mathbb{P}_{\notation{\rvHyp}})} \mathbb{Q}\left[ \sprod{\natPar(\idx)}{\suffStat(\cdot, \notation{\realData})} + \log(\expFamFunc) \right] - \divergence{\rm{KL}}{\mathbb{Q}}{\mathbb{P}_{\notation{\rvHyp}}}$.
    Furthermore, for every $\idx \in \idxset$, the supremum is attained at $\mathbb{Q}_{\idx}(\notation{\realData})$.
\end{Lem}

\begin{Lem}\label{Lem:Compact_lambda}
    Suppose that Assumption~\ref{Ass:Lambda_compact} holds and assume that $\mathbb{P} \{ \kappa(\idx, \notation{\rvData}) > \notation{t}  \} \le \exp(-\notation{t})$ for all $\notation{t} \in \mathbb{R}$ and $\idx \in \idxset$. Then $\mathbb{P} \{ \sup_{\idx \in \idxset} \kappa(\idx, \notation{\rvData}) \le \log ( \coveringNumber/\epsilon ) + \coveringConstant \} \ge 1-\epsilon$.
\end{Lem}

\begin{theorem}\label{Thm_PACBayes_exponential_familiy}
    Suppose that Assumptions~\ref{Ass:exponential_family} and \ref{Ass:Lambda_compact} hold, and assume that $\mathbb{E}_{\notation{\rvData}}[c(\idx, \cdot)] \le 1$ for all $\idx \in \idxset$. Then, it holds that:
    \begin{align*}
        \prob \Bigl\{ &\forall \idx \in \idxset, \ \forall \mathbb{Q} \in \probMeasures(\mathbb{P}_{\notation{\rvHyp}})  : \\ 
        &\mathbb{Q}\left[ \sprod{\natPar(\idx)}{\suffStat(\cdot, \notation{\realData})} + \log(\expFamFunc) \right]\vert_{\notation{\realData = \rvData}} 
        \le 
        \divergence{\rm{KL}}{\mathbb{Q}}{\mathbb{P}_{\notation{\rvHyp}}} + \log\left(\frac{\coveringNumber}{\varepsilon}\right) + \coveringConstant \Bigr \} \ge 1-\varepsilon \,.
    \end{align*}
\end{theorem}

\begin{proof}
Applying Markov's inequality to the non-negative random variable $c(\idx, \notation{\rvData})$ yields for $t \in \R$, $\idx \in \idxset$:
$$
    \mathbb{P} \left\{ c(\idx, \notation{\rvData}) > \exp(\notation{t}) \right\} \le \frac{\mathbb{E}[c(\idx, \notation{\rvData})]}{\exp(\notation{t})} \le \exp(-\notation{t}) \,.
$$
This implies that $\mathbb{P} \left\{\kappa(\idx, \notation{\rvData}) > \notation{t} \right \} \le \exp(-\notation{t})$. Hence, Lemma~\ref{Lem:Compact_lambda} is applicable and gives:
$$
    \mathbb{P} \left\{ \sup_{\idx \in \idxset} \kappa(\idx, \notation{\rvData}) \le \log \left( \frac{\coveringNumber}{\varepsilon} \right) + \coveringConstant \right\} \ge 1-\varepsilon \,.
$$
Using Lemma~\ref{Lem_DonskerVaradhanForExponentialFamilies} gives:
$$
    \mathbb{P} \left\{ \sup_{\idx \in \idxset}  \sup_{\mathbb{Q} \in \probMeasures(\mathbb{P}_{\notation{\rvHyp}})} \mathbb{Q}[\sprod{\natPar(\idx)}{\suffStat(\cdot, \notation{\realData})} + \log(\expFamFunc)]\vert_{\notation{\realData = \rvData}} - \divergence{\rm{KL}}{\mathbb{Q}}{\mathbb{P}_{\notation{\rvHyp}}} \le \log \left( \frac{\coveringNumber}{\varepsilon} \right) + \coveringConstant \right\} \ge 1-\varepsilon \,.
$$
Simply rearranging and reformulating yields the result.
\end{proof}

\begin{Rem}
    \begin{itemize}[leftmargin=25pt]
        \item[(i)] Note that the statement is still true for a data-dependent prior $\prob_{\notation{\rvHyp}}$: Given another independent data set $\notation{\rvData'}$, one needs to assume that $\expectation \left[ c(\idx, (\notation{\rvData, \rvData'})) \right] \le 1$. 
        \item[(ii)] In Section~\ref{Subsec:learning_with_guarantees} we provide sufficient conditions s.t. $\mathbb{E}[c(\idx, \notation{\rvData})] \le 1$ holds for all $\idx > 0$.
        \item[(iii)] Typically, $\coveringNumber$ is (related to) the covering-number of $\idxset$, and $\log(\coveringNumber)$ bears the intrinsic dimension of $\idxset$. Thus, in full generality, it might be large. For us, however, it only has a minor influence, since $\idxset \subset \R$, and the empirical risk is typically much larger.
        \item[(iv)] Thanks to the reviewers we became aware of the monograph by \citet{Hellström_Durisi_Guedj_Raginsky_2023}, which proposes a similar general PAC-Bayesian theorem. On first sight, it seems like theirs is more general than ours. However, Example~\ref{Exa:relation_to_other_PAC_bounds} clarifies this. Furthermore, we want to remark that the first version of our Theorem~\ref{Thm_PACBayes_exponential_familiy} appeared in 2022.
    \end{itemize}
\end{Rem}

\noindent
For the rest of the paper, we set $\expFamFunc \equiv 1$, such that $\log(\expFamFunc) \equiv 0$. The following corollary shows an example of how to transform Theorem~\ref{Thm_PACBayes_exponential_familiy} into a high-probability bound on the risk. The proof is given in Appendix~\ref{App:Proof_Cor_generalization_bound}.

\begin{Cor}[PAC-Bayesian Generalization Bound]\label{Cor_generalization_bound_exp_family}
    Denote $\suffStat$ by $\suffStat = (\suffStat^{(1)}, \suffStat^{(r)})$ with $\suffStat^{(r)} := (\suffStat^{(2)}, ..., \suffStat^{(k)})$ and $\natPar$ by $\natPar = (\natPar^{(1)}, \natPar^{(r)})$ with $\natPar^{(r)} := (\natPar^{(2)}, ..., \natPar^{(k)})$. If $\suffStat^{(1)} = \mathcal{R} - \hat{\mathcal{R}}$ and $\natPar^{(1)} > 0$, the following are equivalent for any $\idx \in \idxset$, $\notation{\realData \in \parSpace^N}$, $\mathbb{Q} \in \probMeasures(\prob_{\notation{\rvHyp}})$:
    \begin{itemize}
        \item[(i)] $\mathbb{Q}\left[ \sprod{\natPar(\idx)}{\suffStat(\cdot, \notation{\realData})} \right] 
        \le \divergence{\rm{KL}}{\mathbb{Q}}{\mathbb{P}_{\notation{\rvHyp}}} + \log\left(\frac{\coveringNumber}{\varepsilon}\right) + \coveringConstant$,
        \item[(ii)] $\mathbb{Q}[\mathcal{R}] \le \mathbb{Q}[\hat{\mathcal{R}}(\cdot, \notation{\realData})] 
        + \frac{1}{\natPar^{(1)}(\idx)} \left(\divergence{\rm{KL}}{\mathbb{Q}}{\mathbb{P}_{\notation{\rvHyp}}} + \log\left(\frac{\coveringNumber}{\varepsilon}\right) + \coveringConstant  - \mathbb{Q}\left[\sprod{\natPar^{(r)}(\idx)}{\suffStat^{(r)}(\cdot, \notation{\realData})}\right] \right) \,.$ 
    \end{itemize}
    In particular, if Theorem~\ref{Thm_PACBayes_exponential_familiy} applies, we can replace (i) with (ii).
\end{Cor}

\noindent
Using similar rearrangements, the following example relates Theorem~\ref{Thm_PACBayes_exponential_familiy} to other known PAC-Bayesian bounds:
\begin{Exa}\label{Exa:relation_to_other_PAC_bounds}
    \begin{itemize}
        \item[(i)] Assume that the loss-function is bounded, that is, $0 \le \ell \le C$, and define $\idxset = \{\idx\}$, $b \equiv 1$, $\coveringConstant = 0$, $\suffStat(\realHyp, \realData) := \left( \mathcal{R}(\realHyp) - \hat{\mathcal{R}}(\realHyp, \notation{\realData}), \ C^2 \right)$, and $\natPar(\idx) := \left( \idx, -\frac{\idx^2}{8} \right)$. Then we recover Catoni's bound \citep{Catoni_2003, Alquier_2024}:
        \begin{align*}
        \prob \Bigl\{ &\forall \mathbb{Q} \in \probMeasures(\mathbb{P}_{\notation{\rvHyp}})  : \\ 
        &\mathbb{Q}[ \mathcal{R}] \le \mathbb{Q}[ \hat{\mathcal{R}}(\cdot, \realData)]\vert_{\notation{\realData = \rvData}} 
        + \frac{1}{\lambda} \left(
        \divergence{\rm{KL}}{\mathbb{Q}}{\mathbb{P}_{\notation{\rvHyp}}} 
        + \log\left(\frac{1}{\varepsilon}\right) 
        + \frac{\lambda^2 C^2}{8N}
        \right)\Bigr \} \ge 1-\varepsilon \,.
        \end{align*}
        \item[(ii)] Assume that $\ell$ takes values in $[0,1]$ and let $D: [0,1]^2 \to \R$ be convex. Further, define $\idxset = \{1\}$, $b \equiv 1$, $\coveringConstant = 0$, $\suffStat(\realHyp, \realData) := \left(n D(\mathcal{R}(\realHyp), \hat{\mathcal{R}}(\realHyp, \notation{\realData})), \ \log(\expectation_{(\rvData, \rvHyp)} [ \exp(n D(\mathcal{R}, \hat{\mathcal{R}})) ] ) \right)$, and $\natPar(\idx) := \left( 1, -1 \right)$.
        Then we get:
        \begin{align*}
        \prob \Bigl\{ &\forall \mathbb{Q} \in \probMeasures(\mathbb{P}_{\notation{\rvHyp}})  :  
        \mathbb{Q}[ D(\mathcal{R}, \hat{\mathcal{R}}(\cdot, \realData))] \vert_{\notation{\realData = \rvData}} \\
        &\le \frac{1}{n}
        \left(\divergence{\rm{KL}}{\mathbb{Q}}{\mathbb{P}_{\notation{\rvHyp}}} 
        + \log\left(\frac{1}{\varepsilon}\right) 
        + \log(\expectation_{(\rvData, \rvHyp)} [ \exp(n D(\mathcal{R}, \hat{\mathcal{R}})) ] )
        \right)\Bigr \} \ge 1-\varepsilon \,.
        \end{align*}
        Applying Jensen's inequality to the left term, we get the bound of \citet{GLL2009}. Similarly, one can obtain the bound of \citet{Begin_Germain_Laviolette_Roy_2014}.
        \item[(iii)] Consider two measurable functions $f, g: \hypSpace \times \parSpace^N \to \R$, and define $\idxset = \{\idx\}$, $b \equiv 1$, $\coveringConstant = 0$, $\suffStat(\realHyp, \realData) := (f(\realHyp, \realData), g(\realHyp, \realData))$, and $\natPar(\idx) := (\idx, -\idx)$.
        Then our assumption $\mathbb{E}_{\notation{\rvData}}[c(\idx, \cdot)] \le 1$ reads
        $\expectation_{(\rvData, \rvHyp)} \left [ \exp(\idx(f - g)) \right] \le 1$,
        which is the same assumption as in \citet[Thm. 5.1]{Hellström_Durisi_Guedj_Raginsky_2023}. Similarly, defining $\natPar(\idx) := (1, -1)$ and $T(\realHyp, \realData) := (f(\realHyp, \realData), \ \log(\expectation_{(\rvData, \rvHyp)} [ \exp(f) ] ))$,
        we also get a similar bound as \citet[Proposition 5.2]{Hellström_Durisi_Guedj_Raginsky_2023}:
        \begin{align*}
        \prob \Bigl\{ &\forall \mathbb{Q} \in \probMeasures(\mathbb{P}_{\notation{\rvHyp}})  :  
        \mathbb{Q}[ f(\cdot, \realData)] \vert_{\notation{\realData = \rvData}} \\
        &\le \divergence{\rm{KL}}{\mathbb{Q}}{\mathbb{P}_{\notation{\rvHyp}}} 
        + \log\left(\frac{1}{\varepsilon}\right) 
        + \log(\expectation_{(\rvData, \rvHyp)} [ \exp(f) ])
        \Bigr \} \ge 1-\varepsilon \,.
        \end{align*}
    \end{itemize}
\end{Exa}

\section{Learning-to-Optimize with Guarantees}\label{Subsec:learning_with_guarantees}

Here, for our setting in Subsection~\ref{subsec:main_assumptions}, we consider properties of optimization algorithms that assert the necessary condition of Theorem~\ref{Thm_PACBayes_exponential_familiy}, namely $\mathbb{E}[c(\idx, \notation{\rvData})] \le 1$ for all $\idx \in \idxset$, with $c$ defined as in Equation~\ref{Eq_DefKappa}, to employ the PAC-Bayesian bound from Section \ref{Sec_PAC}.

\subsection{Worst-Case Bounds}\label{Sec_EnsuredConvergence}
In the next theorem, the additional assumption on $\mathcal{A}$ is sufficient to ensure the conditions of Theorem \ref{Thm_PACBayes_exponential_familiy}. Essentially, it requires the loss of the algorithm's output to be bounded. It can be used, for example, if one wants to combine the learning procedure with existing worst-case guarantees. Yet, as shown in Section~\ref{SubSec_CondOnConv}, it is too restrictive to achieve a significant acceleration compared to the standard choices from a worst-case analysis. For this, please recall our short-hand notation $\ell(\notation{\realHyp}, \notation{\realPar}) = \ell(\mathcal{A}(\notation{\realHyp}, \notation{\realPar}), \notation{\realPar}) = \ell(\mathcal{A}(\notation{\realHyp}, \iter{x}{0}, \notation{\realPar}), \notation{\realPar})$, that is, $\ell(\iter{x}{0}, \notation{\realPar})$ evaluates the loss function at $\iter{x}{0}$, while $\ell(\notation{\realHyp}, \notation{\realPar})$ evaluates the loss function at the output of the algorithm with hyperparameters $h$ and starting from $\iter{x}{0}$.

\begin{Thm}\label{Thm_algo_with_convergence_property}
    Suppose that $\notation{\rvPar}$ and $\ell$ satisfy Assumption~\ref{Assumption_random_parametric_problem}, and suppose that $\mathcal{A}$ satisfies Assumption~\ref{Assumption_parametric_optimization_algorithm}. Further, assume that there is a measurable function $\rho: \notation{\hypSpace} \longrightarrow [0, \infty)$, such that for every $\notation{\realHyp} \in \notation{\hypSpace}$ it holds that $\ell(\notation{\realHyp}, \cdot) \le \rho(\notation{\realHyp}) \ell(\iter{x}{0}, \cdot)$ $\prob_{\notation{\rvPar}}$-a.s. Furthermore, let $\notation{\rvData}$ be a corresponding i.i.d. data set of size $N \in \N$. Finally, assume that $\mathbb{E}\bigl[ \ell(\iter{x}{0}, \notation{\rvPar})^2 \bigr] < \infty$, and define $\natPar: (0, \infty) \longrightarrow \mathbb{R}^2$ and $\suffStat: \notation{\hypSpace} \times \notation{\parSpace^N} \longrightarrow \mathbb{R}^2$ through:
    \begin{align*}
        \natPar(\idx) := \left( \idx, \ -\frac{\idx^2}{2} \right)\,, \ \ \  
        \suffStat (\notation{\realHyp, \realData}) := \left(\mathcal{R}(\notation{\realHyp}) - \hat{\mathcal{R}}(\notation{\realHyp, \realData}), \  \frac{\rho^2(\notation{\realHyp})}{N}\mathbb{E}\left[ \ell(\iter{x}{0}, \notation{\rvPar})^2 \right] \right)\,.
    \end{align*}
    Then it holds that $\mathbb{E}[c(\idx, \notation{\rvData})] \le 1$ for all $\idx > 0$.
\end{Thm}
\begin{proof}
Since $\notation{\rvHyp}$ and $\notation{\rvData}$ are independent, their joint distribution is given by the product measure $\prob_{\notation{\rvData}} \otimes \prob_{\notation{\rvHyp}}$. Thus, by Fubini's theorem we get: 
$$
    \mathbb{E} \left[ \exp\left( \idx (\mathcal{R}(\notation{\rvHyp}) - \hat{\mathcal{R}}(\notation{\rvHyp, \rvData})) \right)\right] 
    = \expectation \left[ \mathbb{E} \left[ \exp\left( \idx (\mathcal{R}(\notation{\realHyp}) - \hat{\mathcal{R}}(\notation{\realHyp, \rvData})) \right)\right]\Bigl\vert_{\notation{\realHyp = \rvHyp}} \right ] \,.
$$
Hence, first consider the inner integral for a fixed $\notation{\realHyp} \in \notation{\hypSpace}$. Then, by definition and the i.i.d. assumption one gets:
\begin{align*}
    \mathbb{E} \left[ \exp\left( \idx (\mathcal{R}(\notation{\realHyp}) - \hat{\mathcal{R}}(\notation{\realHyp, \rvData})) \right)\right] 
    &= \mathbb{E} \left [ \exp\left( 
     -\frac{\idx}{N} \sum_{i=1}^N \left(\ell(\notation{\realHyp, \rvPar_i}) - \mathbb{E}[\ell(\notation{\realHyp, \rvPar})]  \right) \right) \right ]  \\
    &=\prod_{i=1}^N \mathbb{E}_{\notation{\rvPar}} 
     \left [ \exp\left( 
     -\frac{\idx}{N} \left(\ell(\notation{\realHyp}, \cdot) - \mathbb{E}_{\notation{\rvPar}}[\ell(\notation{\realHyp}, \cdot)] \right) \right) \right ] \,.
\end{align*}
The loss-function is non-negative and, by assumption on $\mathcal{A}$, can be bounded $\prob_{\notation{\rvPar}}$-a.s. Thus, for every $\notation{\realHyp} \in \notation{\hypSpace}$, $\ell(\notation{\realHyp, \rvPar})$ is a non-negative random variable with finite second-moment, as $\mathbb{E}_{\notation{\rvPar}}[\ell(\notation{\realHyp}, \cdot)^2] \le \rho(\notation{\realHyp})^2 \mathbb{E}_{\notation{\rvPar}}\left[\ell(\iter{x}{0}, \cdot)^2\right] < \infty$. Hence, by Lemma~\ref{Lemma_subgaussian_lower_tail}, we get:  
\begin{align*}
    \mathbb{E}_{\notation{\rvPar}} \left [ \exp\left(-\frac{\idx}{N} \left(\ell(\notation{\realHyp}, \cdot) - \mathbb{E}_{\notation{\rvPar}}[\ell(\notation{\realHyp}, \cdot)] \right) \right) \right ] 
    &\le \exp\Bigl(\frac{\idx^2}{2N^2} \mathbb{E}_{\notation{\rvPar}} [ \ell(\notation{\realHyp}, \cdot)^2 ] \Bigr) \\
    \le \exp\Bigl( \frac{\idx^2}{2N^2} \rho(\notation{\realHyp})^2 \mathbb{E}_{\notation{\rvPar}} [\ell(\iter{x}{0}, \cdot)^2] \Bigr) \,.
\end{align*} 
Therefore we have the following bound:
\begin{align*}
    \mathbb{E} \left[ \exp\left( \idx (\mathcal{R}(\notation{\realHyp}) - \hat{\mathcal{R}}(\notation{\realHyp, \rvData})) \right)\right] 
    \le \exp\Bigl( \frac{\idx^2}{2N} \rho(\notation{\realHyp})^2 \mathbb{E}_{\notation{\rvPar}} \left[ \ell(\iter{x}{0}, \cdot)^2 \right] \Bigr) \,.
\end{align*}
This can be rearranged into $\mathbb{E} \left [ \exp\left( \idx (\mathcal{R}(\notation{\realHyp}) - \hat{\mathcal{R}}(\notation{\realHyp, \rvData})) - \frac{\idx^2}{2N} \rho(\notation{\realHyp})^2 \mathbb{E}_{\notation{\rvPar}} [\ell(\iter{x}{0}, \cdot)^2] \right) \right ] \le 1$, as the right-hand side does not depend on $\notation{\rvData}$. 
Since $\notation{\rvHyp}$ and $\notation{\rvData}$ are independent, and $\notation{\realHyp} \in \notation{\hypSpace}$ was arbitrary, this inequality does hold $\prob_{\notation{\rvHyp}}$-a.s. Therefore, one directly gets the bound
$\mathbb{E} \left [ \exp\left( \idx (\mathcal{R}(\notation{\rvHyp}) - \hat{\mathcal{R}}(\notation{\rvHyp, \rvData})) - \frac{\idx^2}{2N} \rho(\notation{\rvHyp})^2 \mathbb{E}_{\notation{\rvPar}} [\ell(\iter{x}{0}, \cdot)^2] \right) \right ] \le 1$.
Now, again by Fubini's theorem, one can also switch the order of integration to get:
$$
    \mathbb{E} \left [ \mathbb{E} \left[ \exp\left( \idx (\mathcal{R}(\notation{\rvHyp}) - \hat{\mathcal{R}}(\notation{\rvHyp, \realData})) - \frac{\idx^2}{2N} \rho(\notation{\rvHyp})^2 \mathbb{E}_{\notation{\rvPar}} [\ell(\iter{x}{0}, \cdot)^2] \right) \right]\Bigl\vert_{\notation{\realData = \rvData}} \right ] \le 1 \,.
$$
Inserting the definition of $\natPar$ and $\suffStat$ gives $\mathbb{E} \left [ \mathbb{E} \left[ \exp\left( \sprod{\natPar(\idx)}{\suffStat(\notation{\rvHyp, \realData})} \right) \right] \vert_{\notation{\realData = \rvData}} \right] \le 1$. Here, the inner term is the same as
$\mathbb{E} \left[ \exp\left( \sprod{\natPar(\idx)}{\suffStat(\notation{\rvHyp, \realData})} \right) \right] = \int_{\notation{\hypSpace}} \exp\left( \sprod{\natPar(\idx)}{\suffStat(\notation{\realHyp, \realData}))} \right) \ \prob_{\notation{\rvHyp}}(d\notation{\realHyp})= c(\idx, \notation{\realData})$. Hence, this is the same as $\mathbb{E}[c(\idx, \notation{\rvData})] \le 1$.
\end{proof}
\begin{Rem}
    The argument still works for a data-dependent prior, if the corresponding data sets $\notation{\rvData'}$ and $\notation{\rvData}$ are independent: While interchanging the integration w.r.t. $\notation{\rvData'}$ and $\notation{\rvHyp}$ is not allowed, an interchange w.r.t. $\notation{\rvHyp}$ and $\notation{\rvData}$ is still valid (under the integral), that is, for a function $f$ it would hold
    $\expectation \left[ f(\notation{\rvHyp, \rvData, \rvData'}) \right] 
    = \expectation_{\notation{\rvData'}} \left [ \expectation_{\notation{\rvHyp \vert \rvData' = \realData'}} \left[ \expectation_{\notation{\rvData}} \left[ f(\notation{\realHyp}, \cdot, \notation{\realData'}) \right]\vert_{\notation{\realHyp = \rvHyp}} \right] \right ]
    = \expectation_{\notation{\rvData'}} \left [  \expectation_{\notation{\rvData}} \left[ \expectation_{\notation{\rvHyp \vert \rvData' = \realData'}} \left[ f(\cdot, \notation{\realData, \realData'}) \right]\vert_{\notation{\realData = \rvData}} \right] \right ]$, and the inner term is $\le 1$ in any case.
\end{Rem}

\subsection{Conditional Boundedness}\label{SubSec_CondOnConv}
Typically, the previous approach is too restrictive, because the boundedness assumption on $\mathcal{A}$ already requires theoretical worst-case estimates \emph{almost surely}. For example, if $(\ell(\cdot, \notation{\realPar}))_{\notation{\realPar \in \parSpace}}$ is a family of quadratic functions, and one tries to learn the step-size of gradient descent, the boundedness prevents step-size parameters that lie outside the worst-case convergence regime, as they would lead to a diverging behaviour, which increases the incurred empirical risk dramatically. Thus, to motivate the upcoming discussion, consider the following thought-experiment:
\begin{Exa}
    Consider $\ell(x, \notation{\realPar}) := \frac{\notation{\realPar}}{2} x^2$ and assume that the chosen algorithm is gradient descent, that is $\iter{x}{k+1} = \iter{x}{k} - \realHyp \ell'(\iter{x}{k}, \notation{\realPar})$. For a given $\notation{\realPar}$, the optimal step-size is $\realHyp = \frac{1}{\notation{\realPar}}$, which gives convergence in one step. Then, if $\notation{\realPar}$ is given by samples from the distribution $\prob_{\notation{\rvPar}} = 0.99 \delta_1 + 0.01 \delta_{100}$, a worst-case analysis would suggest to take $\realHyp_w = \frac{1}{100}$. In this case, we would have an algorithm that converges in a single step for 1\% of the problem instances, while having a linear convergence rate of $(\frac{99}{100})^k$ for the other 99\%. Another choice is to take $\realHyp_d = 1$, which leads to an algorithm that does converge in a single step for 99\% of the problem instances, but diverges in 1\% of the cases. By restricting to the 99\% of the cases where convergence does occur, the overall difference in speed is drastic.
\end{Exa}
Hence, in this section, a different approach is taken: We actually allow for divergence, if it only occurs in rare cases with a controllable probability, that is, \say{almost surely} is relaxed to \say{with a sufficiently large probability}. Essentially, we only consider the loss for all those hyperparameters, where the loss is bounded by a certain constant, as well as the probability for that to occur. Then, in Section~\ref{SubSec_GuaranteeConvProb}, we develop a technique that allows the user to actually control this probability. Clearly, a stronger guarantee trades for convergence speed. 

\begin{Def}\label{Def:convergence_set}
    Given a measurable function $\sublevelfunction: \notation{\parSpace} \to \R$, the (parametric) \emph{sublevel set} $\sublevel{\sublevelfunction} \subset \notation{\hypSpace} \times \notation{\parSpace}$ is defined as $\sublevel{\sublevelfunction} := \{ (\notation{\realHyp, \realPar}) \in \notation{\hypSpace} \times \notation{\parSpace} \ : \ \ell(\notation{\realHyp, \realPar}) \le \sublevelfunction(\notation{\realPar})\}$. The sections of $\sublevel{\sublevelfunction}$ for fixed $\notation{\realHyp} \in \notation{\hypSpace}$ will be denoted by $\sublevel{\sublevelfunction, \notation{\realHyp}}$.
\end{Def}

\noindent In Lemma~\ref{Lem:convergence_probability} we show that $\sublevel{\sublevelfunction}$ is indeed a measurable set. This is not obvious, as the loss function and the algorithm are composed in a non-standard way. This result further implies that the sections $\sublevel{\sublevelfunction, \notation{\realHyp}}$ are measurable, too. Since $\notation{\hypSpace}$ and $\notation{\parSpace}$ are Polish spaces, the product $\notation{\hypSpace} \times \notation{\parSpace}$ is again Polish. Hence, there exists a regular version of the conditional probability of $\notation{\rvPar}$, given $\notation{\rvHyp}$, that is, a kernel $\notation{\hypSpace} \to \notation{\parSpace}, \ (\notation{\realHyp}, \set{B}) \mapsto \prob_{\notation{\rvPar \vert \rvHyp = \realHyp}} [\set{B}]$. By \citet[Thm. 1.122, p.124]{Witting_2013}, this determines a regular version of the conditional probability of $(\notation{\rvHyp, \rvPar})$, given $\notation{\rvHyp}$, through $\notation{\hypSpace} \to \notation{\hypSpace} \times \notation{\parSpace}, \ (\notation{\realHyp}, \set{B}) \mapsto \prob_{(\notation{\rvHyp, \rvPar}) \vert \notation{\rvHyp = \realHyp}}[\set{B}] := \prob_{\notation{\rvPar \vert \rvHyp = \realHyp}} [\set{B}_{\notation{\realHyp}}]$, and we have $\prob_{\notation{\rvHyp}}$-a.s. the equality $\prob \{(\notation{\rvHyp, \rvPar}) \in \set{B} \ \vert \ \notation{\rvHyp = \realHyp} \} = \prob_{\notation{\rvPar \vert \rvHyp = \realHyp}} [\set{B}_{\notation{\realHyp}}]$. In particular, this applies to the sublevel set $\sublevel{\sublevelfunction}$, and the map $\notation{\realHyp} \mapsto \prob_{\notation{\rvPar \vert \rvHyp = \realHyp}} [\sublevel{\sublevelfunction, \notation{\realHyp}}]$ is measurable.

\begin{Def}
    Let $\sublevel{\sublevelfunction}$ be a parametric sublevel set. Define the \emph{sublevel probability} as the measurable function $\notation{\realHyp} \mapsto \sublevelprob(\notation{\realHyp}) := \prob_{\notation{\rvPar \vert \rvHyp = \realHyp}}[\sublevel{\sublevelfunction, \notation{\realHyp}}]$.
\end{Def}

\noindent
This construction allows us to give a more fine-grained analysis of the algorithm, as it allows to trade the boundedness assumption for the sublevel probability. This basically extends a worst-case analysis, which would correspond to an uniform upper bound. Motivated by Lemma~\ref{Lem:Sublevel_Risk}, we define the \emph{sublevel risk} as the expect loss \emph{conditioned} on the sublevel set:
\begin{Def}
    Let $\sublevel{\sublevelfunction}$ be a parametric sublevel set. Then the \emph{sublevel risk} $\mathcal{R}_{\sublevelfunction}: \notation{\hypSpace} \to [0, +\infty]$ is defined as the conditional expectation of the loss given $\sublevel{\sublevelfunction, \notation{\realHyp}}$:
    $$
        \notation{\realHyp} \mapsto 
        \sublevelrisk{\sublevelfunction}(\notation{\realHyp}) 
        := \mathbb{E}_{\notation{\rvPar}}\left[\ell(\notation{\realHyp}, \cdot) \ \vert \ \sublevel{\sublevelfunction, \notation{\realHyp}} \right] 
        = \begin{cases}
            \frac{1}{\sublevelprob(\notation{\realHyp})} \mathbb{E}_{\notation{\rvPar}}[\ell(\notation{\realHyp}, \cdot) \mathds{1}_{\sublevel{\sublevelfunction, \notation{\realHyp}}}], &\text{if $\sublevelprob(\notation{\realHyp}) > 0$} \,; \\
            0, &\text{otherwise} \,.
        \end{cases}
    $$
    Given a data set $\notation{\rvData} = (\notation{\rvPar_1, ..., \rvPar_N})$, the \emph{empirical sublevel risk} $\empsublevelrisk{\sublevelfunction}: \notation{\hypSpace} \times \notation{\parSpace^N} \to [0, +\infty]$ is defined as
    $(\notation{\realHyp, \rvData}) \mapsto \empsublevelrisk{\sublevelfunction}(\notation{\realHyp, \rvData}) := \frac{1}{\sublevelprob(\notation{\realHyp})} \frac{1}{N} \sum_{i=1}^N \mathds{1}_{\sublevel{\sublevelfunction, \notation{\realHyp}}}(\notation{\rvPar_i}) \ell(\notation{\realHyp, \rvPar_i})$. 
\end{Def}
\noindent
The following theorem is a direct generalization of Theorem \ref{Thm_algo_with_convergence_property}. Especially, note that the additional assumption on $\mathcal{A}$ is not needed anymore.

\begin{Thm} \label{Thm_PacBayesConditioned}
    Suppose that $\notation{\rvPar}$ and $\ell$ satisfy Assumption~\ref{Assumption_random_parametric_problem}, and suppose that $\mathcal{A}$ satisfies Assumption~\ref{Assumption_parametric_optimization_algorithm}. Further, let $\notation{\rvData}$ be a corresponding i.i.d. data set of size $N \in \N$, and let $\sublevel{\sublevelfunction}$ be a parametric sublevel set with sublevel probability $\sublevelprob$. Assume that $\prob_{\notation{\rvHyp}} \{ \sublevelprob > 0\} = 1$ and $\mathbb{E}_{\notation{\rvPar}}[\sublevelfunction^2] < \infty$. Define $\natPar: (0, \infty) \longrightarrow \mathbb{R}^2$ and $\suffStat: \notation{\hypSpace} \times \notation{\parSpace^N} \longrightarrow \mathbb{R}^2$ as 
    \begin{align*}
        \natPar(\idx) := \left( \idx, \ -\frac{\idx^2}{2} \right)\,, \ \ \
        \suffStat(\notation{\realHyp, \realData}) := \left( \sublevelrisk{\sublevelfunction} (\notation{\realHyp}) - \empsublevelrisk{\sublevelfunction}(\notation{\realHyp, \realData}), \ \frac{1}{\sublevelprob(\notation{\realHyp})^2 N }\mathbb{E}_{\notation{\rvPar}} \left[\sublevelfunction^2 \mathds{1}_{\sublevel{\sublevelfunction, \notation{\realHyp}}} \right] \right) \,.
    \end{align*}
    Then, for all $\idx > 0$, it holds that $\mathbb{E}[c(\idx, \notation{\rvData})] \le 1$.
\end{Thm}
\begin{proof}
    The proof is very similar to the proof of Theorem~\ref{Thm_algo_with_convergence_property} and basically uses the same reasoning. Let $\ell_{\sublevelfunction}(\notation{\realHyp, \realPar}) := \mathds{1}_{\sublevel{\sublevelfunction, \notation{\realHyp}}}(\notation{\realPar}) \ell (\notation{\realHyp, \realPar})$. Since $\notation{\rvHyp}$ and $\notation{\rvData}$ are independent, one gets from Fubini's theorem:
    $$
        \mathbb{E} \left[ \exp(\idx(\sublevelrisk{\sublevelfunction}(\notation{\rvHyp}) - \empsublevelrisk{\sublevelfunction}(\notation{\rvHyp, \rvData}))) \right] 
        = \expectation \left [ \mathbb{E} \left[ \exp(\idx(\sublevelrisk{\sublevelfunction}(\notation{\realHyp}) - \empsublevelrisk{\sublevelfunction}(\notation{\realHyp, \rvData}))) \right]\Bigl\vert_{\notation{\realHyp = \rvHyp}} \right]\,.
    $$
    Thus, first consider a fixed $\notation{\realHyp} \in \notation{\hypSpace}$ with $p(\notation{\realHyp}) > 0$. Then, by definition and the i.i.d. assumption, it holds that:
    \begin{align*}
        \mathbb{E} \left[ \exp(\idx(\sublevelrisk{\sublevelfunction}(\notation{\realHyp}) - \empsublevelrisk{\sublevelfunction}(\notation{\realHyp, \rvData}))) \right] 
        &= \mathbb{E} \left[ \exp\left(-\frac{\idx}{N \sublevelprob(\notation{\realHyp})} \sum_{i=1}^N \left(\ell_{\sublevelfunction}(\notation{\realHyp, \rvPar_i}) - \mathbb{E}_{\notation{\rvPar}}[\ell_{\sublevelfunction}(\notation{\realHyp}, \cdot)] \right)\right) \right] \\
        &= \prod_{i=1}^N \mathbb{E}_{\notation{\rvPar}} \left[  \exp\left(-\frac{\idx}{N \sublevelprob(\notation{\realHyp})} \left( \ell_{\sublevelfunction}(\notation{\realHyp}, \cdot) - \mathbb{E}_{\notation{\rvPar}}[\ell_{\sublevelfunction}(\notation{\realHyp}, \cdot)] \right)\right) \right ] \,.
    \end{align*}
    $\ell_{\sublevelfunction}(\notation{\realHyp}, \cdot)$ is non-negative, and by definition of the parametric sublevel set has a finite second-moment, that is $\mathbb{E}_{\notation{\rvPar}}[\ell_{\sublevelfunction}(\notation{\realHyp}, \cdot)^2] \le \mathbb{E}_{\notation{\rvPar}}[\sublevelfunction^2 \mathds{1}_{\sublevel{\sublevelfunction, \notation{\realHyp}}}] < \infty$. Hence, by Lemma~\ref{Lemma_subgaussian_lower_tail} we have the inequality $\mathbb{E}_{\notation{\rvPar}} \left [  \exp\left(-\frac{\idx}{N \sublevelprob(\notation{\realHyp})} \left( \ell_{\sublevelfunction}(\notation{\realHyp}, \cdot) - \mathbb{E}_{\notation{\rvPar}}[\ell_{\sublevelfunction}(\notation{\realHyp}, \cdot)] \right)\right) \right ] \le \exp\left(\frac{\idx^2}{2 N^2 \sublevelprob(\notation{\realHyp})^2} \mathbb{E}\left[\ell_{\sublevelfunction}(\notation{\realHyp}, \cdot)^2\right]\right)$. Thus:
    $$
        \mathbb{E} \left[ \exp(\idx(\sublevelrisk{\sublevelfunction}(\notation{\realHyp}, \cdot) - \empsublevelrisk{\sublevelfunction}(\notation{\realHyp, \rvData}))) \right] \le 
        \exp\left(\frac{\idx^2}{2 N \sublevelprob(\notation{\realHyp})^2} \mathbb{E}_{\notation{\rvPar}}\left[\sublevelfunction^2 \mathds{1}_{\sublevel{\sublevelfunction, \notation{\realHyp}}}\right]\right) \,.
    $$
    This can be rearranged into $\mathbb{E} \left [ \exp\left(\idx(\sublevelrisk{\sublevelfunction}(\notation{\realHyp}) - \empsublevelrisk{\sublevelfunction}(\notation{\realHyp, \rvData})) - \frac{\idx^2}{2 N \sublevelprob(\notation{\realHyp})^2} \mathbb{E}_{\notation{\rvPar}}\left[\sublevelfunction^2 \mathds{1}_{\sublevel{\sublevelfunction, \notation{\realHyp}}}\right] \right) \right ] \le 1$, since the right-hand side is independent of $\notation{\rvData}$.
    As this holds for any $\notation{\realHyp}$ with $\sublevelprob(\notation{\realHyp}) > 0$, which in turn does hold $\prob_{\notation{\rvHyp}}$-a.s., we get 
    $$
    \mathbb{E} \left [ \exp\left(\idx(\sublevelrisk{\sublevelfunction}(\notation{\rvHyp}) - \empsublevelrisk{\sublevelfunction}(\notation{\rvHyp, \rvData})) - \frac{\idx^2}{2 N \sublevelprob(\notation{\rvHyp})^2} \mathbb{E}_{\notation{\rvPar}}\left[\sublevelfunction^2 \mathds{1}_{\sublevel{\sublevelfunction, \notation{\realHyp}}}\right]\vert_{\notation{\realHyp = \rvHyp}} \right) \right ] \le 1\,.$$
    Changing the order of integration with Fubini's theorem, we get: 
    $$
    \expectation \left [ \expectation \left[ \exp\left(\idx(\sublevelrisk{\sublevelfunction}(\notation{\rvHyp}) - \empsublevelrisk{\sublevelfunction}(\notation{\rvHyp, \realData})) - \frac{\idx^2}{2 N \sublevelprob(\notation{\rvHyp})^2} \mathbb{E}_{\notation{\rvPar}}\left[\sublevelfunction^2 \mathds{1}_{\sublevel{\sublevelfunction, \notation{\realHyp}}}\right]\vert_{\notation{\realHyp = \rvHyp}} \right) \right]\Bigl \vert_{\notation{\realData = \rvData}} \right ] \le 1\,.
    $$
    Using the definition of $\natPar$ and $\suffStat$, this is the same as
    $
        \mathbb{E} \left [ \mathbb{E} \left[ \exp(\sprod{\natPar(\idx)}{\suffStat(\notation{\rvHyp, \realData})}) \right]\vert_{\notation{\realData = \rvData}} \right ] \le 1 \,.
    $
    Thus, in total we get $\mathbb{E} \left[ c(\idx, \notation{\rvData}) \right] \le 1$, because the inner term can be rewritten as $\mathbb{E} \left[ \exp(\sprod{\natPar(\idx)}{\suffStat(\notation{\rvHyp, \realData})}) \right] = \int_{\notation{\hypSpace}} \exp(\sprod{\natPar(\idx)}{\suffStat(\notation{\realHyp, \realData})}) \ \prob_{\notation{\rvHyp}}(d\notation{\realHyp}) = c(\idx, \notation{\realData})$.
\end{proof}

\begin{Rem}
    The assumption $\mathbb{P}_{\notation{\rvHyp}}\{ \sublevelprob > 0 \} = 1$ states that, under the prior, the algorithm should be able to \say{reach} the sublevel set. This is a constraint on the \emph{support} of $\prob_{\notation{\rvHyp}}$, which is not satisfied without further ado. Section~\ref{Alg:Construction_Prior} provides a construction for achieving this.
\end{Rem}

\begin{Exa}\label{Exa:main_result_l2o}
    Combining Theorem~\ref{Thm_PacBayesConditioned} and Theorem~\ref{Thm_PACBayes_exponential_familiy}, we get that:
    \begin{align*}
        \prob \Bigl\{ &\forall \idx \in \idxset, \ \forall \mathbb{Q} \in \probMeasures(\mathbb{P}_{\notation{\rvHyp}})  :  
        \mathbb{Q}[ \sublevelrisk{\sublevelfunction} ] \le \mathbb{Q}[\empsublevelrisk{\sublevelfunction}(\notation{\realData})]\vert_{\notation{\realData = \rvData}} + \\
        &+ \frac{\divergence{\rm{KL}}{\mathbb{Q}}{\mathbb{P}_{\notation{\rvHyp}}} 
        + \log\left(\frac{\coveringNumber}{\varepsilon} \right) + \coveringConstant}{\idx} 
        + \frac{\idx}{2N} \mathbb{Q} \left[\frac{\expectation_{\rvPar} [\sublevelfunction^2 \mathds{1}_{\sublevel{\sublevelfunction, \realHyp}}]\vert_{\realHyp = \cdot}}{\sublevelprob(\cdot)^2} \right]  \Bigr \} \ge 1-\varepsilon \,.
    \end{align*}
    For every fixed $\mathbb{Q} \in \probMeasures(\prob_{\rvHyp})$, optimizing over $\idx$ (assuming that $\idx^*$ is attained in $\Gamma$), gives:
    \begin{align*}
        \prob \Bigl\{ &\forall \mathbb{Q} \in \probMeasures(\mathbb{P}_{\notation{\rvHyp}})  : 
        \mathbb{Q}[ \sublevelrisk{\sublevelfunction} ] \le \mathbb{Q}[\empsublevelrisk{\sublevelfunction}(\notation{\realData})]\vert_{\notation{\realData = \rvData}} + \\
        &+ \sqrt{
        \frac{2 \left( \divergence{\rm{KL}}{\mathbb{Q}}{\mathbb{P}_{\notation{\rvHyp}}} 
        + \log\left(\frac{\coveringNumber}{\varepsilon} \right) + \coveringConstant\right) \mathbb{Q} \left[\frac{\expectation_{\rvPar} [\sublevelfunction^2 \mathds{1}_{\sublevel{\sublevelfunction, \realHyp}}]\vert_{\realHyp = \cdot}}{\sublevelprob(\cdot)^2} \right]
        }{N} 
        }  \Bigr \} \ge 1-\varepsilon \,.
    \end{align*}
    Now, a typical performance-measure in optimization is \emph{complexity}, that is, how many iterations are needed to reach a loss smaller or equal to $\vartheta$. Thus, specifying $\sublevelfunction \equiv \vartheta$ and assuming that $\sublevelprob(\notation{\rvHyp}) \ge \sublevelprob_l$ a.s., this gives rise to:
    \begin{align*}
        \prob \Bigl\{ &\forall \mathbb{Q} \in \probMeasures(\mathbb{P}_{\notation{\rvHyp}})  :  
        \mathbb{Q}[ \sublevelrisk{\sublevelfunction} ] \le \mathbb{Q}[\empsublevelrisk{\sublevelfunction}(\notation{\realData})]\vert_{\notation{\realData = \rvData}} + \\
        &+ \frac{\vartheta}{\sublevelprob_l}\sqrt{
        \frac{2 \left(\divergence{\rm{KL}}{\mathbb{Q}}{\mathbb{P}_{\notation{\rvHyp}}} 
        + \log\left(\frac{\coveringNumber}{\varepsilon} \right) + \coveringConstant\right)  
        }{N}} 
        \Bigr \} \ge 1-\varepsilon \,.
    \end{align*}
\end{Exa}

\section{Implementing the Non-divergence -- Speed Trade-Off}\label{SubSec_GuaranteeConvProb}
In Subsection~\ref{SubSec_CondOnConv}, care has to be taken in the choice of the prior $\mathbb{P}_{\notation{\rvHyp}}$: Just minimizing the upper bound as much as possible can lead to a neglect of a high sublevel probability, that is, the algorithm is especially fast on a small subset of the parameters, while it diverges for the rest. This is due to the fact that the term $\frac{1}{\sublevelprob(\notation{\realHyp})}$ might not compensate for the smaller sublevel risk. Thus, if a certain sublevel probability $\varepsilon_{\rm{conv}} \in [0,1]$ has to be ensured, one has to enforce it. In the case of PAC-Bayesian learning with absolutely continuous distributions, it suffices to have this property for the prior:
\begin{Lem}\label{Lem:absolute_continuity_and_conv_prob}
    Let $\varepsilon_{\rm{conv}} \in [0,1]$ and assume that $\sublevelprob(\notation{\rvHyp}) \ge \varepsilon_{\rm{conv}}$ a.s. Then, for every $\mathbb{Q} \in \probMeasures(\prob_{\notation{\rvHyp}})$ we have $\mathbb{Q} \{ \sublevelprob < \varepsilon_{\rm{conv}} \} = 0$.
\end{Lem}
\begin{proof}
    By assumption we have $\prob_{\notation{\rvHyp}} \{\sublevelprob < \varepsilon_{\rm{conv}}\} = 0$. Thus, the result follows directly by definition of absolute continuity.
\end{proof}
Though the proof is trivial, this lemma has a very important consequence, which we want to stress: If one can guarantee that a required property is satisfied for the prior, it will be \emph{preserved} during the PAC-Bayesian learning process, that is, if the prior only puts mass on hyperparameters that ensure a certain sublevel probability, the posterior will do the same. How to enforce such constraints during construction of the prior is discussed next.

\subsection{Sampling under Probabilistic Constraints}\label{Sec:Sampling_under_probabilistic_constraints}
\begin{figure}[t!]
    \centering
    \begin{tikzpicture}[scale=1]

    \coordinate (y) at (0, 4);
    \coordinate (x) at (5, 0);
    \coordinate (C) at (4.75, 2.25);
    \draw (C) node {$\set{C}$};
    \draw[<->] (y) node[above] {$\genSpaceTwo$} -- (0,0) --  (x) node[right] {$\genSpaceOne$};

    \coordinate (x1) at (1.4, 0.5);
    \coordinate (x2) at (2.5, 0.7);
    \coordinate (x3) at (3.4, 0.5);
    \coordinate (x4) at (4.5, 1.5);
    \coordinate (x5) at (4.2, 2.9);
    \coordinate (x6) at (3.4, 3.2);
    \coordinate (x7) at (2.5, 2.5);
    \coordinate (x8) at (1.4, 3.3);
    \coordinate (x9) at (0.5, 0.75);

    \draw[thick] plot [smooth] coordinates {(x1) (x2) (x3) (x4) (x5) (x6) (x7) (x8) (x9) (x1)};
    
    \draw[thick, my_blue] (x1) -- (x8);
    \draw[thick, my_green] (x3) -- (x6);
    \draw[dashed] (x1) -- (1.4, 0);
    \draw[dashed] (x3) -- (3.4, 0);

    \draw (1.4, 0.05) -- (1.4, -0.05);
    \draw (1.4, -0.05) node[below] {$\genRealOne_1$};
    \draw (x8) node[above] {$\set{C}_{\genRealOne_1}$};

    \draw (3.4, 0.05) -- (3.4, -0.05);
    \draw (3.4, -0.05) node[below] {$\genRealOne_2$};
    \draw (x6) node[above] {$\set{C}_{\genRealOne_2}$};
    
     \begin{scope}[xshift=8cm]
        \coordinate (y2) at (0,4);
        \coordinate (x2) at (5,0);
        \draw[axis] (y2) node[above] {$\prob_{\notation{\genRvTwo}}[\set{C}_{\genRealOne}]$} -- (0,0) --  (x2) node[right] {$\genSpaceOne$};

        \draw (0, 1.5) -- (5, 1.5);
        \draw (-0.05, 1.5) -- (0.05, 1.5);
        \draw (-0.05, 1.5) node[left] {$\sublevelprob_l$};
        \draw (0, 2.7) -- (5, 2.7);
        \draw (-0.05, 2.7) -- (0.05, 2.7);
        \draw (-0.05, 2.7) node[left] {$\sublevelprob_u$};

        \draw (-0.05, 3.5) -- (0.05, 3.5);
        \draw (-0.05, 3.5) node[left] {$1$};
        \draw (-0.05, 0) -- (0.05, 0);
        \draw (-0.05, 0) node[left] {$0$};

        \coordinate (x1) at (0, 0.05);
        \coordinate (x2) at (0.5, 0.1);
        \coordinate (x3) at (1.0, 0.5);
        \coordinate (x4) at (1.5, 1.5);
        \coordinate (x5) at (2.0, 2.7);
        \coordinate (x6) at (2.4, 3.2);
        \coordinate (x6h) at (2.75, 2.7);
        \coordinate (x7) at (3.0, 2.4);
        \coordinate (x8) at (3.5, 2.65);
        \coordinate (x9) at (4.5, 1.5);
        \coordinate (x10) at (5.0, 0.9);

        \draw plot [smooth] coordinates {(x1) (x2) (x3) (x4) (x5) (x6) (x6h) (x7) (x8) (x9) (x10)};
        \draw[thick, light_blue] plot [smooth] coordinates {(x4) (x5)};

        \draw[dashed] (x4) -- (1.5, 0);
        \draw (1.5, 0.05) -- (1.5, -0.05);
        \draw[dashed] (x5) -- (2.0, 0);
        \draw (2.0, 0.05) -- (2.0, -0.05);
        \draw[thick, light_pink] (1.5, 0.0) -- (2.0, 0.0);
        
        \draw[dashed] (x6h) -- (2.75, 0);
        \draw (2.75, 0.05) -- (2.75, -0.05);
        \draw[dashed] (x9) -- (4.5, 0);
        \draw (4.5, 0.05) -- (4.5, -0.05);
        \draw[thick, light_pink] (2.75, 0.0) -- (4.5, 0.0);
        
        \draw[thick, light_blue] plot [smooth] coordinates {(x6h) (x7) (x8) (x9)};

        \coordinate (supp) at (3.0, -0.25);
        \draw (supp) node[below] {\tiny{$\mathrm{supp}(\Tilde{\prob}_{\notation{\genRvOne}})$}};
      \end{scope}
     
\end{tikzpicture}
    \caption{Construction of $\Tilde{\prob}_{\notation{\genRvOne}}$: On the left, the set $\set{C} \subset \genSpaceOne \times \genSpaceTwo$ and two of its sections $\set{C}_{\genRealOne_1}, \set{C}_{\genRealOne_2} \subset \genSpaceTwo$ are visualized. On the right, the function $\sublevelprob(\genRealOne) = \prob_{\notation{\genRvTwo}}[\set{C}_{\genRealOne}]$, the interval $[\sublevelprob_l, \sublevelprob_u]$, and the resulting support $\mathrm{supp}(\Tilde{\prob}_{\notation{\genRvOne}})$ of $\Tilde{\prob}_{\notation{\genRvOne}}$ are visualized. Note that, contrary to the visualization here, $\sublevelprob$ can actually be \emph{highly discontinuous}.}
    \label{fig:Construction_measure_with_support}
\end{figure}
In this section, we describe a methodology that allows for sampling from a distribution that is \emph{probabilistically constrained} in the following sense: We are given two independent random variables $\notation{\genRvOne}: \pspace \to \genSpaceOne$, $\notation{\genRvTwo}: \pspace \to \genSpaceTwo$ taking values in the Polish spaces $\genSpaceOne$ and $\genSpaceTwo$, with joint and marginal distributions $\prob_{(\notation{\genRvOne, \genRvTwo})}$, $\prob_{\notation{\genRvOne}}$ and $\prob_{\notation{\genRvTwo}}$, respectively. Further, we consider a measurable set $\set{C} \subset \genSpaceOne \times \genSpaceTwo$, and we want to generate samples $\notation{\genRvOne = \genRealOne} \in \genSpaceOne$, such that the probability of $(\notation{\genRvOne, \genRvTwo})$ lying in $\set{C}$, given $\notation{\genRvOne = \genRealOne}$, takes values in a certain interval:
$$
    \prob_{(\notation{\genRvOne, \genRvTwo}) \vert \notation{\genRvOne = \genRealOne}}[\set{C}] = \prob_{\notation{\genRvTwo \vert \genRvOne = \genRealOne}}[\set{C}_{\notation{\genRealOne}}] \in [\sublevelprob_l, \sublevelprob_u] \subset [0,1] \,.
$$
This allows us to define the (measurable) function $\sublevelprob: \genSpaceOne \to [0,1]$, $u \mapsto \prob_{\notation{\genRvTwo \vert \genRvOne = \genRealOne}}[\set{C}_{\notation{\genRealOne}}]$. By independence of $\notation{\genRvOne}$ and $\notation{\genRvTwo}$, this is $\prob_{\notation{\genRvTwo}}$-almost surely the same as $\sublevelprob(\genRealOne) = \prob_{\notation{\genRvTwo}} [\set{C}_{\notation{\genRealOne}}] \in [\sublevelprob_l, \sublevelprob_u]$, and we will use the later formulation from now on.\footnote{Note that $\sublevelprob$ is still measurable.}
Thus, for $\sublevelprob_l, \sublevelprob_u \in [0,1]$ with $\sublevelprob_l < \sublevelprob_u$, we can define a measurable set $\set{A} := \bigl \{\genRealOne \in \genSpaceOne \ : \ \prob_{\notation{\genRvTwo}}[\set{C}_{\genRealOne}] \in [\sublevelprob_l, \sublevelprob_u] \bigr\}$, which yields a new measure $\Tilde{\prob}_{\notation{\genRvOne}}$ on $\genSpaceOne$ by restricting to $\set{A}$, that is, for a measurable set $\set{B} \subset \genSpaceOne$ it holds:
$$
\Tilde{\prob}_{\notation{\genRvOne}}[\set{B}] := \left((\mathds{1}_{[\sublevelprob_l, \sublevelprob_u]} \circ \sublevelprob) \cdot \prob_{\notation{\genRvOne}} \right) [\set{B}] = \left(\mathds{1}_\set{A} \cdot \prob_{\notation{\genRvOne}}\right) [\set{B}] = \prob_{\notation{\genRvOne}} [\set{A} \cap \set{B}] \,.
$$
Therefore, as stated before, we have the following goal:
$$
    \text{Goal: Sample from $\Tilde{\prob}_{\notation{\genRvOne}}$, that is, get $\notation{\genRvOne_1, ..., \genRvOne_K} \sim \prob_{\notation{\genRvOne}}$, such that $\prob_{\notation{\genRvTwo}}[\set{C}_{\notation{\genRealOne}}]\vert_{\notation{\genRealOne = \genRvOne_i}} \in [\sublevelprob_l, \sublevelprob_u]$.}
$$
This construction is depicted in Figure~\ref{fig:Construction_measure_with_support}: The left figure visualizes the sections $\{\set{C}_{\genRealOne}\}_{\genRealOne \in \genSpaceOne}$ of the set $\set{C}$, while the right figure shows the corresponding construction of the support of $\Tilde{\prob}_{\notation{\genRvOne}}$. In the following, we implicitly assume that the imposed constraint is realizable, that is, $\Tilde{\prob}_{\notation{\genRvOne}}$ has a non-empty support.

\begin{Exa}
    Consider the random variables $\notation{\rvPar}$ and $\notation{\rvHyp}$ from Section~\ref{Subsec:learning_with_guarantees}. 
    By Lemma~\ref{Lem:absolute_continuity_and_conv_prob} we want to have $\sublevelprob(\notation{\rvHyp}) \in [\varepsilon_{\rm{conv}}, 1]$, where the sublevel probability is given as $\sublevelprob(\notation{\realHyp}) = \prob_{\notation{\rvPar}}[\set{L}_{\sublevelfunction, \notation{\realHyp}}]$ (Lemma~\ref{Lem:Sublevel_Risk}), and the sublevel set $\set{L}_{\sublevelfunction} \subset \notation{\hypSpace} \times \notation{\parSpace}$ is measurable by Lemma~\ref{Lem:convergence_probability}. Thus, this corresponds to the identification $\genSpaceOne = \notation{\hypSpace}$, $\genSpaceTwo = \notation{\parSpace}$, and $\sublevelprob_l = \varepsilon_{\rm{conv}}$, $\sublevelprob_u = 1$.
\end{Exa}

\subsubsection{Incorporation into a Sampling Procedure}
\begin{algorithm}[t!]
	\caption{Iterative estimation of the probability $\sublevelprob$}\label{Alg:Estimation_Probabilistic_Constraint}
	\begin{algorithmic}
		\Require $\quantile_l, \quantile_u, \varepsilon \in [0, 1]$.
		\State $a, b \gets 1, 1$    \Comment{Initialize with uninformative prior.}
		\While{$Q_{a,b}(\quantile_u) - Q_{a,b}(\quantile_l) \ge \varepsilon$} \Comment{$Q_{a,b}$ is the quantile function for $\mathrm{Beta}(a,b)$.}
		\State Draw $\notation{\rv{I}} \sim \mathrm{Ber}(\sublevelprob)$
		\State $a \gets a + \notation{\rv{I}}$ and $b \gets b + (1 - \notation{\rv{I}})$
		\EndWhile
	\end{algorithmic}
\end{algorithm}
The only distinction between samples from $\Tilde{\prob}_{\notation{\genRvOne}}$ and samples from $\prob_{\notation{\genRvOne}}$ is the restriction to $\set{A}$. Since many sampling algorithms access the unnormalised density anyway, it suffices to be able to sample from $\prob_{\notation{\genRvOne}}$, if the restriction to $\set{A}$ can be satisfied differently.
Thus, we have to integrate this constraint into a sampling procedure for $\prob_{\notation{\genRvOne}}$. Because we do not have any geometrical or topological information about the set $\set{C}$, we resort to statistical information: Given i.i.d. samples $\notation{\genRvTwo_1, ..., \genRvTwo_n} \sim \prob_{\notation{\genRvTwo}}$, for a given $\genRealOne \in \genSpaceOne$, we are able to evaluate the Bernoulli random variables $\notation{\rv{I}_n}  := \mathds{1}\{\notation{\genRvTwo_n} \in \set{C}_{\genRealOne}\}$, $n \in \N$. These have the parameter $\prob \{\notation{\rv{I}_n} = 1\} = \prob \{\notation{\genRvTwo_n} \in \set{C}_{\genRealOne}\} = \prob_{\notation{\genRvTwo}} [\set{C}_{\genRealOne}] = \sublevelprob(\genRealOne)$. Thus, by estimating $\sublevelprob(\genRealOne)$ with an estimator $\hat{\sublevelprob}(\genRealOne)$, we approximate the constraint $\set{A}$ with $\hat{\set{A}}$:
$$
    \set{A} = \{\genRealOne \in \genSpaceOne \ : \ \sublevelprob(\genRealOne) \in [\sublevelprob_l, \sublevelprob_u]\} \approx \{\genRealOne \in \genSpaceOne \ : \ \hat{\sublevelprob}(\genRealOne) \in [\sublevelprob_l, \sublevelprob_u]\} =: \hat{\set{A}}\,.
$$
To decide whether a given sample $\notation{\genRvOne_i} \sim \prob_{\genRvOne}$ does lie in $\set{A}$, that is, whether $\notation{\genRvOne_i}$ can actually be regarded as a sample from $\Tilde{\prob}_{\notation{\genRvOne}}$,  we resort to a simple accept-reject mechanism as in \emph{Metropolis-Hastings}-type algorithms \citep{Robert_Casella_2004}. Note that this allows to keep an algorithm \emph{inside} $\hat{\set{A}}$. However, it does not provide a way \emph{into} $\hat{\set{A}}$, let alone $\set{A}$.

\begin{figure}[t!]
    \centering
    \includegraphics[width=\textwidth]{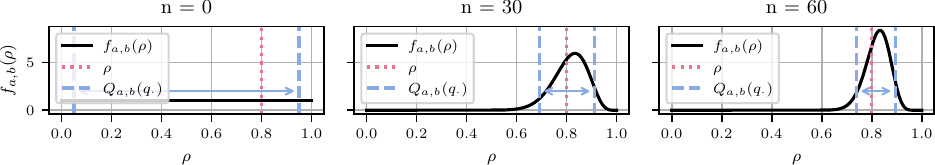}
    \caption{Iterative estimation of $\sublevelprob(\genRealOne)$: The black line shows the density $f_{a^{(n)}, b^{(n)}}$ of $\mathrm{Beta}(a^{(n)}, b^{(n)})$ after having observed $\notation{\rv{I}_1, ..., \rv{I}_n}$. The red dotted line indicates the true probability $\sublevelprob(\genRealOne)$, which we are trying to estimate, and the blue dashed lines indicate the lower and upper quantiles corresponding to $\quantile_l, \quantile_u$. The procedure stops as soon as $Q_{a^{(n)}, b^{(n)}}(\quantile_u) - Q_{a^{(n)}, b^{(n)}}(\quantile_l) < \varepsilon$, which is indicated by the double-headed arrow. Here, we use $\quantile_l = 0.05, \quantile_u = 0.95, \varepsilon = 0.15$ and $\sublevelprob(\genRealOne) = 0.8$.}
    \label{fig:estimation_convergence_probability}
\end{figure}

\noindent
We estimate $\sublevelprob(\genRealOne)$ in a Bayesian way, as it allows us to balance accuracy against computational complexity through uncertainty-quantification, which we use as a stopping criterion: We place a Beta-prior $\mathrm{Beta}(a^{(0)}, b^{(0)})$ over the interval $[0,1]$. As we do not have prior knowledge, and the map $\genRealOne \mapsto \sublevelprob(\genRealOne)$ can be discontinuous\footnote{Consider learning the step-size parameter $\notation{\realHyp} > 0$ for gradient descent on quadratic functions with largest eigenvalue $L$: The algorithm converges for $\notation{\realHyp} < \frac{2}{L}$ ($\sublevelprob(\notation{\realHyp}) = 1$) and diverges for $\notation{\realHyp} > \frac{2}{L}$ ($\sublevelprob(\notation{\realHyp}) = 0$) .}, we use a noninformative prior \citep[Ch. 3.3]{Berger_1985}, that is, $a_0 = b_0 = 1$. Since the Beta distribution is the conjugate prior for the Bernoulli distribution \citep[p.130]{Berger_1985}, that is, the posterior is again a Beta-distribution, after observing a sample $\notation{\rv{I}_{k+1}}$, the parameters $a^{(k)}, b^{(k)}$ get updated as:
$$
    a^{(k+1)} = a^{(k)} + \notation{\rv{I}_{k+1}}, \qquad b^{(k+1)} = b^{(k)} + (1 - \notation{\rv{I}_{k+1}}) \,.
$$
This allows us to do the estimation iteratively: We only draw a new sample $\notation{\rv{I}_{n+1}}$ as long as $Q^{(n)}(\quantile_u) - Q^{(n)}(\quantile_l) \ge \varepsilon$, where $Q^{(n)}$ denotes the quantile-function of $\mathrm{Beta}(a^{(n)}, b^{(n)})$, and $\quantile_u, \quantile_l, \varepsilon \in [0,1]$ are parameters that specify the accuracy of the estimation. Finally, one can use the posterior mean $\frac{a^{(n)}}{a^{(n)}+b^{(n)}}$ or posterior mode $\frac{a^{(n)}-1}{a^{(n)}+b^{(n)}-2}$ (provided $a^{(n)}, b^{(n)} > 1$) as point estimate $\hat{\sublevelprob}_{\genRealOne}$. By adjusting $\quantile_l, \quantile_u$ or $\varepsilon$, one can balance between accuracy and computational complexity. However, the number of iterations needed also depends on the true probability: For $\sublevelprob(\genRealOne) \approx 0$ or $\sublevelprob(\genRealOne) \approx 1$, the uncertainty decreases significantly faster than for $\sublevelprob(\genRealOne) \approx 0.5$. This procedure is summarized in Algorithm~\ref{Alg:Estimation_Probabilistic_Constraint} and depicted in Figure~\ref{fig:estimation_convergence_probability}. 

\subsubsection{Broader Context} 
Different, yet conceptually similar ideas for how to cut the computational cost of Bayesian Markov-Chain-Monte-Carlo algorithms through subsampling have been proposed: \citet{Korattikara_Chen_Welling_2014} use sequential hypothesis tests to reach the binary accept-reject decision in the Metropolis-Hastings algorithm. \citet{Bardenet_Doucet_Holmes_2014} estimate the accept-reject step in such a way that it coincides with the true accept-reject step with a user-specified probability. \citet{Maclaurin_Adams_2014} introduce an auxiliary binary variable $z_n \in \{0,1\}$, which allows for querying only a subset of the data for the computation of the exact likelihood. And \cite{Quiroz_Kohn_Villani_Tran_2018} combine subsampling with a bias-correction strategy to speed-up the sampling procedure. A summary of different approaches is given by \citet{Bardenet_Doucet_Holmes_2017}. We leave the analysis for our proposed approximation to future work.

\subsubsection{Choice of the Sampling Procedure}
Often, the hyperparameters $\notation{\realHyp} \in \notation{\hypSpace}$ are high-dimensional. Thus, we use \emph{stochastic gradient Langevin dynamics}~\citep{WellingTeh_2011} (SGLD) as the underlying sampling algorithm, and constrain it to the set $\hat{\set{A}}$ by use of the previously described procedure. This is summarized in Algorithm~\ref{Alg:Probabilistically_Constrained_Sampling}. However, if it fits the application, other sampling algorithms can be used, too. The computational overhead of the additional estimation depends on the cost of evaluating $\mathds{1}\{\notation{\genRvTwo_n} \in \set{C}_{\genRealOne}\}$. In our case it is expensive: Every sample $\notation{\rv{I}_n}$ requires to run the algorithm $\mathcal{A}$, which corresponds to approximating the solution of a minimization problem.

\begin{Rem}
    Algorithm~\ref{Alg:Probabilistically_Constrained_Sampling} requires to start in the set $\hat{\set{A}}$. If such a point is not known, one can still run the algorithm and just \say{start} the accept-reject mechanism as soon as one has found a point $\genRealOne \in \hat{\set{A}}$. However, it is not guaranteed that such a point will actually be found. 
\end{Rem}

\begin{algorithm}[t!]
\caption{Probabilistically constrained sampling}\label{Alg:Probabilistically_Constrained_Sampling}
    \begin{algorithmic}
    \Require $\sublevelprob_l, \sublevelprob_u \in [0, 1]$, $n_{\rm{max}} \in \N$, $\genRealOne_0 \in \hat{\set{A}}$. 
    \State $n \gets 0$ and $\genRealOne \gets \genRealOne_0$
    \While{$n \le n_{\rm{max}}$}
        \State 1) Draw a proposal $\genRealOne'$ with SGLD starting from $\genRealOne$.
        \State 2) Estimate $\sublevelprob(\genRealOne') = \prob_{\notation{\genRvTwo}}[\set{C}_{\genRealOne'}]$ by $\hat{\sublevelprob}(\genRealOne')$ with Algorithm~\ref{Alg:Estimation_Probabilistic_Constraint}. 
        \If{ $\hat{\sublevelprob}(\genRealOne') \in [\sublevelprob_l, \sublevelprob_u]$}
        \State $\genRealOne \gets \genRealOne'$
        \Else
        \State Reject $\genRealOne'$.
        \EndIf
    \EndWhile
    \end{algorithmic}
\end{algorithm}

\begin{figure}[t!]
    \centering
    \includegraphics[width=\textwidth]{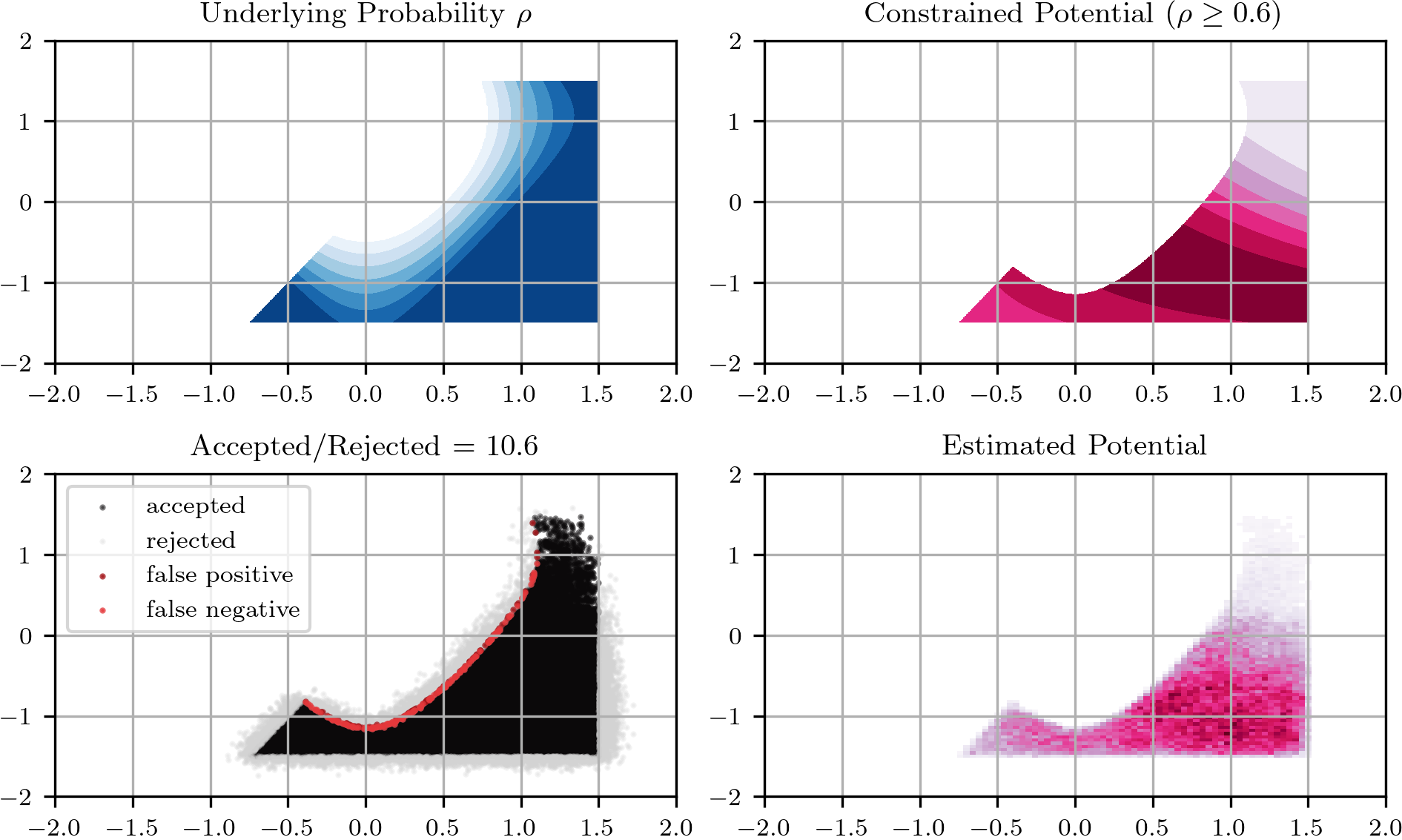}
    \caption{Example for probabilistically constrained sampling: The upper left plot shows the underlying function $\sublevelprob(\genRealOne)$. It is discontinuous and defines a non-convex set $\set{A}$. The upper right plot shows the probabilistically constrained potential ($\sublevelprob(\genRealOne) \in [0.6, 1]$), from which we want to sample. The lower left plot shows the accepted (black) and the rejected (gray) samples (in a ratio of about 10:1). Further, we can see that some of them are false-positives (dark red) or false-negatives (red). Especially, this happens for $\sublevelprob(\genRealOne) \approx 0.6$, where the remaining uncertainty can easily lead to a wrong decision. Here, we have chosen the $\quantile_l = 0.01$, $\quantile_u = 0.99$, and $\varepsilon = 0.05$ in Algorithm~\ref{Alg:Estimation_Probabilistic_Constraint}. Finally, the lower right plot shows the estimated potential.}
    \label{fig:example_probabilistically_constrained_sampling}
\end{figure}

\noindent
The results of applying this procedure for a two-dimensional toy example are shown in Figure~\ref{fig:example_probabilistically_constrained_sampling}: The upper row shows the function $\genRealOne \mapsto \sublevelprob(\genRealOne)$, and the potential from which we want to sample with the constraint $\sublevelprob(\genRealOne) \in [0.6, 1]$. The lower row shows the accepted (black) and rejected (gray) samples, and the final estimate of the constrained potential. While most samples get accepted/rejected correctly, some are actually false-positives (dark red) or false-negatives (red). Yet, this is to be expected. Note that, for simplicity, we did use full gradients here. We are now in a position to describe the whole learning procedure.

\section{Learning Procedure}\label{Sec:Learning_Procedure}

\begin{figure}[t!]
    \centering
    \input{Learning_Procedure/Images/learning_procedure_four_steps}
    \caption{Learning procedure: \textbf{1)} Imitation learning. \textbf{2)} Probabilistically constrained stochastic empirical risk minimization. \textbf{3)} Construct prior through sampling. \textbf{4)} Compute posterior by performing the PAC-Bayesian learning step.}
    \label{fig:learning_procedure_highlevel}
\end{figure}

This section deals with the implementation of the learning procedure, and translates the abstract framework discussed in Sections~\ref{Sec_PAC} and \ref{Subsec:learning_with_guarantees} into concrete design choices. Thus, this marks the beginning of the second part of the paper, which is less theoretical. The resulting learning procedure is visualized in Figure~\ref{fig:learning_procedure_highlevel} and consists of four steps:  
\begin{itemize}[leftmargin=25pt]
    \item[(i)] Step one: Train the algorithm to \say{mimic} another algorithm $\mathcal{A}'$. This is needed only, if one cannot choose stable initial hyperparameters directly, for example, when the update includes a neural network. Otherwise, the algorithm might predict points that are so far off that one encounters numerical instabilities.  
    \item[(ii)] Step two: Find a point $\iter{\notation{\realHyp}}{0} \in \notation{\hypSpace}$ that a) satisfies the constraint in Subsection~\ref{SubSec_CondOnConv} and b) yields a good performance. For this, we perform a constrained version of stochastic empirical risk minimization with a new, specifically designed loss function.
    \item[(iii)] Step three: Starting from $\iter{\notation{\realHyp}}{0}$, construct the prior distribution by running a constrained version of a sampling algorithm.
    \item[(iv)] Step four: Find the optimal $\idx^* \in \idxset$, which allows for computing the optimal posterior distribution $\mathbb{Q}_{\idx^*}$ in closed-form.
\end{itemize}  
The outline of this section is as follows:
In Subsection~\ref{Sec:minimization_PAC_bound} we identify the optimal posterior $\mathbb{Q}^*$ in the abstract setting. In Subsection~\ref{Subsec:Initialization}, we describe the pre-computation phase in (i). Subsections~\ref{Subsec:Good_Performance_and_Constraint} and \ref{Subsec:Sampling_Probabilistic_Constraints} deal with the concrete design choices in (ii) and (iii) to construct the prior, and Subsection~\ref{Subsec:Compute_Posterior} yields the posterior distribution in (iv).
Since the prior has to be independent of the data set that is used in the PAC-Bayesian step, we split the data set $\notation{\rvData}$ into independent parts $\notation{\rvData}_{\rm{prior}}$, $\notation{\rvData}_{\rm{val}}$, $\notation{\rvData}_{\rm{train}}$ and $\notation{\rvData}_{\rm{test}}$, where $\notation{\rvData}_{\rm{prior}}$ and $\notation{\rvData}_{\rm{val}}$ are used for the construction of the prior distribution, $\notation{\rvData}_{\rm{train}}$ is used for the PAC-Bayesian learning step, and $\notation{\rvData}_{\rm{test}}$ is the test set which is only needed for the experiments. Nevertheless, for notational simplicity, we will use the generic $\notation{\rvData}$, implicitly assuming the above partitioning. 
\begin{Rem}
    Through the choice of the sampling algorithm, the concrete learning procedure described here mainly applies to the case $\notation{\hypSpace} = \R^d$, $d \in \N$. Nevertheless, the general methodology is still applicable to other Polish spaces, if this choice can be adjusted accordingly.
\end{Rem}

\subsection{Minimization of the PAC-Bound}\label{Sec:minimization_PAC_bound}
Learning is phrased as minimizing the PAC-Bayesian upper-bound. Hence, in this subsection we consider $\natPar$, $\suffStat$ and item $(ii)$ from Corollary~\ref{Cor_generalization_bound_exp_family}, and we seek for $\idx \in \idxset$ and $\mathbb{Q} \in \probMeasures(\mathbb{P}_{\notation{\rvHyp}})$ that minimize the upper-bound, that is, we want to solve:
$$
    \inf_{\idx \in \idxset} \ \inf_{\mathbb{Q} \in \probMeasures(\prob_{\notation{\rvHyp}})} \ \mathbb{Q}[\hat{\mathcal{R}}(\cdot, \notation{\realData})] + \frac{\left(\divergence{\rm{KL}}{\mathbb{Q}}{\mathbb{P}_{\notation{\rvHyp}}} + \log\left(\frac{\coveringNumber}{\varepsilon}\right) + \coveringConstant - \mathbb{Q}\left[\sprod{\natPar^{(r)}(\idx)}{\suffStat^{(r)}(\cdot, \notation{\realData})}\right] \right)}{\natPar^{(1)}(\idx)}  \,.
$$
By factoring out $-\frac{1}{\natPar^{(1)}(\idx)}$ again, this is actually the same as:
\begin{align*}
    \inf_{\idx \in \idxset} \ -\frac{1}{\natPar^{(1)}(\idx)} \left(\sup_{\mathbb{Q} \in \probMeasures(\mathbb{P}_{\notation{\rvHyp}})} 
    \mathbb{Q}[\sprod{\natPar(\idx)}{\Tilde{\suffStat}(\cdot, \notation{\realData})}]
    - \divergence{\rm{KL}}{\mathbb{Q}}{\mathbb{P}_{\notation{\rvHyp}}} 
    - \log \left( \frac{\coveringNumber}{\varepsilon} \right) - \coveringConstant \right),
\end{align*}
where $\Tilde{\suffStat}(\notation{\realHyp, \realData}) := \bigl(-\hat{\mathcal{R}}(\notation{\realHyp, \realData}), \ \suffStat^{(r)}(\notation{\realHyp, \realData}) \bigr)$. Since $\log(\coveringNumber/\varepsilon) + \coveringConstant$ is a constant, Lemma \ref{Lem_DonskerVaradhanForExponentialFamilies} shows that the term inside the brackets is given by $\Tilde{\kappa}(\idx, \notation{\realData}) - \log( K/\varepsilon) - C_\mathcal{O}$, where $\Tilde{\kappa}$ corresponds to the exponential family $(\Tilde{\mathbb{Q}}_{\idx})_{\idx \in \idxset}$ built upon $\Tilde{\suffStat}$ and $\natPar$ (with $\expFamFunc \equiv 1$). Furthermore, the optimal posterior distribution $\mathbb{Q} \in \probMeasures(\prob_{\notation{\rvHyp}})$ is given by the corresponding member of the data-dependent exponential family $\Tilde{\mathbb{Q}}_{\idx}(\notation{\realData}) \propto \exp(\sprod{\natPar(\idx)}{\Tilde{\suffStat}(\cdot, \notation{\realData})}) \cdot \prob_{\notation{\rvHyp}}$, usually called the Gibbs posterior \citep{Alquier_2024}.
By denoting $F(\idx, \notation{\realData}) := -\frac{1}{\natPar^{(1)}(\idx)}( \Tilde{\kappa}(\idx, \notation{\realData}) - \log(\coveringNumber / \varepsilon) - \coveringConstant)$, one is left with solving the following problem:
\begin{equation}\label{Eq:MinF}
    \inf_{\idx \in \idxset} \ F(\idx, \notation{\realData}),
\end{equation}
which for $\idxset \subset \mathbb{R}$ is one-dimensional. Based on Theorem~\ref{Thm_PacBayesConditioned}, we restrict to $\idxset \subset (0, +\infty)$, such that the solution to \eqref{Eq:MinF} can be seen as an approximation to the global minimum $\inf_{\idx > 0} F(\idx, \notation{\realData})$. For the latter one, one can show that the solution set lies in a compact interval $[\idxset_{\rm{min}}, \idxset_{\rm{max}}]$, since $F(\idx, \notation{\realData}) \to \infty$ as $\idx \to 0$ or $\idx \to \infty$. Under our assumptions, $F(\cdot, \notation{\realData})$ is continuously differentiable. Hence, since $\idxset$ is compact, $F(\cdot, \notation{\realData})$ is Lipschitz-continuous on $\idxset$ and the minimum in \eqref{Eq:MinF} is attained. 
For a finite set $\idxset = \{\idx_1, ..., \idx_K\} \subset [\idxset_{\rm{min}}, \idxset_{\rm{max}}]$, the optimization reduces to grid search. For $\idxset = [\idxset_{\rm{min}}, \idxset_{\rm{max}}]$, we employ grid search as initialization for gradient-based optimization. Here, the computational bottleneck is given by evaluating $\idx \mapsto \Tilde{\kappa}(\idx, \notation{\realData})$. In Sections~\ref{Subsec:Sampling_Probabilistic_Constraints} and \ref{Subsec:Compute_Posterior} we will ensure that this is cheap.

\subsection{Finding a Trainable Initialization}\label{Subsec:Initialization}

\begin{algorithm}[t!]
\caption{Procedure to find an initialization}\label{Alg:Finding_Initialization}
    \begin{algorithmic}
    \Require Data set $\notation{\realData}_{\rm{prior}}$, $\iter{x}{0} \in \R^n$, $\trajectorylength, n_{\rm{init}} \in \N$ and $\varepsilon > 0$.
    \State $m \gets +\infty$ and sample $\notation{\realPar} \sim \mathit{U}_{\notation{\realData}_{\rm{prior}}}$.
    \While{$\frac{1}{n_{\rm{init}}} m > \varepsilon$} 
        \State $m \gets 0$
        \For{$i = 1, ..., n_{\rm{init}}$}
            \State 1) Compute $(\iter{x}{1}, \iter{y}{1}),...,(\iter{x}{\trajectorylength}, \iter{y}{\trajectorylength})$ with $\mathcal{A}(\notation{\realHyp, \realPar}, \iter{x}{0})$ and $\mathcal{A}'(\notation{\realPar}, \iter{x}{0})$, resp.
            \State 2) Compute $\ell_{\rm{init}}(\notation{\realHyp, \realPar}, \iter{x}{0}, \trajectorylength) = \frac{1}{\trajectorylength} \sum_{k=1}^{\trajectorylength} \norm{\iter{x}{k} - \iter{y}{k}}{2}^2$.
            \State 3) Update $m \gets m + \ell_{\rm{init}}(\notation{\realHyp, \realPar}, \iter{x}{0}, \trajectorylength)$
            \State 4) Update $\notation{\realHyp}$ by backpropagation and Adam.
            \Comment{Other algorithms possible.}
            \State 5) Update $\notation{\realPar}$, $\iter{x}{0}$ and $\trajectorylength$ based on Section~\ref{Subsec:Good_Performance_and_Constraint}.
        \EndFor
    \EndWhile
    \end{algorithmic}
\end{algorithm}

To increase numerical stability, we start with \say{imitation learning} \citep{Chen_Zhang_Jingyang_Chang_Liu_Amini_Wang_2020}, that is, the algorithm $\mathcal{A}$ should \say{follow} another algorithm $\mathcal{A}'$, for example, gradient descent. For this, we minimize the mean squared error between the iterates of the two algorithms: Given a starting point $\iter{x}{0} \in \R^n$, an iteration number $\trajectorylength \in \N$, and a parameter $\notation{\realPar \in \parSpace}$, denote the first $\trajectorylength$ iterates of $\mathcal{A}(\notation{\realHyp, \realPar}, \iter{x}{0})$ by $\iter{x}{1}, ..., \iter{x}{\trajectorylength} \in \R^n$ and the ones of $\mathcal{A}'(\iter{x}{0}, \notation{\realPar})$ by $\iter{y}{1}, ..., \iter{y}{\trajectorylength} \in \R^n$. Then, define the loss as the mean squared error over these iterations:
$$
    \ell_{\rm{init}}(\notation{\realHyp, \realPar}, \iter{x}{0}, \trajectorylength) := \frac{1}{\trajectorylength} \sum_{k=1}^{\trajectorylength} \norm{\iter{x}{k} - \iter{y}{k}}{2}^2 \,.
$$
In each iteration, that is, each prediction of tuples $(\iter{x}{1}, \iter{y}{1}), ..., (\iter{x}{\trajectorylength}, \iter{y}{\trajectorylength})$, the parameters $\notation{\realPar}$, $x^{(0)}$ and $\trajectorylength$ are randomized as described in Section~\ref{Subsec:Good_Performance_and_Constraint}. It is not necessary to reach a high accuracy here, as the purpose is to prevent divergence, and not actual imitation of $\mathcal{A}'$. The procedure is summarized in Algorithm~\ref{Alg:Finding_Initialization}.

\subsection{Locating the Prior}\label{Subsec:Good_Performance_and_Constraint}

\begin{algorithm}[t!]
\caption{Procedure to locate the prior}\label{Alg:Training_Procedure}
    \begin{algorithmic}
    \Require Data sets $\notation{\realData}_{\rm{prior}}, \notation{\realData}_{\rm{val}}$, numbers $n_{\rm{max}}, n_{\rm{train}}, \trajectorylength \in \N$ with $\trajectorylength \le n_{\rm{train}}$, initialization $\iter{x}{0}$ and thresholds $\sublevelprob_l, \sublevelprob_u \in [0,1]$ with $\sublevelprob_l < \sublevelprob_u$. 
    \State ...........................................................................................................................................
    \State Set $x \gets \iter{x}{0}$, $b \gets \mathbf{false}$, and sample $\notation{\realPar} \sim \mathrm{U}_{\notation{\realData}_{\rm{prior}}}$ \Comment{$b$ = Point inside constraint?}
    \State ...........................................................................................................................................
    \For{$i = 1,...,n_{\rm{max}}$} \Comment{Other stopping criteria possible.}
        \State ......................................................................................................................................
        \State 1.a) Compute $\iter{x}{1}, ..., \iter{x}{\trajectorylength}$ with $\mathcal{A}(\notation{\realHyp, \realPar}, \iter{x}{0})$. 
        \State 1.b) Compute $\ell_{\rm{train}}(\notation{\realHyp, \realPar}, \iter{x}{0}, \trajectorylength) = \sum_{i=1}^{\trajectorylength} \mathds{1}_{\{\ell(\iter{x}{i-1}, \notation{\realPar}) > 0\}} \frac{\ell(\iter{x}{i}, \notation{\realPar})}{\ell(\iter{x}{i-1}, \notation{\realPar})}$.
        \State 1.c) Construct a proposal $\Tilde{\notation{\realHyp}}$ by using backpropagation and Adam.
        \State ......................................................................................................................................
        \State 2) Estimate $\sublevelprob(\Tilde{\notation{\realHyp}})$ by $\hat{\sublevelprob}(\Tilde{\notation{\realHyp}})$ with Algorithm~\ref{Alg:Estimation_Probabilistic_Constraint} on $\notation{\realData}_{\rm{val}}$.
        \If{ $\hat{\sublevelprob}(\Tilde{\notation{\realHyp}}) \in [\sublevelprob_l, \sublevelprob_u]$} \Comment{If point inside constraint, just update.}
            \State $\notation{\realHyp} \gets \Tilde{\notation{\realHyp}}$ and $b \gets \mathbf{true}$
        \Else \Comment{If not...}
            \If{$b = \mathbf{true}$} \Comment{...reject moving outside constraint.}
                \State Reject $\Tilde{\notation{\realHyp}}$, set $\iter{x}{0} \gets x$, sample $\notation{\realPar} \sim \mathrm{U}_{\notation{\realData}_{\rm{prior}}}$, and continue with 1).
            \Else \Comment{...accept, if constraint has not been found yet.}
                \State $\notation{\realHyp} \gets \notation{\Tilde{\realHyp}}$
            \EndIf
        \EndIf
        \State ......................................................................................................................................
        \State 3) Draw $\notation{\rv{R}} \sim \mathrm{Ber}(\frac{\trajectorylength}{n_{\rm{train}}})$. 
        \If{ $\notation{\rv{R}} = 0$}
            \State $\iter{x}{0} \gets \iter{x}{\trajectorylength}$
        \Else{ $\notation{\rv{R}} = 1$}
            \State $\iter{x}{0} \gets x$ and sample $\notation{\realPar} \sim \mathrm{U}_{\notation{\realData}_{\rm{prior}}}$
        \EndIf
        \State ......................................................................................................................................
    \EndFor
    \end{algorithmic}
\end{algorithm}

Empirically, the performance of the learned algorithm is significantly improved by the following two design choices. The motivation is to prevent overfitting and to learn a scale-independent contraction of the loss: 

\subsubsection{Ratio of Losses}
The canonical loss function to be minimized is the empirical risk $\hat{\mathcal{R}}(\notation{\realHyp, \realData}) = \frac{1}{N} \sum_{i=1}^N \ell(\notation{\realHyp, \realPar_i})$, and, if $\notation{\hypSpace}$ is high-dimensional or if $N$ is large, one resorts to stochastic empirical risk minimization. While this kind of loss was used extensively before, for learning-to-optimize it has a strong disadvantage: Only the overall outcome after $n_{\rm{max}}$ iterations gets penalized. Thus, it does not take the performance along the trajectory into account. Further, often it is hard to minimize (due to training instabilities) and does not lead to the desired performance. To circumvent this, \cite{Andrychowicz_Denil_Gomez_Hoffman_Pfau_Schaul_Shillingford_DeFreitas_2016} proposed to use $\Tilde{\ell}_{\rm{train}}(\notation{\realHyp, \realPar}, \iter{x}{0}) := \sum_{i=1}^n \ell(\iter{x}{i}, \notation{\realPar})$. Again, this formulation has a decisive flaw: Under most objectives, if the algorithm performs reasonably well, the loss at the beginning is \emph{several orders} of magnitude larger than the loss at the end. Hence, $\Tilde{\ell}_{\rm{train}}$ mainly penalizes the loss at the beginning, leading to an algorithm that minimizes the loss very fast in early iterations, yet slows down a lot in later iterations. This is due to $\Tilde{\ell}_{\rm{train}}$ being \emph{scale-sensitive}. Additionally, the incurred loss might vary strongly with the initialization $\iter{x}{0}$ alone, thereby introducing ambiguity into the incurred losses. We propose to use the \emph{ratio of consecutive losses}:
$$
    \ell_{\rm{train}}(\notation{\realHyp, \realPar}, \iter{x}{0}, \trajectorylength) := \sum_{i=1}^{\trajectorylength} \mathds{1}_{\{\ell(\iter{x}{i-1}, \notation{\realPar}) > 0\}} \frac{\ell(\iter{x}{i}, \notation{\realPar})}{\ell(\iter{x}{i-1}, \notation{\realPar})} \,, \qquad \trajectorylength \in \N, \ \trajectorylength \le n \,.
$$
This has several advantages: First, the loss is not scale-sensitive anymore, such that it favors hyperparameters that yield a good performance in each iteration. Second, there is no ambiguity in the observed loss through the initialization, as the only criterion is a strong contraction of the loss (instead of a small loss). Third, the incurred losses do not vary too much, which empirically makes it easier to choose hyperparameters of the learning procedure. However, it also has a disadvantage: If the function values do indeed converge in a setting where the optimal loss is strictly greater than zero, this gets fully penalized, as then $\frac{\ell(\iter{x}{i}, \notation{\realPar})}{\ell(\iter{x}{i-1}, \notation{\realPar})} \equiv 1$. For now, we do not know how to avoid this problem (apart from just stopping the iterations in case of convergence) while keeping the advantages.

\begin{figure}
    \centering
    \includegraphics[width=1.\linewidth]{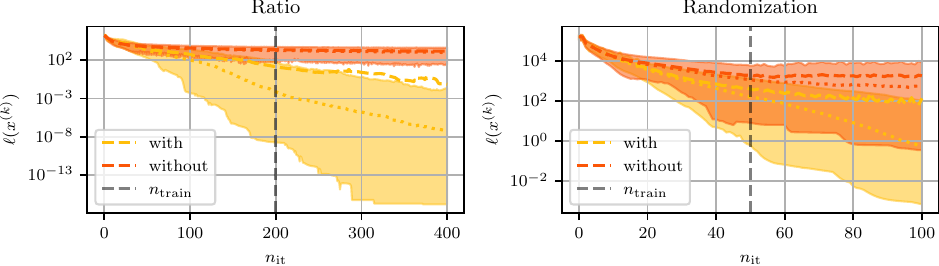}
    \caption{Effect of our design choices: Dashed lines represent the mean losses, dotted lines represent the median losses, and the shaded region represents 95\% of the data. The yellow algorithm was trained with our design choice and the orange one without. Besides that, everything else was kept the same. In the left plot we can see that using the ratio of consecutive losses strongly improves the performance, and in the right plot we can see that the randomization procedure yields generalization beyond $n_{\rm{train}}$ \emph{and} an overall better performance.}
    \label{fig:effect_of_design_choices}
\end{figure}

\subsubsection{Randomized Trajectory Length}\label{Paragraph:randomized_trajectory_lenght} 
Training $\mathcal{A}$ with fixed initialization $\iter{x}{0}$ and fixed trajectory length leads to overfitting: Applying it at another starting point $\iter{\Tilde{x}}{0}$ or applying it for more iterations typically does not work, or even leads to divergence. To avoid this, we propose the following randomization: Fix $\trajectorylength \le n_{\rm{train}}$ and set $y := \iter{x}{0}$.
\begin{itemize}[leftmargin=25pt]
    \item[0)] Sample a parameter $\notation{\realPar}$ uniformly at random from $\notation{\realData}$.
    \item[1)] Compute $\iter{x}{1}, ..., \iter{x}{\trajectorylength}$ with $\mathcal{A}(\notation{\realHyp, \realPar}, y)$ and the loss $\ell_{\rm{train}}(\notation{\realHyp, \realPar}, y, \trajectorylength)$, and update $\notation{\realHyp}$.
    \item[2)] Sample $\notation{\iter{\rv{R}}{k}} \sim \mathrm{Ber}(\frac{\trajectorylength}{n_{\rm{train}}})$. If $\notation{\iter{\rv{R}}{k}} = 0$, set $y := \iter{x}{\trajectorylength}$ and go to step 1). If $\notation{\iter{\rv{R}}{k}} = 1$, set $y := \iter{x}{0}$ and go to step 0). 
\end{itemize}
The random variable $\notation{\iter{\rv{R}}{k}} $ decides whether the algorithm gets restarted from $\iter{x}{0}$ with a new parameter $\notation{\Tilde{\realPar}}$, or if one continuous the current trajectory. The choice $\frac{\trajectorylength}{n_{\rm{train}}}$ ensures that the expected trajectory-length equals $n_{\rm{train}}$: Define $\stoppingtime := \inf \{k \in \N \ : \ \notation{\iter{\rv{R}}{k}} = 1\}$.
Then, $\stoppingtime \sim \mathrm{Geo}(\frac{\trajectorylength}{n_{\rm{train}}})$ is a geometrically distributed with expectation $\expectation[\stoppingtime] = \frac{n_{\rm{train}}}{\trajectorylength}$. Therefore, for the actual length $\notation{\rv{L}} = \trajectorylength \cdot \stoppingtime$ of the trajectory we get $\expectation [\notation{\rv{L}}] = \trajectorylength \expectation[\stoppingtime] = n_{\rm{train}}$. 
\begin{Rem}
    Similarly to \cite{Andrychowicz_Denil_Gomez_Hoffman_Pfau_Schaul_Shillingford_DeFreitas_2016}, we omit the computation of second-order derivatives during training. Additionally, and surprisingly, it usually suffices to consider single iterates, that is $\trajectorylength = 1$. That amounts to learning an update step that is agnostic to the recurrent nature of the optimization algorithm and just learns to adapt to the local geometry of the loss function \emph{along the iterations}.
\end{Rem}
\noindent
Figure~\ref{fig:effect_of_design_choices} shows the effect of these two design choices: The left plot shows the effect of using the ration of consecutive losses and the right plot shows the effect of randomizing the trajectory. In both cases, we train two times the same algorithm: One time with our proposed choice (yellow), and one time without (orange). Everything else is kept the same, that is, both were trained with Algorithm~\ref{Alg:Training_Procedure}. In the left plot one can see that the ratio of losses strongly improves the performance compared to using normal function-values, and in the right plot one can see that the randomization procedure improves the generalization to more iterations \emph{and} its performance. However, please note that there might be some bias: The architecture of the algorithm is one that we have found using our proposed training procedure. Further details can be found in the GitHub-repository. The overall procedure is summarized in Algorithm~\ref{Alg:Training_Procedure}.

\subsection{Constructing the Prior}\label{Subsec:Sampling_Probabilistic_Constraints}

\begin{algorithm}[t!]
\caption{Procedure to construct the prior}\label{Alg:Construction_Prior}
    \begin{algorithmic}
    \Require Data sets $\notation{\realData}_{\rm{prior}}$ (sampling) and $\notation{\realData}_{\rm{val}}$ (constraint), $\nSample \in \N$ and $\notation{\realHyp} \in \mathrm{supp} ({\Tilde{\prob}_{\notation{\realHyp}}})$. 
    \State 1) Starting from $\notation{\realHyp}$, run Algorithm~\ref{Alg:Probabilistically_Constrained_Sampling} (with $\ell_{\rm{train}}$) to get the points $\notation{\realHyp_1, ..., \realHyp_{\nSample}} \in \notation{\hypSpace}$.
    \State 2) Evaluate $\varphi_{\rm{prior}}$ on $\{\notation{\realHyp_1, ..., \realHyp_{\nSample}}\}$ by evaluating $\empsublevelrisk{g}$ corresponding to $\notation{\realData}_{\rm{prior}}$.
    \State 3) Compute $\prob_{\notation{\rvHyp}} \{\notation{\realHyp_j}\}$, that is,
    $
        \mathbb{P}_{\notation{\rvHyp}}\{\notation{\realHyp_j}\} = \sigma \left( \varphi_{\rm{prior}}(\notation{\realHyp_1}), ..., \varphi_{\rm{prior}}(\notation{\realHyp_{\nSample}}) \right)_j \,.
    $
    \end{algorithmic}
\end{algorithm}

Besides the performance and the sublevel guarantees, the only assumption on the prior $\prob_{\notation{\rvHyp}}$ is its independence of $\notation{\rvData}_{\rm{train}}$. Further, by Lemma~\ref{Lem_DonskerVaradhanForExponentialFamilies} the functional form of the posterior is fully specified, namely it is of the form:
\begin{equation}\label{Eq:Gibbs_Posterior}
    \mathbb{Q}_{\idx}(\notation{\realData}) \propto \exp(\varphi_{\idx}(\cdot, \notation{\realData})) \cdot \mathbb{P}_{\notation{\rvHyp}} , \qquad \idx \in \idxset\,,
\end{equation}
where the potential is given by $\varphi_{\idx}(\notation{\realHyp}, \notation{\realData}) = \sprod{\natPar(\idx)}{\Tilde{\suffStat}(\notation{\realHyp, \realData})}$. Hence, for mathematical convenience, we will construct $\prob_{\notation{\rvHyp}}$ by approximating the distribution $\mathbb{P}'$ given by
$$
    \mathbb{P}' \propto \exp \left( - \empsublevelrisk{\sublevelfunction, \rm{prior}} - \iota_{[\sublevelprob_l, \sublevelprob_u]} \circ \sublevelprob \right) \cdot \mu \,,
$$
where $\mu$ is a measure on $\notation{\hypSpace}$, which allows to sample from $\mathbb{P}'$ (possibly unnormalized). In our experiments it holds $\notation{\hypSpace} = \R^d$ and we choose $\mu = \lambda^d$, where $\lambda^d$ is the $d$-dimensional Lebesgue measure. For sampling, we use the \emph{stochastic gradient Langevin dynamics} algorithm, where we use the output of the backpropagation algorithm as proxy for the (sub)gradient. Finally, since anyway we have to resort to a sampling algorithm to get points $\notation{\realHyp_1, ..., \realHyp_{\nSample}} \in \notation{\hypSpace}$, $\nSample \in \N$, we define the prior distribution directly as a discrete distribution, that is $\mathbb{P}_{\notation{\rvHyp}}\{\notation{\realHyp}\} := \frac{1}{Z} \sum_{i=1}^{\nSample} \notation{w}_i \delta_{\notation{\realHyp_i}}\{\notation{\realHyp}\}$. Thus, $\prob_{\notation{\rvHyp}}$ is the suitably normalized discrete measure on $\notation{\hypSpace}$ corresponding to $\notation{\realHyp_1, ..., \realHyp_{\nSample}}$, where the normalization constant is given by
$Z = \sum_{i=1}^{\nSample} \notation{w}_i$ with $\notation{w}_i = \exp \left(-\empsublevelrisk{\sublevelfunction, \rm{prior}}(\notation{\realHyp_i}) - \iota_{[\sublevelprob_l, \sublevelprob_u]} \left( \hat{\sublevelprob}(\notation{\realHyp_i}) \right) \right)$. When $\notation{\realHyp_1, ..., \realHyp_{\nSample}} \in \notation{\hypSpace}$ are given, the corresponding probabilities can equivalently be expressed with the so-called \emph{softmax} function $\sigma(x_1, ..., x_n)_j = \frac{\exp(x_j)}{\sum_{i=1}^{n} \exp(x_i)}$ and the potential $\varphi_{\rm{prior}}(\notation{\realHyp})= -\empsublevelrisk{\sublevelfunction, \rm{prior}}(\notation{\realHyp}) - \iota_{[\sublevelprob_l, \sublevelprob_u]} \left( \hat{\sublevelprob}(\notation{\realHyp}) \right)$:
\begin{align*}
    \mathbb{P}_{\notation{\rvHyp}}\{\notation{\realHyp_j}\} 
    &= \frac{\exp \left( \varphi_{\rm{prior}}(\notation{\realHyp_j}) \right) }{\sum_{i=1}^{\nSample} \exp \left(\varphi_{\rm{prior}}(\notation{\realHyp_i}) \right)} 
    = \sigma \left( \varphi_{\rm{prior}}(\notation{\realHyp_1}), ..., \varphi_{\rm{prior}}(\notation{\realHyp_{\nSample}}) \right)_j \,.
\end{align*}
Here, the potentials $\varphi_{\rm{prior}}$ have to be computed \emph{only once} for every $\notation{\realHyp_i}$, $i=1,...,\nSample$. This is summarized in Algorithm~\ref{Alg:Construction_Prior}.
\begin{Rem}
    As one would approximate the intractable integrals with Monte-Carlo estimates anyway, choosing a discrete measure is less restrictive than it seems, and it has the additional advantage of allowing for \emph{exact} instead of approximate inference.
\end{Rem}

\subsection{Computing the Posterior}\label{Subsec:Compute_Posterior}

Given a discrete prior $\prob_{\notation{\rvHyp}}$, every posterior $\mathbb{Q} \in \probMeasures(\prob_{\notation{\rvHyp}})$ is also discrete with the same support $\{ \notation{\realHyp_1, ..., \realHyp_{\nSample}} \}$. Then, by the closed-form solution~\eqref{Eq:Gibbs_Posterior}, for every $\idx \in \idxset$ the optimal posterior $\mathbb{Q}_{\idx}(\notation{\realData})$ is given by:
\begin{align*}
    \mathbb{Q}_{\idx} (\notation{\realData}, \{ \notation{\realHyp_j}\}) 
    &= \frac{\exp \left( \sprod{\natPar(\idx)}{\Tilde{\suffStat}(\notation{\realHyp_j, \realData})} + \varphi_{\rm{prior}}(\notation{\realHyp_j}) \right)}{\sum_{i=1}^{\nSample} \exp\left( \sprod{\natPar(\idx)}{\Tilde{\suffStat}(\notation{\realHyp_i, \realData})} + \varphi_{\rm{prior}}(\notation{\realHyp_j}) \right)}
    = \sigma \left(\varphi_{\idx}(\notation{\realHyp_1, \realData}), ..., \varphi_{\idx}(\notation{\realHyp_{\nSample}, \realData}) \right)_j\,,
\end{align*}
with the potential $\varphi_{\idx}(\notation{\realHyp, \realData}) = \sprod{\natPar(\idx)}{\Tilde{\suffStat}(\notation{\realHyp, \realData})} + \varphi_{\rm{prior}}(\notation{\realHyp})$. Thus, to get the distribution $\mathbb{Q}_{\idx}(\notation{\realData})$ as a function of $\idx$, one has to compute $\Tilde{\suffStat}(\notation{\realHyp_i, \realData})$ \emph{only once} for every $i=1,...,\nSample$, such that it can be evaluated with the softmax function. Hence, the only missing ingredient is the optimal $\idx^* \in \idxset$, which is found as described in Section~\ref{Sec:minimization_PAC_bound}. After evaluating the potentials $\varphi_{\idx}(\cdot, \notation{\realData})$, which has to be done anyway, evaluating $\Tilde{\kappa}(\cdot, \notation{\realData})$ in $\idx$ is cheap. The process is summarized in Algorithm~\ref{Alg:Construction_Posterior}.

\begin{algorithm}[t!]
\caption{Procedure to construct the posterior}\label{Alg:Construction_Posterior}
    \begin{algorithmic}
    \Require Points $\{\notation{\realHyp_1, ..., \realHyp_{\nSample}}\}$, values $\{\varphi_{\rm{prior}}(\notation{\realHyp_1}), ..., \varphi_{\rm{prior}}(\notation{\realHyp_{\nSample}})\}$, data set $\realData = \notation{\realData}_{\rm{train}}$. 
    \State 1) Evaluate $\Tilde{T}(\notation{\realHyp_i, \realData})$, $i=1, ..., \nSample$.
    \State 2) Setup $\{\varphi_{\idx} (\notation{\realHyp_1, \realData}), ..., \varphi_{\idx} (\notation{\realHyp_{\nSample}, \realData})\}$ as functions in $\idx$.
    \State 3) Solve $\idx^* \in \argmin_{\idx \in \idxset} \ F(\idx, \notation{\realData})$. \Comment{$F(\idx^*, \notation{\realData})$ is the predicted PAC-bound.}
    \State 4) Compute $\mathbb{Q}_{\idx^*}(\notation{\realData, \{\realHyp_j\}}) = \sigma \left( \varphi_{\idx^*}(\notation{\realHyp_1, \realData}), ..., \varphi_{\idx^*}(\notation{\realHyp_{\nSample}, \realData}) \right)_j$, $j=1, ..., \nSample$.
    \State 5) Optional: Choose $\notation{\realHyp^*} = \argmax_{i = 1, ..., \nSample} \mathbb{Q}_{\idx^*}(\notation{\realData, \{\realHyp_{i}\}})$ as final point-estimate.
    \end{algorithmic}
\end{algorithm}

\section{Experiments}\label{Sec:Experiments}

We consider the \emph{smooth and strongly convex} problem of minimizing quadratic functions with varying strong convexity and smoothness constants, a \emph{high-dimensional} image processing problem, the \emph{non-smooth} Lasso problem, and the \emph{non-smooth and non-convex} problem of training a neural network. 
More details on the implementation, especially a detailed description of the architectures of the algorithms and how we construct the parameters for each problem, is given in Appendix~\ref{App:Implementation_details}. Alternatively, the code can be found in the GitHub-repository. 
In the evaluation, we will always show the \emph{loss over the iterations} in the upper left plot, the \emph{performance in terms of computation time} in the upper right plot, the \emph{loss histogram with predicted PAC-bound} in the lower left plot, and the final \emph{estimate for the sublevel probability}, that is, whether the probabilistically constrained optimization/sampling procedure did work correctly, in the lower right plot. Finally, note that we always show the performance of a single sample $\notation{\realHyp^*}$ (mode of the posterior), and not the mean performance. 

\subsection{Quadratics}\label{Subsec:Quadratics}

\begin{figure}[t!]
    \centering
    \includegraphics[width=\textwidth]{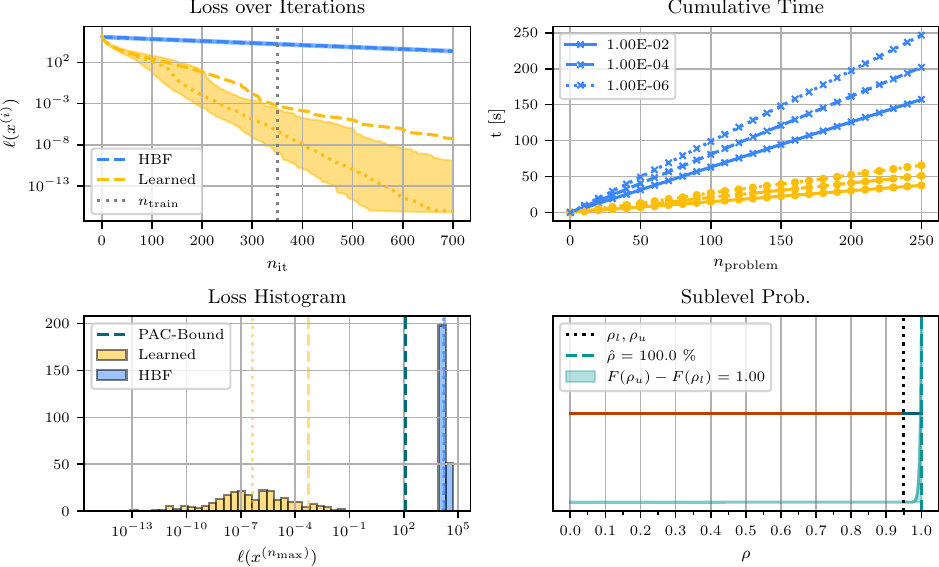}
    \caption{\textbf{Upper left:} Dashed lines represent the mean losses, dotted lines represent the median losses, and the shaded regions represent the 10th to 90th percentile. The learned algorithm $\mathcal{A}$ is shown in yellow, while \emph{heavy-ball with friction} (HBF) is shown in blue. 
    \textbf{Upper right:} The different lines indicate the cumulative computation time the  algorithms need to solve all the test problems up to a certain accuracy (in function values) measured by $\ell(\iter{x}{i}, \notation{\realPar}) < \varepsilon$. However, note that both algorithms are run for maximally $n_{\rm{max}} = 1\mathrm{e}4$ iterations.
    \textbf{Lower left:} Loss histogram after $n_{\rm{train}} = 350$ iterations with predicted PAC-bound.
    \textbf{Lower right:} The teal dashed line shows the point estimate for the sublevel probability, while the teal solid line shows the $\mathrm{Beta}$-posterior. The black dotted lines indicate the constraints $\sublevelprob_l, \sublevelprob_u$ and show the feasible region as dark teal line.} 
    \label{fig:experiments_quadratics}
\end{figure}

As first problem we consider strongly convex quadratic functions with varying strong convexity, varying smoothness and varying right-hand side, that is, each optimization problem is of the form:
$$
    \min_{x \in \R^n} \frac{1}{2} \Vert Ax - b \Vert^2 \,, \qquad A \in \R^{n \times n}, \ b \in \R^n\,.
$$
Thus, the parameters are given by $\notation{\realPar} = (A, b) \in \R^{n^2 + n} =: \notation{\parSpace}$, while the optimization variable is $x \in \R^n$, where we use $n = 200$.
By construction, each of these functions is $L$-smooth and $m$-strongly convex, with $L \in [L_-, L_+]$ and $m \in [m_-, m_+]$. Hence, assuming that it is not feasible to compute the smoothness and strong-convexity constants for each problem separately, the given class of functions is $L_+$-smooth and $m_-$-strongly convex. Therefore, we use \emph{heavy-ball with friction} (HBF) due to \citet{Polyak_1964} as baseline. Its update is given by
$\iter{x}{k+1} = \iter{x}{k} - \notation{\alpha} \nabla f (\iter{x}{k}) + \beta \left( \iter{x}{k} - \iter{x}{k-1} \right)$, where the optimal worst-case convergence rate is attained for $\notation{\alpha} = \Bigl( \frac{2}{\sqrt{L_+} + \sqrt{\mu_-}} \Bigr)^2$ and $\beta = \Bigl( \frac{\sqrt{L_+} - \sqrt{\mu_-}}{\sqrt{L_+} + \sqrt{\mu_-}} \Bigr)^2$ \citep{Nesterov_2018}. Further details can be found in Appendix~\ref{App:Implementation_details_quadratics}.
Figure~\ref{fig:experiments_quadratics} shows the results of this experiment: The upper left plot shows that the learned algorithm outperforms HBF by orders of magnitude and, despite being trained for $n_{\rm{train}} = 350$ iterations, one can use it until convergence. Here, the mean indicates that there are single instances for which instabilities occur, and, by comparing it to the median, one observes that the mean is far from being representative of the \say{typical} performance. Further, the algorithm performs well on very different orders of magnitude, ranging from about $1\mathrm{e}5$ to $1\mathrm{e}{-15}$. The upper right plot confirms that also in terms of computation time the learned algorithm is way faster than HBF, and the lower left plot shows that while the predicted PAC-bound is not tight, it still provides the guarantee to outperform HBF. Lastly, the lower right plot shows that the algorithm did satisfy the specified constraints $p_l$ and $p_u$ in all test cases.

\subsection{Image Processing}\label{Subsec:Image_Processing}

\begin{figure}[t!]
    \centering
    \includegraphics[width=\textwidth]{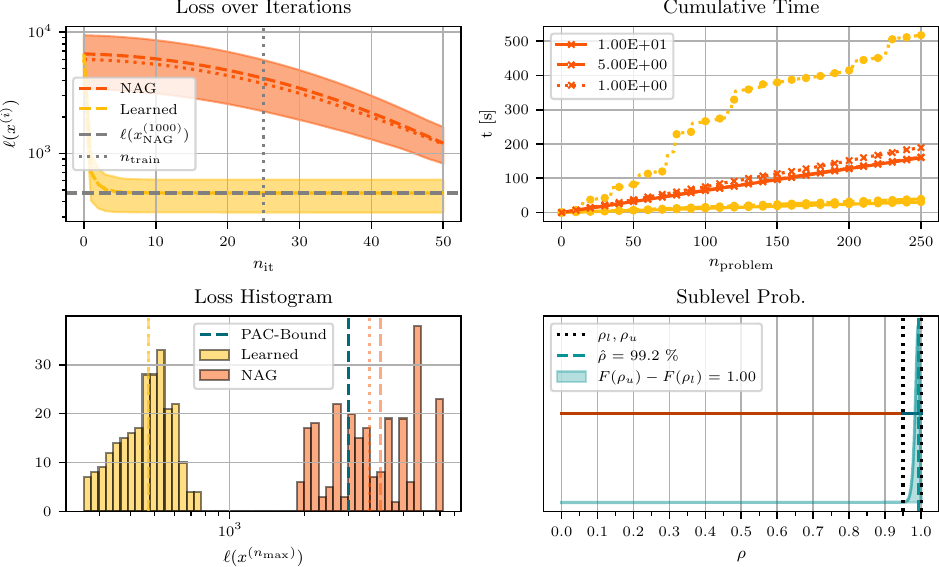}
    \caption{\textbf{Upper left:} Dashed lines represent mean losses, dotted lines show median losses, and the shaded regions represent the 10th to 90th percentile. The learned algorithm $\mathcal{A}$ is shown in yellow, while Nesterovs \emph{accelerated gradient descent} (NAG) is shown in orange. 
    \textbf{Upper right:} The different lines show the cumulative computation time the algorithms need to solve the test problems up to a certain accuracy (in function values) measured by $\ell(\iter{x}{i}, \notation{\realPar}) - \ell(x_{\rm{std}}^{(1000)}, \notation{\realPar}) < \varepsilon$. However, note that both algorithms are run for maximally $n_{\rm{max}} = 1000$ iterations. 
    \textbf{Lower left:} Loss histogram after $n_{\rm{train}} = 50$ iterations with predicted PAC-bound. 
    \textbf{Lower right:} The teal dashed line shows the point estimate for the sublevel probability, while the teal solid line shows the $\mathrm{Beta}$-posterior. The black dotted lines indicate the constraints $\sublevelprob_l, \sublevelprob_u$ and show the feasible region as dark teal line.}
    \label{fig:experiment_image_processing}
\end{figure}

We consider (gray-scale) \emph{image denoising/deblurring} with a regularizer given as smooth approximation to the $L_1$-norm of the image derivative, that is, problems of the form:
$$
    \min_{x \in \R^n} \frac{1}{2} \Vert Ax - b \Vert^2 + \lambda \sum_{i,j=1}^n \sqrt{(D_h x)_{i,j}^2 + (D_w x)_{i,j}^2 + \varepsilon^2} \qquad \lambda \in \R, \ A, D_h, D_w \in \R^{n \times n}, \ b \in \R^n \,.
$$
The matrix $A$ describes the \say{blurring} of the image, while $D_h$ and $D_w$ are the discrete image derivatives in h- and w-direction, respectively, which are used to penalize local changes in the image. We use images of height $N_h = 250$ and width $N_w = \texttt{int}(0.75 \cdot N_h) = 187$. Thus, the dimension $n$ of the optimization space is given by $n = 46750$. Further, as parameters $\notation{\realPar}$ we use the observed image and the regularization parameter, that is, $\notation{\realPar} = (b, \lambda) \in \R^{n+1} =: \notation{\parSpace}$.
Since the problem is smooth and convex, yet not strongly convex, the baseline algorithm is given by the \emph{accelerated gradient descent} (NAG) algorithm due to \citet{Nesterov_1983}. Its update is given by first computing $\iter{y}{k+1} = \iter{x}{k} + \frac{t_k - 1}{t_{k+1}} (\iter{x}{k} - \iter{x}{k-1})$ followed by setting $\iter{x}{k+1} = \iter{y}{k} - \notation{\alpha} \nabla f(\iter{y}{k+1})$. We use the optimal choices $t_{k+1} = \frac{1}{2}\left(1 + \sqrt{1 + 4 t_k^2}\right)$ and $\notation{\alpha} = \frac{1}{L}$. Here, the smoothness constant $L$ is given by the largest eigenvalue of $A^T A + \frac{\lambda}{\varepsilon} D^T D$, where $D \in \R^{2n \times n}$ is given by \say{stacking} $D_h$ and $D_w$, that is, $D = \begin{pmatrix} D_h & D_w \end{pmatrix}^T$. Further details can be found in Appendix~\ref{App:Implementation_details_image_processing}.
The results of this experiment are summarized in Figure~\ref{fig:experiment_image_processing}: The upper left plot shows that the learned algorithm is much faster than NAG in terms of the loss over the iterations, reaching a loss close to the ground-truth after only 5 iterations. The upper right plot confirms this finding also in terms of computation time. Yet, one can observe a strong increase in computation time for the dotted line (loss per pixel of about $\frac{1}{46750}$), indicating that the learned algorithm often is not able to reach this accuracy. In the lower left plot, one can observe that the predicted PAC-bound is not perfectly tight, yet provides the guarantee to outperform NAG. Finally, the lower right plot shows that, while the algorithm did not reach the sublevel set in all of the test cases, the probabilistically constraint optimization/sampling procedure did work correctly.

\subsection{Lasso-Problem}\label{Subsec:Lasso}

Here we consider the Lasso problem \citep{Tibshirani_1996}, that is, a non-smooth problem of the form:
$$
    \min_{x \in \R^n} \frac{1}{2} \Vert Ax - b \Vert_2^2 + 
    \lambda \Vert x \Vert_1 \qquad A \in \R^{\notation{m} \times n}, \ b \in \R^{\notation{m}} \,,
$$
with $\notation{m} \le n$. Thus, we are solving an underdetermined system of linear equations with an additional $\ell_1$-regularization term, which promotes sparsity in the solution \citep[see][]{Hastie_Tibshirani_Friedman_2009}. Hence, the optimization variable is given by $x \in \R^n$.
As baseline we use the \emph{fast iterative shrinkage-thresholding algorithm} (FISTA) by \citet{Beck_Teboulle_2009}, which performs an extrapolation step followed by a proximal gradient step, that is, abbreviating $h(x) := \frac{1}{2} \Vert Ax - b\Vert^2$ and $g(x) := \lambda \Vert x \Vert_1$, the update is given by first computing $\iter{y}{k} = \iter{x}{k} + \iter{\beta}{k} \left( \iter{x}{k} - \iter{x}{k-1} \right)$ followed by setting $\iter{x}{k+1} = \text{prox}_{\notation{\alpha} g} \left( \iter{y}{k} - \notation{\alpha} \nabla h(\iter{y}{k})\right)$. Here, the proximal mapping is defined as $\hat{x} = \text{prox}_{\notation{\alpha} g}(\Bar{x}) = \argmin_{x \in \R^n} \ \lambda \Vert x \Vert_1 + \frac{1}{2\notation{\alpha}} \Vert x - \Bar{x} \Vert_2^2$, and can be computed efficiently in closed-form yielding the so-called \emph{soft-thresholding operator}:
$$
    \hat{x}_i = \begin{cases}
        \Bar{x}_i - \notation{\alpha} \lambda \frac{\Bar{x}_i}{\abs{\Bar{x}_i}}\,, & \abs{\Bar{x}_i} > \notation{\alpha} \lambda \,; \\
        0 \,, &\text{otherwise} \,,
    \end{cases}\,, \quad i = 1,...,n \,.
$$
We choose $\notation{\alpha} = 1/L$, where $L$ is the largest eigenvalue of $A^T A$, that is, the smoothness parameter of $h$, while $\iter{\beta}{k}$ is set to $\iter{\beta}{k} := (t_k - 1)/t_{k+1}$ with $t_{k+1} = (1 + \sqrt{1 + 4 t_k^2})/2$. Further details about the experiment can be found in Appendix~\ref{App:Implementation_details_lasso}.
\begin{figure}[t!]
    \centering
    \includegraphics[width=\textwidth]{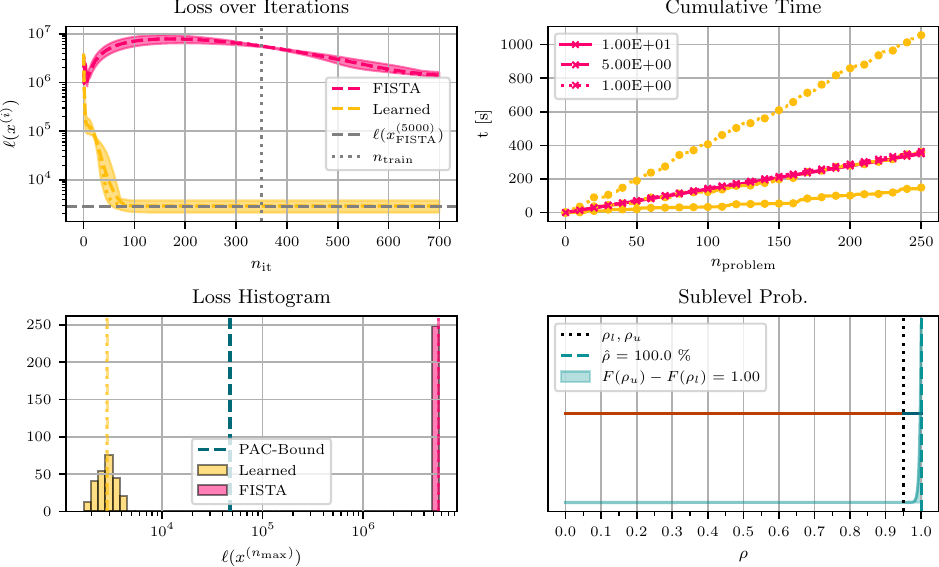}
    \caption{\textbf{Upper left:} Dashed lines represent the mean losses, dotted lines represent the median losses, and the shaded region represents the 10th to 90th percentile. Here, the \emph{fast iterative shrinkage-thresholding algorithm} (FISTA) is shown in pink and the learned algorithm in yellow. The gray horizontal lines represent the loss achieved by FISTA after 5000 iterations, which serves as approximation for the solution.  
    \textbf{Upper right:} The different lines show the cumulative computation time the algorithms need to solve all the test problems up to a certain accuracy (in function-values) measured by $\ell(\iter{x}{i}, \notation{\realPar}) - \ell(x_{\rm{std}}^{(5000)}, \notation{\realPar}) < \varepsilon$. However, note that both algorithms are run for maximally $n_{\rm{max}} = 5000$ iterations.
    \textbf{Lower left:} Loss histogram (after $n_{\rm{train}} = 350$ iterations) with the predicted PAC-bound.
    \textbf{Lower right:} The teal dashed line shows the point estimate for the sublevel probability, while the teal solid line shows the $\mathrm{Beta}$-posterior. The black dotted lines indicate the constraints $\sublevelprob_l, \sublevelprob_u$ and show the feasible region as dark teal line.}
    \label{fig:experiment_lasso}
\end{figure}
The results of this experiment are summarized in Figure~\ref{fig:experiment_lasso}: The upper left plot shows that the learned algorithm outperforms FISTA by several orders of magnitude, achieving a loss that is similar to the one of $\iter{x}{5000}_{\rm{FISTA}}$ after only 100 iterations, and one can observe that the learned algorithm can be used for more iterations than it was trained for. The upper right plot shows that, up to a certain accuracy, it is also way faster in terms of computation time. Yet, it seems that $\mathcal{A}$ does not reach arbitrary levels of accuracy. The lower left plot shows that the predicted PAC-bound is not perfectly tight, yet guarantees that $\mathcal{A}$ will outperform FISTA for the given number of iterations. And the lower right plot indicates that the algorithm did reach the sublevel set in all of the test cases.

\subsection{Training Neural Networks}\label{Subsec:Neural_Network_Training}

This experiment considers the problem of training a neural network on a regression problem, that is, $\mathcal{A}$ is trained to predict the parameters $\beta \in \R^{\notation{m}}$ of a neural network $\mathtt{N}_\beta$, which then is used to predict a function $g: \R \to \R$. Hence, the optimization variable is given by $\beta \in \R^{\notation{m}}$.
\begin{figure}[t!]
    \centering
    \includegraphics[width=\textwidth]{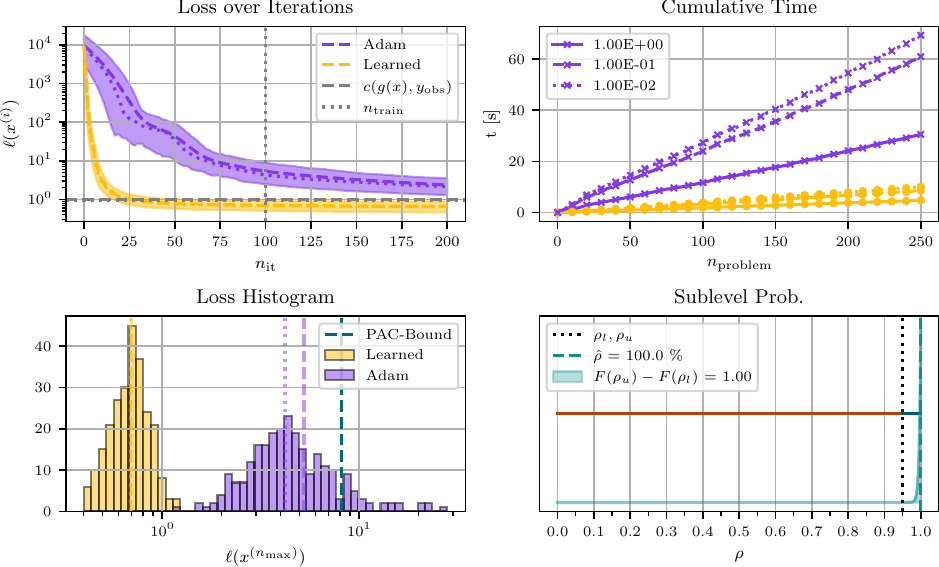}
    \caption{\textbf{Upper left:} Dashed lines represent the mean losses, dotted lines represent the median losses, and the shaded regions indicate the 10th to 90th percentile. The vertical dotted line shows $n_{\rm{train}}$, and the horizontal gray dashed line represents the average loss of the ground-truth function $g$ (equal to one, as we added standard Gaussian noise). Here, Adam is shown in purple and the learned algorithm in yellow. 
    \textbf{Upper right:} The different lines show the cumulative computation time of the algorithms to solve all the test problems up to a certain accuracy (in function values) measured by $\ell(\beta^{(i)}, \notation{\realPar}) - c(X_i, Y_i) < \varepsilon$. However, note that both algorithms are run for maximally $n_{\rm{max}} = 5000$ iterations. 
    \textbf{Lower left:} Loss histogram (after $n_{\rm{train}}$ iterations) and PAC-bound. 
    \textbf{Lower right:} The teal dashed line shows the point estimate for the sublevel probability, while the teal solid line shows the $\mathrm{Beta}$-posterior. Here, the black dotted lines indicate the constraints $\sublevelprob_l, \sublevelprob_u$ and show the feasible region as dark teal line.}
    \label{fig:neural_network_training_summary}
\end{figure}
As baseline we use Adam \citep{Kingma_Ba_2015} (as it is implemented in PyTorch), which is a widely used optimization algorithm for training neural networks. For tuning, we perform a grid search over 100 step-size parameters, such that its performance is best for the given $n_{\notation{\rm{train}}}$ iterations. Note that originally Adam was introduced for stochastic optimization, while we use it in the \say{full-batch setting} here. Further details can be found in Appendix~\ref{App:Implementation_details_nn_training}.
Figure~\ref{fig:neural_network_training_summary} shows the results of this experiment: The upper left plot shows that the learned algorithm clearly outperforms Adam, reaching the ground-truth loss after about 25 iterations, while Adam is not able to reach it within 200 iterations. Further, while the algorithm was trained for 100 iterations, it can be applied for more. The upper right plot confirms that, also in terms of computation time, $\mathcal{A}$ is way faster in training the neural network than Adam. The lower left plot shows that the predicted PAC-bound is not perfectly tight, yet yields a reasonable bound on the average performance, and guarantees to perform roughly as good as Adam (on average). The lower right plot indicates that the algorithm did reach the sublevel set in all test cases.

\section{Discussion and Limitations}

The motivation for this paper was to use more structure in a given problem than is analytically tractable. For this, we considered a distribution over parametric loss functions and formulated the (ultimate) goal in \eqref{Eq_random_parametric_opt_prob}, that is, to find a solution to each realization from this distribution. Under reasonable assumptions, this problem is too general to be solved. This led to the formulation of the performance of an algorithm in terms of its risk. However, since this is intractable, we derived PAC-Bayesian generalization bounds relating the risk to the empirically observable performance on a data set. This resulted in the formulation of a training objective, which relies heavily on the existence of a prior distribution satisfying our assumptions and yielding a good performance. As such a distribution is typically not known, we derived a procedure to construct it. This involved several key design choices, such as the loss-function, specific randomization steps, and, especially, the probabilistic constraints. Finally, we validated the resulting learning procedure on four practically relevant problems and showed that it yields a superior performance. While these experimental results are promising, we nevertheless see five main limitations of our work. First, the only guarantee that is provided by the PAC-Bayesian bound is an upper bound on the function value after a specified number of iterations. In particular, it does not guarantee that the function values, the iterates, or the gradient norm actually do converge. Second, our learning procedure is \emph{not guaranteed to work} and still involves many design choices. Third, one still has to find a good architecture for each given problem, which can be time-consuming. Fourth, the presented algorithmic procedure has a high computational cost (offline), which however, at least in part, is due to the nature of learning-to-optimize. Finally, the procedure only applies to deterministic algorithms. All these are promising directions of research that we leave to a future work.

\acks{M. Sucker and P. Ochs acknowledge funding by the German Research Foundation under Germany’s Excellence Strategy – EXC number 2064/1 – 390727645. Furthermore, J. Fadili and P. Ochs are supported by the ANR-DFG joint project TRINOM-DS under the numbers ANR-20-CE92-0037-01 and OC150/5-1.}


\newpage
\appendix

\section{Supplementary Definitions}\label{App:Missing_definitions}
\begin{Def}[Probability Kernel]\label{Def_Kernel}
    Let $\left(\genSpaceOne, \algebraGenSpaceOne\right)$, $\left(\genSpaceTwo, \algebraGenSpaceTwo\right)$ be measurable spaces. A function $\mu: \genSpaceOne \times \algebraGenSpaceTwo \to [0, \infty], \ (\genRealOne, \set{A}) \mapsto \mu (\genRealOne, \set{A})$ is called a \emph{kernel} from $\genSpaceOne$ to $\genSpaceTwo$, written as $\mu: \genSpaceOne \to \genSpaceTwo$, if for every set $\set{A} \in \algebraGenSpaceTwo$, the map $\genRealOne \mapsto \mu(\genRealOne, \set{A})$ is measurable, and for every point $\genRealOne \in \genSpaceOne$, the map $\set{A} \mapsto \mu(\genRealOne, \set{A})$ is a measure. Furthermore, $\mu$ is called a \emph{probability kernel} from $\genSpaceOne$ to $\genSpaceTwo$, if $\mu(\genRealOne, \genSpaceTwo) = 1$ for every $\genRealOne \in \genSpaceOne$.
\end{Def}

\begin{Def}[Exponential Family]\label{Def_Exponential_family}
    Let $\idxset$ be a non-empty index set. A family of probability measures $(\mathbb{Q}_{\idx})_{\idx \in \idxset}$ on a measurable space $\genSpaceOne$ is called an \emph{exponential family} (in $\natPar$ and $\suffStat$), if there is a dominating probability measure $\mu$, that is, $(\mathbb{Q}_{\idx})_{\idx \in \idxset} \subset \probMeasures(\mu)$, functions $\natPar: \idxset \to \R^k$, $\logPartition: \idxset \to (0, +\infty)$, and measurable functions $\suffStat: \genSpaceOne \to \R^k$, $\expFamFunc: \genSpaceOne \to (0, +\infty)$, such that for every $\idx \in \idxset$ we have 
    $\mathbb{Q}_{\idx} = \expFamFunc \logPartition(\idx) \exp\left(\sprod{\natPar(\idx)}{\suffStat}\right) \cdot \mu$, that is, $\mathbb{Q}_{\idx}[\set{B}] = \int_\set{B} \expFamFunc(\genRealOne) \logPartition(\idx) \exp(\sprod{\natPar(\idx)}{\suffStat(\genRealOne)}) \ \mu(d\genRealOne)$, $\set{B} \in \algebra{B}(\genSpaceOne)$.
\end{Def}

\begin{Def}[Support of a Measure]\label{Def_Support}
    Let $\genSpaceOne$ be a topological space, and let $\mu$ be a measure on $\genSpaceOne$. The \emph{support} of $\mu$ is defined as:
    $$
        \supp{\mu} := \{\genRealOne \in \genSpaceOne \ : \ \text{$\mu[\set{B}] > 0$ for every neighborhood $\set{B}$ of $\genRealOne$}\} \,.
    $$
\end{Def}

\section{Supplementary Lemmas}\label{App:Missing_Lemmas}
\begin{Lem}\label{Lem:Q_data_dependent}
    Under Assumption~\ref{Ass:exponential_family}, $\mathbb{Q}_{\idx}$ is a data-dependent distribution for every $\idx \in \idxset$.
\end{Lem}

\begin{proof}
    Denote the density of $\mathbb{Q}_{\idx}$ w.r.t. $\prob_{\notation{\rvHyp}}$ by $f_{\idx}(\notation{\realHyp, \realData}) := \frac{\expFamFunc(\notation{\realHyp})}{c(\idx, \notation{\realData})} \exp\left(\sprod{\natPar(\idx)}{\suffStat(\notation{\realHyp, \realData})}\right)$. The map $c(\idx, \cdot): \notation{\parSpace^N} \to [0, \infty)$ is $\algebra{B}(\notation{\parSpace^N})$ measurable, as $\suffStat$ is measurable w.r.t. the product-$\sigma$-algebra and $\prob_{\notation{\rvHyp}}$ is a finite measure \citep[Lemma 1.28, p.25]{Kallenberg_2021}. Hence, $f_{\idx}$ is measurable w.r.t. $\algebra{B}(\notation{\hypSpace}) \otimes \algebra{B}(\notation{\parSpace^N})$, since $c(\idx, \notation{\realData}) \in (0, \infty)$. Thus, it holds that $\mathbb{Q}_{\idx} = f_{\idx} \cdot \prob_{\notation{\rvHyp}}$ is a kernel from from $\notation{\parSpace^N}$ to $\notation{\hypSpace}$ \citep[Lem. 3.2, p.56]{Kallenberg_2021}. Finally, $\mathbb{Q}_{\idx}: \notation{\parSpace^N} \to \notation{\hypSpace}$ is actually a probability kernel, since $c(\idx, \notation{\realData})$ is the corresponding normalization constant. 
\end{proof}

\noindent
The following result states that non-negative random variables with finite second moment satisfy a one-sided sub-Gaussian inequality \citep[p.47]{Boucheron_Lugosi_Massart_2013}.
\begin{Lem}\label{Lemma_subgaussian_lower_tail}
    Let $\notation{\genRvOne}$ be a non-negative random variable with finite second moment. Then, for every $\idx > 0$ it holds $\mathbb{E}\left[ \exp \left( -\idx(\notation{\genRvOne} - \mathbb{E}[\notation{\genRvOne}]) \right) \right] \le \exp \left( \frac{\idx^2}{2} \mathbb{E}[\notation{\genRvOne}^2] \right)$.
\end{Lem}

\begin{Lem}\label{Lem:convergence_probability}
    The sublevel set $\sublevel{\sublevelfunction}$ is measurable.
\end{Lem}
\begin{proof}
    As $\sublevelfunction$ is assumed to be measurable, it suffices to show that the specific composition of $\ell$ and $\mathcal{A}$ is measurable, that is, $\ell \circ \mathcal{A}: \notation{\hypSpace} \times \notation{\parSpace} \to [0, +\infty], \ (\notation{\realHyp, \realPar}) \mapsto \ell(\mathcal{A}(\notation{\realHyp, \realPar}), \notation{\realPar})$ is measurable w.r.t. $\algebra{B}(\notation{\hypSpace}) \otimes \algebra{B}(\notation{\parSpace})$ and $\algebra{B}([0, +\infty])$. Since $\ell \ge 0$ is measurable, there exists a sequence of simple\footnote{A function is called simple, if it is of the form $\ell_n = \sum_{i=1}^K a_n^i \mathds{1}_{\set{A}_n^i}$ with disjoint sets $\set{A}_n^i$.} functions $\ell_n$ with $\ell = \lim_{n \to \infty} \ell_n$. Thus, since limits of measurable functions are measurable, it suffices to consider the case of a simple function $\ell: \R^n \times \notation{\parSpace} \to \R$. Then, however, it suffices to consider characteristic functions of the form $\mathds{1}_\set{A}$ for a measurable set $\set{A} \in \algebra{B}(\R^n) \otimes \algebra{B}(\notation{\parSpace})$. Since the product-$\sigma$-algebra is generated by cylinder sets, it actually suffices to consider the case $\ell = \mathds{1}_{\set{B} \times \set{D}}$, that is, $(\ell \circ \mathcal{A})(\notation{\realHyp, \realPar}) = \mathds{1}_{\set{B} \times \set{D}}(\mathcal{A}(\notation{\realHyp, \realPar}), \notation{\realPar}) = \mathds{1}_{\set{B}}(\mathcal{A}(\notation{\realHyp, \realPar})) \mathds{1}_\set{D}(\notation{\realPar})$. The second term is obviously measurable, and the first term is measurable as a composition of two measurable functions.
\end{proof}

\begin{Lem}\label{Lem:Sublevel_Risk}
    Suppose Assumption~\ref{Assumption_parametric_optimization_algorithm} holds, and let $\prob_{\notation{\rvPar}}[\sublevel{\sublevelfunction, \notation{\realHyp}}] > 0$ for every $\notation{\realHyp} \in \notation{\hypSpace}$. Then we have $\prob_{\notation{\rvHyp}}$-a.s.:
    \begin{itemize}
        \item[(i)] $\sublevelprob(\notation{\realHyp}) = \prob_{\notation{\rvPar}}[\sublevel{\sublevelfunction, \notation{\realHyp}}]$,
        \item[(ii)] $\expectation \left [ \ell(\notation{\rvHyp, \rvPar}) \cdot \mathds{1}_{\sublevel{\sublevelfunction}}(\notation{\rvHyp, \rvPar}) \ \vert \ \notation{\rvHyp = \realHyp} \right ] = \expectation_{\notation{\rvPar}}[\ell(\notation{\realHyp}, \cdot) \mathds{1}_{\sublevel{\sublevelfunction, \notation{\realHyp}}}] = \sublevelprob(\notation{\realHyp}) \cdot \expectation_{\notation{\rvPar}} [\ell(\notation{\realHyp}, \cdot) \ \vert \ \sublevel{\sublevelfunction, \notation{\realHyp}}] $.
    \end{itemize}
\end{Lem}
\begin{proof}
    By the independence of $\notation{\rvPar}$ and $\notation{\rvHyp}$, we have $\expectation \left \{ \ell(\notation{\rvHyp, \rvPar}) \mathds{1}_{\sublevel{\sublevelfunction}}(\notation{\rvHyp, \rvPar}) \ \vert \ \notation{\rvHyp = \realHyp} \right \} = \int_{\notation{\parSpace}} \ell(\notation{\realHyp, \realPar}) \mathds{1}_{\sublevel{\sublevelfunction, \notation{\realHyp}}}(\notation{\realPar}) \ \prob_{\notation{\rvPar}}(d\notation{\realPar}) = \expectation_{\notation{\rvPar}} [\ell(\notation{\realHyp}, \cdot) \mathds{1}_{\sublevel{\sublevelfunction, \notation{\realHyp}}}]$ $\prob_{\notation{\rvHyp}}$-a.s., which shows the first equality of (ii). Since $\prob_{\notation{\rvPar}} [\sublevel{\sublevelfunction, \notation{\realHyp}}] > 0$, the elementary conditional expectation is defined as $\expectation_{\notation{\rvPar}} [\ell(\notation{\realHyp}, \cdot) \ \vert \ \sublevel{\sublevelfunction, \notation{\realHyp}}] = \frac{\expectation_{\notation{\rvPar}}[\ell(\notation{\realHyp}, \cdot) \mathds{1}_{\sublevel{\sublevelfunction, \notation{\realHyp}}}]}{\prob_{\notation{\rvPar}}[\sublevel{\sublevelfunction, \notation{\realHyp}}]}$. Again by independence we have $\prob_{\notation{\rvHyp}}$-a.s. the equality $\prob_{\notation{\rvPar}}[\sublevel{\sublevelfunction, \notation{\realHyp}}] = \prob_{\notation{\rvPar \vert \rvHyp = \realHyp}}[\sublevel{\sublevelfunction, \notation{\realHyp}}] = \sublevelprob(\notation{\realHyp})$, which shows (i) and the second equality of (ii). 
\end{proof}

\section{Proof of Lemma~\ref{Lem_DonskerVaradhanForExponentialFamilies}}\label{App:Proof_Donsker_Varadhan}
\begin{proof}
Take any $\idx \in \idxset$ and $\notation{\realData \in \parSpace^N}$. First, let $\mathbb{Q} \in \probMeasures(\mathbb{P}_{\notation{\rvHyp}})$  be arbitrary. By the Radon-Nikodym theorem, there exists a measurable function $f \ge 0$, s.t. $\mathbb{Q} = f \cdot \prob_{\notation{\rvHyp}}$. Since the convention $0 \cdot \infty = 0$ applies throughout measure theory, one has:
\begin{align*}
    \divergence{\rm{KL}}{\mathbb{Q}}{\prob_{\notation{\rvHyp}}} 
    &= \mathbb{Q} \left[\log(f) \right] 
    = \prob_{\notation{\rvHyp}} \left[f \log(f)\right] 
    = \prob_{\notation{\rvHyp}} \left[\mathds{1}_{\{f > 0\}} f \log(\mathds{1}_{\{f > 0\}} f)\right] \,.
\end{align*}
Hence, w.l.o.g. we can assume that $f > 0 \ \ \prob_{\notation{\rvHyp}}$-a.s. Then, by Jensen's inequality, one gets:
\begin{align*}
    &\mathbb{Q}[\sprod{\natPar(\idx)}{\suffStat(\cdot, \notation{\realData})} + \log(\expFamFunc)] - \mathbb{Q}[\log(f)]
    = \mathbb{Q}\left[\log\left(\frac{\expFamFunc}{f} \exp\left(\sprod{\natPar(\idx)}{\suffStat(\cdot, \notation{\realData})}\right)\right)\right] \\
    &\le \log \left( \mathbb{Q} \left[ \frac{\expFamFunc}{f} \exp\left(\sprod{\natPar(\idx)}{\suffStat(\cdot, \notation{\realData})}\right) \right] \right) 
    = \log \left( (f \cdot \prob_{\notation{\rvHyp}}) \left[ \frac{\expFamFunc}{f} \exp\left(\sprod{\natPar(\idx)}{\suffStat(\cdot, \notation{\realData})}\right) \right] \right) \\
    &= \log \left( \prob_{\notation{\rvHyp}} \left[ \expFamFunc \exp\left(\sprod{\natPar(\idx)}{\suffStat(\cdot, \notation{\realData})}\right) \right] \right) = \kappa(\idx, \notation{\realData}) \,.
\end{align*}
Hence, we have $\kappa(\idx, \notation{\realData}) \ge \mathbb{Q}[\sprod{\natPar(\idx)}{\suffStat(\cdot, \notation{\realData})} + \log(\expFamFunc)] - \divergence{\rm{KL}}{\mathbb{Q}}{\prob_{\notation{\rvHyp}}}$ for any probability measure $\mathbb{Q} \ll \prob_{\notation{\rvHyp}}$. Now consider the member of the exponential family:
\begin{align*}
    \divergence{\rm{KL}}{\mathbb{Q}_{\idx}(\notation{\realData})}{\mathbb{P}_{\notation{\rvHyp}}} 
    &= \int_{\notation{\hypSpace}} \log\left( \expFamFunc(\notation{\realHyp}) \logPartition(\idx, \notation{\realData}) \exp(\sprod{\natPar(\idx)}{\suffStat(\notation{\realHyp, \realData})})\right) \ \mathbb{Q}_{\idx}(\notation{\realData}, d\notation{\realHyp}) \\
    &= \int_{\notation{\hypSpace}} \log (\expFamFunc(\notation{\realHyp})) + \sprod{\natPar(\idx)}{\suffStat(\notation{\realHyp, \realData})} \ \mathbb{Q}_{\idx}(\notation{\realData}, d\notation{\realHyp}) - \log\left(c(\idx, \notation{\realData}) \right) \\
    &= \int_{\notation{\hypSpace}} \log (\expFamFunc(\notation{\realHyp})) + \sprod{\natPar(\idx)}{\suffStat(\notation{\realHyp, \realData})} \ \mathbb{Q}_{\idx}(\notation{\realData}, d\notation{\realHyp}) - \kappa(\idx, \notation{\realData}).
\end{align*}
Rearranging yields $\kappa(\idx, \notation{\realData}) = \int_{\notation{\hypSpace}} \log (\expFamFunc(\notation{\realHyp})) + \sprod{\natPar(\idx)}{\suffStat(\notation{\realHyp, \realData})} \ \mathbb{Q}_{\idx}(\notation{\realData}, d\notation{\realHyp}) - \divergence{\rm{KL}}{\mathbb{Q}_{\idx}(\realData)}{\mathbb{P}_{\notation{\rvHyp}}}$.
\end{proof}

\section{Proof of Lemma~\ref{Lem:Compact_lambda}}\label{App:Proof_Compact_Lambda}
\begin{proof}
    W.l.o.g. assume that $\set{O}_i \neq \emptyset$ and choose $\idx_i \in \set{O}_i$, $i = 1, ..., \coveringNumber$. Then, for every $\notation{\realData \in \parSpace^N}$, it holds that:
    \begin{align*}
        \sup_{\idx \in \idxset} \kappa(\idx, \notation{\realData}) 
        &\le \max_{i=1,...,\coveringNumber} \ \sup_{\idx \in \set{O}_i} \kappa(\idx, \notation{\realData}) 
        = \max_{i=1,...,\coveringNumber} \left\{\kappa(\idx_i, \notation{\realData}) + \sup_{\idx \in \set{O}_i} \left(\kappa(\idx, \notation{\realData}) - \kappa(\idx_i, \notation{\realData}) \right)  \right\} \\
        &\le \max_{i=1,...,\coveringNumber} \kappa(\idx_i, \notation{\realData}) + \coveringConstant \,.
    \end{align*}
    Thus, in total one gets for $\notation{t} \in \mathbb{R}$:
    \begin{align*}
        \prob \left\{ \sup_{\idx \in \idxset} \kappa(\idx, \notation{\rvData}) > \notation{t} \right\} 
        &\le \prob \left\{ \max_{i=1,...,\coveringNumber} \kappa(\idx_i, \notation{\rvData}) + \coveringConstant > \notation{t} \right\}  
        \le \sum_{i=1}^{\coveringNumber} \prob \left\{ \kappa(\idx_i, \notation{\rvData}) + \coveringConstant > \notation{t} \right\} \\
        &\le \sum_{i=1}^{\coveringNumber} \exp(\coveringConstant-\notation{t}) 
        = \coveringNumber \exp(\coveringConstant-\notation{t}) \,.
    \end{align*}
    Taking $\notation{t} = \log\left( \frac{\coveringNumber}{\epsilon} \right) + \coveringConstant$ gives $\mathbb{P} \left\{ \sup_{\idx \in \idxset} \kappa(\idx, \notation{\rvData}) > \log\left( \frac{\coveringNumber}{\epsilon} \right) + \coveringConstant \right\} \le \epsilon$.
\end{proof}

\section{Proof of Corollary~\ref{Cor_generalization_bound_exp_family}}\label{App:Proof_Cor_generalization_bound}
\begin{proof}
    The two forumlas are simply rewritings of each other: By assumption, bilinearity and definition of the euclidean scalar product, and linearity of the integral, the term $\mathbb{Q}[\sprod{\natPar(\idx)}{\suffStat(\cdot, \notation{\realData})}]$ can be split up as:
    \begin{align*}
        \mathbb{Q}[\sprod{\natPar(\idx)}{\suffStat(\cdot, \notation{\realData})}] 
        &= \mathbb{Q}[\natPar^{(1)}(\idx) (\mathcal{R} - \hat{\mathcal{R}}(\cdot, \notation{\realData}))] +
        \mathbb{Q}[\sprod{\natPar^{(r)}(\idx)}{\suffStat^{(r)}(\cdot, \notation{\realData})}] \\
        &= \natPar^{(1)}(\idx) \mathbb{Q}[\mathcal{R}] - \natPar^{(1)}(\idx) \mathbb{Q}[\hat{\mathcal{R}}(\cdot, \notation{\realData})] + 
        \mathbb{Q}[\sprod{\natPar^{(r)}(\idx)}{\suffStat^{(r)}(\cdot, \notation{\realData})}] \,.
    \end{align*}
    Simply rearranging the terms then yields the result, as $\natPar^{(1)} > 0$.
\end{proof}

\section{Implementation Details}\label{App:Implementation_details}
We use the following training procedure in all experiments: $N = N_{\rm{prior}} + N_{\rm{train}} + N_{\rm{val}} + N_{\rm{test}}$ denotes the total number of datapoints, and we use $N_{\rm{prior}} = ... = N_{\rm{test}} = 250$. (Sub)Gradients are defined by the output of backpropagation as it is implemented in PyTorch \citep{PyTorch_2019}, and we use $g(\notation{\realPar}) := \notation{\alpha} \ell(\iter{x}{0}, \notation{\realPar})^{\notation{\beta}}$, $\notation{\alpha}, \notation{\beta} > 0$, to define the sublevel set $\sublevel{\sublevelfunction}$. In Algorithm~\ref{Alg:Estimation_Probabilistic_Constraint}, we use $\sublevelprob_l = 0.95$, $\sublevelprob_u = 1.0$, $\quantile_l = 0.01$, $\quantile_u = 0.99$, and $\varepsilon = 0.075$. Thus, the algorithm should reach $\sublevel{\sublevelfunction}$ in at least 95\% of the cases, and for the estimation of the sublevel probability it should concentrate 98\% of the mass within a distance of 0.075. In Algorithm~\ref{Alg:Probabilistically_Constrained_Sampling}, we use \emph{stochastic gradient Langevin dynamics} to draw $10^2$ samples, where we decay the step-size starting from $10^{-6}$. In Algorithm~\ref{Alg:Finding_Initialization}, we use Adam with an initial step-size of $10^{-3}$, which gets reduced by a factor of 0.5 every 200 iterations, until an accuracy of $\varepsilon = 10^{-2}$ is reached, or for at most $n_{\rm{init}} = 10^3$ iterations. In Algorithm~\ref{Alg:Training_Procedure}, we use Adam with an initial step-size of $10^{-4}$, which gets reduced by a factor of 0.5 every $2 \cdot 10^4$ iterations, for a total of $n_{\rm{max}} = 2 \cdot 10^5$ iterations. We use a trajectory length of $\trajectorylength = 1$, that is, only single points, and update the constraint only every $2 \cdot 10^4$ iterations (with a reset to previous iterates, if we have left the set $\hat{\set{A}}$). In Algorithm~\ref{Alg:Construction_Posterior}, we use a finite $\idxset$ with $\vert \idxset \vert = 75 \cdot 10^3$, and an accuracy (of the PAC-bound) of $\varepsilon = 0.05$. 
As we contrast the learned algorithm to first-order methods, in each iteration $\mathcal{A}$ has access to iterates, (sub)gradients, and function values, and the update is solely based on these. Here, we perform preprocessing: The (sub)gradient is split into its norm $\norm{\nabla \ell(\iter{x}{k}, \realPar)}{}$ and the corresponding unit vector $\iter{d}{k}_1$. Further, the norm is transformed to $\iter{n}{k}_1 := \log(1 + \norm{\nabla \ell(\iter{x}{k}, \realPar)}{})$ to be less scale-sensitive. The iterates $\iter{x}{k}, \iter{x}{k-1}$ are combined into the momentum term $\iter{m}{k} := \iter{x}{k} - \iter{x}{k-1}$, which also is split into the unit vector $\iter{d}{k}_2$ and the transformed norm $\iter{n}{k}_2$. Similarly, we also transform the function values into $\iter{\ell}{k}_1 = \log(1 + \ell(\iter{x}{k}, \realPar))$ and $\iter{\ell}{k}_2 = \log(1 + \ell(\iter{x}{k-1}, \realPar))$.

\begin{Rem}
    \begin{itemize}
        \item[(i)] We always use the output of the backpropagation algorithm instead of exact (sub-)gradients, that is, the learned algorithms \emph{do not rely on smoothness}.
        \item[(ii)] We use 100 samples only, as they are very costly: To evaluate the potentials $\varphi_{\rm{prior}}$ and $\varphi_{\idx}(\cdot, \notation{\realData})$ on a single sample $\notation{\realHyp} \in \notation{\hypSpace}$, one has to compute all losses $\ell(\notation{\realHyp, \realPar_i})$, $i=1,...,N_{\rm{prior}} + N_{\rm{train}}$, that is, \say{solving} $N_{\rm{prior}} + N_{\rm{train}}$ optimization problems.
    \end{itemize}
\end{Rem}

\subsection{Details for the Experiment on Quadratic Functions}\label{App:Implementation_details_quadratics}
This subsection describes the missing details for the experiment on quadratic functions.

\subsubsection{Construction of the Parameters} 
To control the strong-convexity and smoothness of $\ell$, we specify intervals $[m_-, m_+], [L_-, L_+] \subset (0, +\infty)$, and sample $m_1, ..., m_N \overset{iid}{\sim} \mathrm{U}_{[m_-, m_+]}$, $L_1, ..., L_N \overset{iid}{\sim} \mathrm{U}_{[L_-, L_+]}$. Then, the matrices $A_j$, $j = 1, ..., N$, are created as \emph{diagonal matrices} with entries $a_{ii}^j = \sqrt{m_j} + i \cdot \frac{\sqrt{L_j} - \sqrt{m_j}}{n}$, $i = 1,...,n$, that is, we linearly interpolate from $\sqrt{m_j}$ to $\sqrt{L_j}$. Hence, the matrix $A_j^T A_j$ has smallest and largest eigenvalue $m_j$ and $L_j$, respectively. To change the relative position between the ellipsoid of the quadratic and the initialization, we randomize the right-hand side by sampling $b_1, ..., b_N \overset{iid}{\sim} \mathcal{N}(\mu, \Sigma)$, where we create $\mu$ and $\Sigma = C^T C$ by sampling $\mu_i, C_{i,k} \overset{iid}{\sim} \mathrm{U}_{[-5, 5]}$, $i,k = 1,...,n$.

\begin{figure}[t!]
    \centering
        \begin{tikzpicture}
        \tikzset{conv/.style={black,draw=black,fill=test_1,rectangle,minimum height=0.5cm}}
        \tikzset{linear/.style={black,draw=black,fill=test_2,rectangle,minimum height=0.5cm}}
        \tikzset{relu/.style={black,draw=black,fill=test_3,rectangle,minimum height=0.25cm}}

        \node (d1) at (0, 1.0) {\tiny{$\iter{d}{k}_1$}};
        \node (d2) at (0, 0.0) {\tiny{$\iter{d}{k}_2$}};
        \node (d3) at (-0.4, -1.0) {\tiny{$\iter{d}{k}_1 \odot \iter{d}{k}_2$}};

        \node[conv,rotate=90,minimum width=2.75cm] (conv1) at (2,0) {\tiny{\texttt{Conv2d(3,16,1,bias=F)}}};
        \node[relu,rotate=90,minimum width=2.75cm] (relu1) at (2.5,0) {\tiny{\texttt{ReLU}}};
        
        \node[conv,rotate=90,minimum width=2.75cm] (conv2) at (3,0) {\tiny{\texttt{Conv2d(16,16,1,bias=F)}}};
        \node[conv,rotate=90,minimum width=2.75cm] (conv3) at (3.5,0) {\tiny{\texttt{Conv2d(16,16,1,bias=F)}}};
        \node[relu,rotate=90,minimum width=2.75cm] (relu2) at (4,0) {\tiny{\texttt{ReLU}}};

        \node[conv,rotate=90,minimum width=2.75cm] (conv4) at (4.5,0) {\tiny{\texttt{Conv2d(16,16,1,bias=F)}}};
        \node[conv,rotate=90,minimum width=2.75cm] (conv5) at (5,0) {\tiny{\texttt{Conv2d(16,16,1,bias=F)}}};
        \node[relu,rotate=90,minimum width=2.75cm] (relu3) at (5.5,0) {\tiny{\texttt{ReLU}}};

        \node[conv,rotate=90,minimum width=2.75cm] (conv6) at (6,0) {\tiny{\texttt{Conv2d(16,1,1,bias=F)}}};

        \node (d) at (7, 0) {\tiny{$\iter{d}{k}$}};

        \draw[->] (d1) to [out=0, in=180] (conv1.north);
        \draw[->] (d2) -- (conv1);
        \draw[->] (d3) to [out=0, in=180] (conv1.north);

        \draw[-] (conv1) -- (relu1) -- (conv2) -- (conv3) -- (relu2) -- (conv4) -- (conv5) -- (relu3) -- (conv6);
        \draw[->] (conv6) -- (d);

        \node (n1) at (0, -1.6) {\tiny{$\iter{n}{k}_1$}};
        \node (n2) at (0, -2.3) {\tiny{$\iter{n}{k}_2$}};
        \node (l1) at (0, -2.9) {\tiny{$\iter{\ell}{k}_1$}};
        \node (l2) at (0, -3.6) {\tiny{$\iter{\ell}{k}_2$}};

        \node[linear,rotate=90,minimum width=2.3cm] (lin1) at (2, -2.6) {\tiny{\texttt{Linear(4,8,bias=F)}}};
        \node[relu,rotate=90,minimum width=2.3cm] (relu2_1) at (2.5, -2.6) {\tiny{\texttt{ReLU}}};

        \node[linear,rotate=90,minimum width=2.3cm] (lin2) at (3, -2.6) {\tiny{\texttt{Linear(8,8,bias=F)}}};
        \node[linear,rotate=90,minimum width=2.3cm] (lin3) at (3.5, -2.6) {\tiny{\texttt{Linear(8,8,bias=F)}}};
        \node[relu,rotate=90,minimum width=2.3cm] (relu2_2) at (4.0, -2.6) {\tiny{\texttt{ReLU}}};

        \node[linear,rotate=90,minimum width=2.3cm] (lin4) at (4.5, -2.6) {\tiny{\texttt{Linear(8,8,bias=F)}}};
        \node[linear,rotate=90,minimum width=2.3cm] (lin5) at (5.0, -2.6) {\tiny{\texttt{Linear(8,8,bias=F)}}};
        \node[relu,rotate=90,minimum width=2.3cm] (relu2_3) at (5.5, -2.6) {\tiny{\texttt{ReLU}}};

        \node[linear,rotate=90,minimum width=2.3cm] (lin6) at (6, -2.6) {\tiny{\texttt{Linear(8,1,bias=F)}}};

        \node (s) at (7, -2.6) {\tiny{$\iter{\notation{\alpha}}{k}$}};

        \draw[->] (n1) to [out=0, in=180] (lin1.north);
        \draw[->] (n2) to [out=0, in=180] (lin1.north);
        \draw[->] (l1) to [out=0, in=180] (lin1.north);
        \draw[->] (l2) to [out=0, in=180] (lin1.north);

        \draw[-] (lin1) -- (relu2_1) -- (lin2) -- (lin3) -- (relu2_2) -- (lin4) -- (lin5) -- (relu2_3) -- (lin6);
        \draw[->] (lin6) -- (s);

        \node (xk) at (9, 0) {\tiny{$\iter{x}{k}$}};
        \node (new) at (9, -1.3) {\tiny{$\iter{x}{k+1} := \iter{x}{k} + \iter{\notation{\alpha}}{k} \cdot \iter{d}{k}$}};
        \draw[->] (d) to [out=0, in=90] (new.north);
        \draw[->] (xk) -- (new);
        \draw[->] (s) to [out=0, in=270] (new.south);
        
    \end{tikzpicture}
    \caption{Update step of $\mathcal{A}$ for quadratic problems: The directions $\iter{d}{k}_1$, $\iter{d}{k}_2$ and $\iter{d}{k}_1 \odot \iter{d}{k}_2$ are inserted as different channels into the \texttt{Conv2d}-block, which performs $1 \times 1$ \say{convolutions}, that is, the algorithm acts coordinate-wise on the input, and yields an new update-direction $\iter{d}{k}$. The scales $\iter{n}{1}$, $\iter{n}{2}$, and the function values $\iter{\ell}{k}_1$, $\iter{\ell}{k}_2$ get transformed separately by the fully-connected block to yield the step-size $\iter{\notation{\alpha}}{k}$.}
    \label{architecture:quadratics}
\end{figure}

\subsubsection{Algorithm} The algorithmic update of the learned algorithm $\mathcal{A}$ is visualized in Figure~\ref{architecture:quadratics} and consists of two blocks: The update-block combines the gradient direction $\iter{d}{k}_1$, the momentum direction $\iter{d}{k}_2$, and their \say{interaction} $\iter{d}{k}_1 \odot \iter{d}{k}_2$ into the new update-direction $\iter{d}{k}$, while the other block computes a step-size based on the corresponding logarithmically transformed norms $\iter{n}{k}_1$ and $\iter{n}{k}_2$, and the logarithmically transformed function values $\iter{\ell}{k}_1$ and $\iter{\ell}{k}_2$. 

\subsection{Details for the Image-Processing Experiment}\label{App:Implementation_details_image_processing}
This subsection describes the missing details for the image-processing experiment.

\subsubsection{Construction of the Parameters} Throughout, we use $\varepsilon = 0.01$. For computational efficiency, the matrices $A, D_h, D_w$ are implemented through the convolution of the image $x$ with a corresponding kernel (with reflective boundary conditions). For $A$, we use a standard $(5 \times 5)$-Gaussian kernel, while $D_h$ and $D_w$ are given through the kernels:
$$
    k_h = \begin{pmatrix}
        0 & 0 & 0 \\ 0 & -1 & 0 \\ 0 & 1 & 0
    \end{pmatrix} \in \R^{3 \times 3}
    \qquad \text{and} \qquad 
    k_w = \begin{pmatrix}
        0 & 0 & 0 \\ 0 & -1 & 1 \\ 0 & 0 & 0
    \end{pmatrix} \in \R^{3 \times 3} \,.
$$
Additionally, after blurring an image with $A$, we add centered Gaussian noise $\varepsilon_{i,j}$ with standard deviation $\sigma = \frac{25}{256}$ to each pixel, that is, $b_{i,j} = (Ax^*)_{i,j} + \varepsilon_{i,j}$ with $\varepsilon_{i,j} \overset{iid}{\sim} \mathcal{N} \left( 0, \sigma \right)$, $i=1,...,N_h$, $j = 1,...,N_w$. The regularization parameters $\lambda_i \in \R$, $i = 1,...,N$, are given by sampling uniformly, that is, $\lambda_i \overset{iid}{\sim} \mathrm{U}_{[\lambda_-, \lambda_+]}$, where we use $\lambda_- =0.05$ and $\lambda_+ = 0.5$.

\begin{figure}[t!]
    \centering
        \begin{tikzpicture}
        \tikzset{conv/.style={black,draw=black,fill=test_1,rectangle,minimum height=0.5cm}}
        \tikzset{linear/.style={black,draw=black,fill=test_2,rectangle,minimum height=0.5cm}}
        \tikzset{relu/.style={black,draw=black,fill=test_3,rectangle,minimum height=0.25cm}}

        \node (d1) at (0, 1.0) {\tiny{$\iter{d}{k}_1$}};
        \node (d2) at (0, 0.0) {\tiny{$\iter{d}{k}_2$}};
        \node (d3) at (-0.4, -1.0) {\tiny{$\iter{d}{k}_1 \odot \iter{d}{k}_2$}};

        \node[conv,rotate=90,minimum width=2.75cm] (conv1) at (2,0) {\tiny{\texttt{Conv2d(3,16,3,bias=F)}}};
        \node[relu,rotate=90,minimum width=2.75cm] (relu1) at (2.5,0) {\tiny{\texttt{ReLU}}};
        
        \node[conv,rotate=90,minimum width=2.75cm] (conv2) at (3,0) {\tiny{\texttt{Conv2d(16,16,3,bias=F)}}};
        \node[relu,rotate=90,minimum width=2.75cm] (relu2) at (3.5,0) {\tiny{\texttt{ReLU}}};

        \node[conv,rotate=90,minimum width=2.75cm] (conv3) at (4,0) {\tiny{\texttt{Conv2d(16,1,3,bias=F)}}};

        \node (d) at (6, 0) {\tiny{$\iter{d}{k}$}};

        \draw[->] (d1) to [out=0, in=180] (conv1.north);
        \draw[->] (d2) to [out=0, in=180] (conv1.north);
        \draw[->] (d3) to [out=0, in=180] (conv1.north);

        \draw[-] (conv1) -- (relu1) -- (conv2) -- (relu2) -- (conv3);
        \draw[->] (conv3) -- (d);

        \node (n1) at (0, -1.9) {\tiny{$\iter{n}{k}_1$}};
        \node (n2) at (0, -3.2) {\tiny{$\iter{n}{k}_2$}};

        \node[linear,rotate=90,minimum width=2.3cm] (lin1) at (2, -2.6) {\tiny{\texttt{Linear(2,8,bias=F)}}};
        \node[relu,rotate=90,minimum width=2.3cm] (relu2_1) at (2.5, -2.6) {\tiny{\texttt{ReLU}}};

        \node[linear,rotate=90,minimum width=2.3cm] (lin2) at (3, -2.6) {\tiny{\texttt{Linear(8,8,bias=F)}}};
        \node[relu,rotate=90,minimum width=2.3cm] (relu2_2) at (3.5, -2.6) {\tiny{\texttt{ReLU}}};

        \node[linear,rotate=90,minimum width=2.3cm] (lin3) at (4, -2.6) {\tiny{\texttt{Linear(8,1,bias=F)}}};

        \node (s) at (6, -2.6) {\tiny{$\iter{\notation{\alpha}}{k}$}};

        \draw[->] (n1) to [out=0, in=180] (lin1.north);
        \draw[->] (n2) to [out=0, in=180] (lin1.north);

        \draw[-] (lin1) -- (relu2_1) -- (lin2) -- (relu2_2) -- (lin3);
        \draw[->] (lin3) -- (s);

        \node (xk) at (8, 0) {\tiny{$\iter{x}{k}$}};
        \node (new) at (8, -1.3) {\tiny{$\iter{x}{k+1} := \iter{x}{k} + \iter{\notation{\alpha}}{k} \cdot \iter{d}{k}$}};
        \draw[->] (d) to [out=0, in=90] (new.north);
        \draw[->] (xk) -- (new);
        \draw[->] (s) to [out=0, in=270] (new.south);
        
    \end{tikzpicture}
    \caption{Update step of $\mathcal{A}$ for the image-processing problems: The directions $\iter{d}{k}_1$, $\iter{d}{k}_2$ and $\iter{d}{k}_1 \odot \iter{d}{k}_2$ are inserted as different channels (in the shape of the image) into the \texttt{Conv2d}-block, which performs a $3 \times 3$-convolution. The scales $\iter{n}{1}$, $\iter{n}{2}$ get transformed separately by the fully-connected block.}
    \label{architecture:image_processing}
\end{figure}

\subsubsection{Algorithm} The algorithmic update of $\mathcal{A}$ is visualized in Figure~\ref{architecture:image_processing} and consists of an update-block, which combines $\iter{d}{k}_1$, $\iter{d}{k}_2$ and their \say{interaction} $\iter{d}{k}_1 \odot \iter{d}{k}_2$ into the new update direction $\iter{d}{k}$, and a block to compute a step-size from the norms of the gradient and momentum term. Note that we use $3\times3$-convolutions this time, that is, we incorporate the knowledge about an image-processing problem into the design of the optimization algorithm.

\subsection{Details for the LASSO Experiment}\label{App:Implementation_details_lasso}
This subsection describes the missing details for the LASSO experiment.

\subsubsection{Construction of the Parameters}
The same matrix $A \in \R^{p \times n}$ with dimensions $n = 350$ and $\notation{m} = 70$ is used for all problem instances. Here, we sample each entry uniformly, that is, $a_{i,j} \overset{iid}{\sim} \mathrm{U}_{[-0.5, 0.5]}$, $i=1,...,\notation{m}$, $j = 1,...,n$. Thus, the parameters $\notation{\realPar}$ are given by the right-hand side and the regularization parameter, that is, $\notation{\realPar} = (b, \lambda) \in \R^{\notation{m} + 1} =: \notation{\parSpace}$. For this, the regularization parameter $\lambda$ is also sampled uniformly, that is, $\lambda_i \overset{iid}{\sim} \mathrm{U}_{[\lambda_-, \lambda_+]}$, $i = 1,...,N$, with $\lambda_- = 5$ and $\lambda_+ = 10$, while the right-hand side is sampled from a multivariate normal distribution, that is, $b_i \overset{iid}{\sim} \mathcal{N}(\mu, \Sigma)$, $i=1,...,N$, where we first construct $\mu$ and $\Sigma = C^T C$ by sampling each entry of $\mu$ and $C$ uniformly at random in $[-5, 5]$.

\begin{figure}[h!]
    \centering
    \begin{tikzpicture}
        \tikzset{conv/.style={black,draw=black,fill=test_1,rectangle,minimum height=0.5cm}}
        \tikzset{linear/.style={black,draw=black,fill=test_2,rectangle,minimum height=0.5cm}}
        \tikzset{relu/.style={black,draw=black,fill=test_3,rectangle,minimum height=0.25cm}}

        \node (gn_1) at (-4.05, 2.2) {\tiny{$\iter{n}{k}_{1,\neq}$}};
        \node (gn_2) at (-4, 1.8) {\tiny{$\iter{n}{k}_{1,0}$}};
        \node (mn_1) at (-4.05, 1.4) {\tiny{$\iter{n}{k}_{2,\neq}$}};
        \node (mn_2) at (-4, 1.0) {\tiny{$\iter{n}{k}_{2,0}$}};
        \node (sn_1) at (-4, 0.6) {\tiny{$\iter{n}{k}_{3,\neq}$}};
        \node (sn_2) at (-3.95, 0.2) {\tiny{$\iter{n}{k}_{3,0}$}};
        \node (ld) at (-4, -0.2) {\tiny{$\iter{\Delta \ell}{k}$}};
        \node (ldns) at (-4.05, -0.6) {\tiny{$\iter{\Delta g}{k}$}};
        \node (lds) at (-4.075, -1.0) {\tiny{$\iter{\Delta h}{k}$}};
        \node (sp_1) at (-3.975, -1.4) {\tiny{$\iter{s}{k}_{\neq}$}};
        \node (sp_2) at (-4, -1.8) {\tiny{$\iter{s}{k}_0$}};
        \node (reg) at (-3.875, -2.2) {\tiny{$\lambda$}};

        \node[linear,rotate=90, minimum width=2.75cm] (lin1) at (-2.5, 0) {\tiny{\texttt{Linear(12,30,bias=F)}}};
        \node[relu,rotate=90, minimum width=2.75cm] (relu2_1) at (-2.0, 0) {\tiny{\texttt{ReLU}}};

        \node[linear,rotate=90, minimum width=2.75cm] (lin2) at (-1.5, 0) {\tiny{\texttt{Linear(30,20,bias=F)}}};
        \node[relu,rotate=90, minimum width=2.75cm] (relu2_2) at (-1.0, 0) {\tiny{\texttt{ReLU}}};
        
        \node[linear,rotate=90, minimum width=2.75cm] (lin3) at (-0.5, 0) {\tiny{\texttt{Linear(20,10,bias=F)}}};
        \node[relu,rotate=90, minimum width=2.75cm] (relu2_3) at (0, 0) {\tiny{\texttt{ReLU}}};

        \node[linear,rotate=90,minimum width=2.75cm] (lin4) at (0.5, 0) {\tiny{\texttt{Linear(10,8,bias=F)}}};

        \draw[->] (gn_1) to [out=0, in=180] (lin1.north);
        \draw[->] (gn_2) to [out=0, in=180] (lin1.north);
        \draw[->] (mn_1) to [out=0, in=180] (lin1.north);
        \draw[->] (mn_2) to [out=0, in=180] (lin1.north);
        \draw[->] (sn_1) to [out=0, in=180] (lin1.north);
        \draw[->] (sn_2) to [out=0, in=180] (lin1.north);
        \draw[->] (ld) to [out=0, in=180] (lin1.north);
        \draw[->] (ldns) to [out=0, in=180] (lin1.north);
        \draw[->] (lds) to [out=0, in=180] (lin1.north);
        \draw[->] (sp_1) to [out=0, in=180] (lin1.north);
        \draw[->] (sp_2) to [out=0, in=180] (lin1.north);
        \draw[->] (reg) to [out=0, in=180] (lin1.north);
        
        \node (s1) at (1.8, 1.4) {\tiny{$w_1$}};
        \node (s2) at (1.8, 1.0) {\tiny{$w_2$}};
        \node (s3) at (1.8, 0.6) {\tiny{$w_3$}};
        \node (s4) at (1.8, 0.2) {\tiny{$w_4$}};
        \node (s5) at (1.8, -0.2) {\tiny{$w_5$}};
        \node (s6) at (1.8, -0.6) {\tiny{$w_6$}};
        \node (s7) at (1.8, -1.0) {\tiny{$w_7$}};
        \node (s8) at (1.8, -1.4) {\tiny{$w_8$}};

        \draw[->] (lin4.south) to [out=0, in=180] (s1.west);
        \draw[->] (lin4.south) to [out=0, in=180] (s2.west);
        \draw[->] (lin4.south) to [out=0, in=180] (s3.west);
        \draw[->] (lin4.south) to [out=0, in=180] (s4.west);
        \draw[->] (lin4.south) to [out=0, in=180] (s5.west);
        \draw[->] (lin4.south) to [out=0, in=180] (s6.west);
        \draw[->] (lin4.south) to [out=0, in=180] (s7.west);
        \draw[->] (lin4.south) to [out=0, in=180] (s8.west);

        \draw[-] (lin1) -- (relu2_1) -- (lin2) -- (relu2_2) -- (lin3) -- (relu2_3) -- (lin4);

        \node (d1) at (3.4, 1.4) {\tiny{$w_1 \cdot \iter{d}{k}_{1, \neq}$}};
        \node (d2) at (3.4, 1.0) {\tiny{$w_2 \cdot \iter{d}{k}_{1, 0}$}};
        \node (d3) at (3.4, 0.6) {\tiny{$w_3 \cdot \iter{d}{k}_{2, \neq}$}};
        \node (d4) at (3.4, 0.2) {\tiny{$w_4 \cdot \iter{d}{k}_{2, 0}$}};
        \node (d5) at (3.4, -0.2) {\tiny{$w_5 \cdot \iter{d}{k}_{3, \neq}$}};
        \node (d6) at (3.4, -0.6) {\tiny{$w_6 \cdot \iter{d}{k}_{3, 0}$}};
        \node (d7) at (3.4, -1.0) {\tiny{$w_7 \cdot \iter{d}{k}_{4, \neq}$}};
        \node (d8) at (3.4, -1.4) {\tiny{$w_8 \cdot \iter{d}{k}_{4, 0}$}};

        \draw[->] (s1) to (d1.west);
        \draw[->] (s2) to (d2.west);
        \draw[->] (s3) to (d3.west);
        \draw[->] (s4) to (d4.west);
        \draw[->] (s5) to (d5.west);
        \draw[->] (s6) to (d6.west);
        \draw[->] (s7) to (d7.west);
        \draw[->] (s8) to (d8.west);

        \node[conv,rotate=90,minimum width=2.7cm] (conv1) at (5.25,0) {\tiny{\texttt{Conv2d(8,20,1,bias=F)}}};
        \node[relu,rotate=90,minimum width=2.7cm] (relu1) at (5.75,0) {\tiny{\texttt{ReLU}}};
        
        \node[conv,rotate=90,minimum width=2.7cm] (conv2) at (6.25,0) {\tiny{\texttt{Conv2d(20,20,1,bias=F)}}};
        \node[relu,rotate=90,minimum width=2.7cm] (relu2) at (6.75,0) {\tiny{\texttt{ReLU}}};

        \node[conv,rotate=90,minimum width=2.7cm] (conv3) at (7.25,0) {\tiny{\texttt{Conv2d(20,20,1,bias=F)}}};
        \node[relu,rotate=90,minimum width=2.7cm] (relu3) at (7.75,0) {\tiny{\texttt{ReLU}}};

        \node[conv,rotate=90,minimum width=2.7cm] (conv4) at (8.25,0) {\tiny{\texttt{Conv2d(20,2,1,bias=F)}}};

        \node (d_1) at (9.3, 0.5) {\tiny{$\iter{d}{t}_{out, 1}$}};
        \node (d_2) at (9.3, -0.5) {\tiny{$\iter{d}{t}_{out, 2}$}};
        
        \draw[->] (d1) to [out=0, in=180] (conv1.north);
        \draw[->] (d2) to [out=0, in=180] (conv1.north);
        \draw[->] (d3) to [out=0, in=180] (conv1.north);
        \draw[->] (d4) to [out=0, in=180] (conv1.north);
        \draw[->] (d5) to [out=0, in=180] (conv1.north);
        \draw[->] (d6) to [out=0, in=180] (conv1.north);
        \draw[->] (d7) to [out=0, in=180] (conv1.north);
        \draw[->] (d8) to [out=0, in=180] (conv1.north);

        \draw[-] (conv1) -- (relu1) -- (conv2) -- (relu2) -- (conv3) -- (relu3) -- (conv4);
        \draw[->] (conv4.south) to [out=0, in=180] (d_1.west);
        \draw[->] (conv4.south) to [out=0, in=180] (d_2.west);

        \node (xk) at (10.6, 2.) {\tiny{$\iter{x}{t}$}};
        \node (new) at (10.6, 0) {\tiny{$\iter{x}{t+1}$}};
        \draw[->] (xk) to [out=270, in=90] (new.north);
        \draw[->] (d_1) to [out=0, in=180] (new.west);
        \draw[->] (d_2) to [out=0, in=180] (new.west);
        
    \end{tikzpicture}
    \caption{Algorithmic update for the LASSO problem: Based on the given features (split into zero and non-zero parts), the first block computes eight weights, which are used to perform a weighting of the different directions, which then get used in the second block. This second block predicts two directions $d_{out, 1}, d_{out, 2}$, where $d_{out, 1}$ only acts on the non-zero entries, and $d_{out, 2}$ acts on the zero entries. These are used in the update $\iter{x}{k+1} := \mathrm{prox}_{\beta g}\left(\iter{x}{k} + \left(\iter{d}{k}_{out, 1, \neq} - \nabla \ell(\iter{x}{k}, \realPar) + \norm{\iter{x}{k} - \iter{x}{k-1}}{} \cdot \iter{d}{k}_{out, 2, 0} \right) / L \right)$.}
\end{figure}

\subsubsection{Algorithm}
The solutions of the Lasso problem are typically sparse. To achieve this, the algorithm has to identify the coordinates which are non-zero. Therefore, in each iteration, we treat the zero and non-zero entries of $\iter{x}{k}$ (and derived quantities) separately. Here, non-zero entries are written with a $\neq$-subscript, while zero entries are written with a $0$-subscript, for example, $\iter{x}{k}_{\neq}$ and $\iter{x}{k}_0$.Then, first, we compute weights $w_1, ..., w_8$ with a fully-connected block with ReLU-activation functions, where we use the features $\iter{n}{k}_1 = \log(1 + \norm{\nabla \ell(\iter{x}{k}, \realPar)}{})$, $\iter{n}{k}_2 = \log(1 + \norm{\iter{x}{k} - \iter{x}{k-1}}{})$, $\iter{n}{k}_3 = \log(1 + \norm{\iter{p}{k}}{})$, where $\iter{p}{k} = \mathrm{prox}_{\beta g} \left( \iter{x}{k} - \beta \nabla \ell(\iter{x}{k}, \realPar) \right)$, $\Delta \iter{\ell}{k} := \ell(\iter{x}{k}, \realPar) - \ell(\iter{x}{k-1}, \realPar)$, $\Delta \iter{g}{k} := g(\iter{x}{k}) - g(\iter{x}{k-1})$, $\Delta \iter{h}{k} := h(\iter{x}{k}, \realPar) - h(\iter{x}{k-1}, \realPar)$, the scalar products $\iter{s}{k}_{\neq}$ and $\iter{s}{k}_0$ between the (normalized) gradient and (normalized) momentum, and the regularization parameter $\lambda$. Then, we use these weights to perform a reweighting of the directions $\iter{d}{k}_1, ..., \iter{d}{k}_4$, which are the normalized gradient, the normalized momentum, the normalized residual $\iter{x}{k} - \iter{p}{k}$, and the coordinate-wise product between (normalized) gradient and (normalized) momentum. Then, these reweighted directions get fed into a 1x1-convolutional block, which predicts the two directions $\iter{d}{k}_{out, 1}$ and $\iter{d}{k}_{out, 2}$, which we use to compute the final update with the proximal mapping, given by
$$
\iter{x}{k+1} := \mathrm{prox}_{\beta g}\left(\iter{x}{k} + \left(\iter{d}{k}_{out, 1, \neq} - \nabla \ell(\iter{x}{k}, \realPar) + \norm{\iter{x}{k} - \iter{x}{k-1}}{} \cdot \iter{d}{k}_{out, 2, 0} \right) / L \right) \,.
$$

\subsection{Details for the Neural-Network-Training Experiment}\label{App:Implementation_details_nn_training}
This subsection describes the missing details for the neural-network-training experiment.

\begin{figure}[t!]
    \centering
        \begin{tikzpicture}
        \tikzset{conv/.style={black,draw=black,fill=test_1,rectangle,minimum height=0.5cm}}
        \tikzset{linear/.style={black,draw=black,fill=test_2,rectangle,minimum height=0.5cm}}
        \tikzset{relu/.style={black,draw=black,fill=test_3,rectangle,minimum height=0.25cm}}

        \node (n1) at (-4, 1.8) {\tiny{$\iter{n}{k}_1$}};
        \node (n2) at (-4, 0.9) {\tiny{$\iter{n}{k}_2$}};
        \node (l1) at (-4, 0) {\tiny{$\iter{\ell}{k}$}};
        \node (l2) at (-4, -0.9) {\tiny{$\iter{\ell}{k-1}$}};
        \node (s) at (-4, -1.8) {\tiny{$\iter{s}{k}$}};

        \node[linear,rotate=90,minimum width=2.3cm] (lin1) at (-3, 0) {\tiny{\texttt{Linear(5,8,bias=F)}}};
        \node[relu,rotate=90,minimum width=2.3cm] (relu2_1) at (-2.5, 0) {\tiny{\texttt{ReLU}}};

        \node[linear,rotate=90,minimum width=2.3cm] (lin2) at (-2.0, 0) {\tiny{\texttt{Linear(8,8,bias=F)}}};
        \node[relu,rotate=90,minimum width=2.3cm] (relu2_2) at (-1.5, 0) {\tiny{\texttt{ReLU}}};

        \node[linear,rotate=90,minimum width=2.3cm] (lin3) at (-1.0, 0) {\tiny{\texttt{Linear(8,4,bias=F)}}};

        \draw[->] (n1) to [out=0, in=180] (lin1.north);
        \draw[->] (n2) to [out=270, in=180] (lin1.north);
        \draw[->] (l1) to [out=0, in=180] (lin1.north);
        \draw[->] (l2) to [out=90, in=180] (lin1.north);
        \draw[->] (s) to [out=0, in=180] (lin1.north);

        \node (s1) at (0.1, 1.25) {\tiny{$\notation{w}_1$}};
        \node (s2) at (0.1, 0.325) {\tiny{$\notation{w}_2$}};
        \node (s3) at (0.1, -0.325) {\tiny{$\notation{w}_3$}};
        \node (s4) at (0.1, -1.25) {\tiny{$\notation{w}_4$}};

        \draw[->] (lin3.south) to [out=0, in=180] (s1.west);
        \draw[->] (lin3.south) to [out=0, in=180] (s2.west);
        \draw[->] (lin3.south) to [out=0, in=180] (s3.west);
        \draw[->] (lin3.south) to [out=0, in=180] (s4.west);

        \draw[-] (lin1) -- (relu2_1) -- (lin2) -- (relu2_2) -- (lin3);

        \draw (1, 2.) circle [radius=0.18] node (g) {\tiny{$g$}};
        \draw (1, -2.) circle [radius=0.18] node (m) {\tiny{$m$}};
        \node (d1) at (1.5, 1.25) {\tiny{$\notation{w}_1 \cdot g \odot \iter{d}{k}_1$}};
        \node (d2) at (1.8, 0.325) {\tiny{$\notation{w}_2 \cdot \iter{d}{k}_1$}};
        \node (d3) at (1.8, -0.325) {\tiny{$\notation{w}_3 \cdot \iter{d}{k}_2$}};
        \node (d4) at (1.5, -1.25) {\tiny{$\notation{w}_4 \cdot m \odot \iter{d}{k}_2$}};
        
        \node (xk) at (2.5, 2.) {\tiny{$\iter{x}{k}$}};
        \node (xkm1) at (2.5, -2.) {\tiny{$\iter{x}{k-1}$}};

        \draw[->] (g) to [out=0,in=90] (d1.north);
        \draw[->] (s1) to (d1.west);
        \draw[->] (s2) to (d2.west);
        \draw[->] (s3) to (d3.west);
        \draw[->] (s4) to (d4.west);
        \draw[->] (m) to [out=0, in=270] (d4.south);

        \node[conv,rotate=90,minimum width=2.7cm] (conv1) at (3.25,0) {\tiny{\texttt{Conv2d(6,16,1,bias=F)}}};
        \node[relu,rotate=90,minimum width=2.7cm] (relu1) at (3.75,0) {\tiny{\texttt{ReLU}}};
        
        \node[conv,rotate=90,minimum width=2.7cm] (conv2) at (4.25,0) {\tiny{\texttt{Conv2d(16,16,1,bias=F)}}};
        \node[relu,rotate=90,minimum width=2.7cm] (relu2) at (4.75,0) {\tiny{\texttt{ReLU}}};

        \node[conv,rotate=90,minimum width=2.7cm] (conv3) at (5.25,0) {\tiny{\texttt{Conv2d(16,16,1,bias=F)}}};
        \node[relu,rotate=90,minimum width=2.7cm] (relu3) at (5.75,0) {\tiny{\texttt{ReLU}}};

        \node[conv,rotate=90,minimum width=2.7cm] (conv4) at (6.25,0) {\tiny{\texttt{Conv2d(16,16,1,bias=F)}}};
        \node[relu,rotate=90,minimum width=2.7cm] (relu4) at (6.75,0) {\tiny{\texttt{ReLU}}};

        \node[conv,rotate=90,minimum width=2.7cm] (conv5) at (7.25,0) {\tiny{\texttt{Conv2d(16,16,1,bias=F)}}};
        \node[relu,rotate=90,minimum width=2.7cm] (relu5) at (7.75,0) {\tiny{\texttt{ReLU}}};

        \node[conv,rotate=90,minimum width=2.7cm] (conv6) at (8.25,0) {\tiny{\texttt{Conv2d(16,1,1,bias=F)}}};

        \node (d) at (9.2, 0) {\tiny{$\iter{d}{k}$}};

        \draw[->] (xk) to [out=0,in=90] (conv1.east);
        \draw[->] (xkm1) to [out=0,in=270] (conv1.west);
        
        \draw[->] (d1) to [out=0, in=180] (conv1.north);
        \draw[->] (d2) to [out=0, in=180] (conv1.north);
        \draw[->] (d3) to [out=0, in=180] (conv1.north);
        \draw[->] (d4) to [out=0, in=180] (conv1.north);

        \draw[-] (conv1) -- (relu1) -- (conv2) -- (relu2) -- (conv3) -- (relu3) -- (conv4) -- (relu4) -- (conv5) -- (relu5) -- (conv6);
        \draw[->] (conv6) -- (d);

        \node (new) at (10.4, 0) {\tiny{$\iter{x}{k+1}$}};
        \draw[->] (xk) to [out=0, in=90] (new.north);
        \draw[->] (d) -- (new);
        
    \end{tikzpicture}
    \caption{Algorithmic update for training the neural network: Based on the two norms $\iter{n}{k}_1$ and $\iter{n}{k}_2$, the scalar product $\iter{s}{k} := \sprod{\iter{d}{k}_1}{\iter{d}{k}_2}$, and the current and previous (logarithmically scaled) loss $\iter{\ell}{k}$, $\iter{\ell}{k-1}$, we compute four scalars $\notation{w}_1, ..., \notation{w}_4$, which are used for weighting $\iter{d}{k}_1$, $\iter{d}{k}_2$, and their corresponding pre-conditioned versions $g \odot \iter{d}{k}_1$ and $m \odot \iter{d}{k}_2$. Together with the current and previous point, they get fed (as separate channels) into the first layer of a 1x1-convolutional block, which computes an update direction $\iter{d}{k}$. Then, we update $\iter{x}{k+1} := \iter{x}{k} + \iter{d}{k}$.}
    \label{architecture:neural_network}
\end{figure}

\subsubsection{Construction of the Parameters}
We assume that the neural network should learn a function $g: \R \to \R$ from noisy observations $y_j$, that is $y_j = g(x_j) + \varepsilon$ with $\varepsilon \sim \mathcal{N}(0,1)$.
For this, we construct polynomials $g_i$, $i=1,...,N$, of degree $d=5$ as follows: For every function $g_i$, we sample points $\{x_{i,j}\}_{j=1}^K$ (here: $K = 50$) uniformly in $[-2, 2]$, that is, $x_{i,j} \overset{iid}{\sim} \mathrm{U}[-2, 2]$, $i = 1 ,..., N$, $j = 1, ..., K$. Then, we sample the coefficients $(c_{i, 0}, ..., c_{i, 5})$ of $g_i$ uniformly in $[-5, 5]$, that is, $c_{i, l} \overset{iid}{\sim} \mathrm{U}[-5, 5]$, $i = 1 ,..., N$, $l = 0, ..., 5$. Lastly, we get the values $y_{i,j}$ as:
$$
    y_{i,j} = g_i(x_{i,j}) + \varepsilon_{i, j} \quad \text{with} \quad \varepsilon_{i, j} \overset{iid}{\sim} \mathcal{N}(0,1), \quad i = 1 ,..., N, \ j = 1, ..., 50\,.
$$
For every function $g_i: \R \to \R$ the neural network is trained on the data set $\notation{\realPar_i} := \{X_i, Y_i\}$ with $X_i = (x_{i,1}, ..., x_{i,K}) \in \R^K$ and $Y_i = (y_{i,1}, ..., y_{i,K}) \in \R^K$. Hence, the data set will serve as the parameter $\notation{\realPar}$ of the loss function $\ell: \R^p \times \notation{\parSpace} \to \R_{\ge 0}$, such that the parameter space $\notation{\parSpace}$ can be identified as the space of these data sets, that is, $\notation{\parSpace} = \R^{K \times 2}$. 

\subsubsection{Loss Function and Architecture}
Since the mean square error is the standard choice for training models on regression tasks, the loss is given by $\ell(\beta, \notation{\realPar_i}) := c(\mathtt{N}(\beta, X_i), Y_i) := \frac{1}{K} \sum_{j=1}^{K} (\mathtt{N}_\beta(x_{i,j}) - y_{i,j})^2$. \\
As model we use a fully-connected two layer neural network with \texttt{ReLU}-activation functions. To have more features in the input layer, the input $x$ is transformed into the vector $(x, x^2, ..., x^5)$. Hence, the parameters $\beta \in \R^{\notation{m}}$ are given by the weights $A_1 \in \R^{50 \times 5}, A_2 \in \R^{1 \times 50}$ and biases $b_1 \in \R^{50}, b_2 \in \R$ of the two fully-connected layers. Therefore, the optimization space is of dimension $\notation{m} = (5 \cdot 50) + (1 \cdot 50) + 50 + 1 = 351$.

\subsubsection{Algorithm} The algorithmic update in Figure~\ref{architecture:neural_network} consists of two blocks: A weighting block, which computes four weights $\notation{w}_1, ..., \notation{w}_4$ based on the norms $\iter{n}{k}_1, \iter{n}{k}_2$, the losses $\ell(\iter{x}{k}, \notation{\realPar}), \ell(\iter{x}{k-1}, \notation{\realPar})$, and the scalar product $\sprod{\iter{d}{k}_1}{\iter{d}{k}_2}$. Each of these gets multiplied with $\iter{d}{k}_1, \iter{d}{k}_2$, or the \say{pre-conditioned} versions, which we compute by pointwise multiplication with the learned vectors $g$ and $m$. Then, additionally to the $\iter{x}{k}$ and $\iter{x}{k-1}$, these weighted directions get fed into an update-block, which computes the final update direction $\iter{d}{k}$.

\section{Additional Experiment on MNIST}

\begin{figure}[t!]
    \centering
        \begin{tikzpicture}
        \tikzset{conv/.style={black,draw=black,fill=test_1,rectangle,minimum height=0.5cm}}
        \tikzset{linear/.style={black,draw=black,fill=test_2,rectangle,minimum height=0.5cm}}
        \tikzset{relu/.style={black,draw=black,fill=test_3,rectangle,minimum height=0.25cm}}

        \node (n1) at (-4, 1.5) {\tiny{$\iter{n}{k}_1$}};
        \node (n2) at (-4, 0.5) {\tiny{$\iter{n}{k}_2$}};
        \node (l1) at (-4, -0.5) {\tiny{$\iter{\ell}{k}$}};
        \node (l2) at (-4.3, -1.5) {\tiny{$\iter{\ell}{k-1}$}};

        \node[linear,rotate=90,minimum width=2.8cm] (lin1) at (-3, 0) {\tiny{\texttt{Linear(4,15,bias=F)}}};
        \node[relu,rotate=90,minimum width=2.8cm] (relu2_1) at (-2.5, 0) {\tiny{\texttt{ReLU}}};

        \node[linear,rotate=90,minimum width=2.8cm] (lin2) at (-2.0, 0) {\tiny{\texttt{Linear(15,20,bias=F)}}};
        \node[relu,rotate=90,minimum width=2.8cm] (relu2_2) at (-1.5, 0) {\tiny{\texttt{ReLU}}};

        \node[linear,rotate=90,minimum width=2.8cm] (lin3) at (-1.0, 0) {\tiny{\texttt{Linear(20,15,bias=F)}}};
        \node[relu,rotate=90,minimum width=2.8cm] (relu2_3) at (-0.5, 0) {\tiny{\texttt{ReLU}}};

        \node[linear,rotate=90,minimum width=2.8cm] (lin4) at (0, 0) {\tiny{\texttt{Linear(15,7,bias=F)}}};
        
        \draw[->] (n1) to [out=0, in=180] (lin1.north);
        \draw[->] (n2) to [out=270, in=180] (lin1.north);
        \draw[->] (l1) to [out=90, in=180] (lin1.north);
        \draw[->] (l2) to [out=0, in=180] (lin1.north);

        \node (s1) at (1.05, 1.5) {\tiny{$\notation{w}_1$}};
        \node (s2) at (1.05, 1.0) {\tiny{$\notation{w}_2$}};
        \node (s3) at (1.05, 0.5) {\tiny{$\notation{w}_3$}};
        \node (s4) at (1.05, 0.0) {\tiny{$\notation{w}_4$}};
        \node (s5) at (1.05, -0.5) {\tiny{$\notation{w}_5$}};
        \node (s6) at (1.05, -1.0) {\tiny{$\notation{w}_6$}};
        \node (s7) at (1.05, -1.5) {\tiny{$\notation{w}_7$}};

        \draw[->] (lin4.south) to [out=0, in=180] (s1.west);
        \draw[->] (lin4.south) to [out=0, in=180] (s2.west);
        \draw[->] (lin4.south) to [out=0, in=180] (s3.west);
        \draw[->] (lin4.south) to [out=0, in=180] (s4.west);
        \draw[->] (lin4.south) to [out=0, in=180] (s5.west);
        \draw[->] (lin4.south) to [out=0, in=180] (s6.west);
        \draw[->] (lin4.south) to [out=0, in=180] (s7.west);

        \draw[-] (lin1) -- (relu2_1) -- (lin2) -- (relu2_2) -- (lin3) -- (relu2_3) -- (lin4);

        \node (d1) at (2.6, 1.5) {\tiny{$\notation{w}_1 \cdot g_1 \odot \iter{d}{k}_1$}};
        \node (d2) at (2.6, 1.0) {\tiny{$\notation{w}_2 \cdot g_2 \odot \iter{d}{k}_1$}};
        \node (d3) at (2.9, 0.5) {\tiny{$\notation{w}_3 \cdot \iter{d}{k}_1$}};
        \node (d4) at (2.6, 0.0) {\tiny{$\notation{w}_4 \cdot \iter{d}{k}_1 \odot \iter{d}{k}_2$}};
        \node (d5) at (2.9, -0.5) {\tiny{$\notation{w}_5 \cdot \iter{d}{k}_2$}};
        \node (d6) at (2.6, -1.0) {\tiny{$\notation{w}_6 \cdot m_1 \odot \iter{d}{k}_2$}};
        \node (d7) at (2.6, -1.5) {\tiny{$\notation{w}_7 \cdot m_2 \odot \iter{d}{k}_2$}};

        \draw[->] (s1) to (d1.west);
        \draw[->] (s2) to (d2.west);
        \draw[->] (s3) to (d3.west);
        \draw[->] (s4) to (d4.west);
        \draw[->] (s5) to (d5.west);
        \draw[->] (s6) to (d6.west);
        \draw[->] (s7) to (d7.west);

        \node[conv,rotate=90,minimum width=2.8cm] (conv1) at (4.5,0) {\tiny{\texttt{Conv2d(7,15,1,bias=F)}}};
        \node[relu,rotate=90,minimum width=2.8cm] (relu1) at (5.,0) {\tiny{\texttt{ReLU}}};
        
        \node[conv,rotate=90,minimum width=2.8cm] (conv2) at (5.5,0) {\tiny{\texttt{Conv2d(15,20,1,bias=F)}}};
        \node[relu,rotate=90,minimum width=2.8cm] (relu2) at (6.,0) {\tiny{\texttt{ReLU}}};

        \node[conv,rotate=90,minimum width=2.8cm] (conv3) at (6.5,0) {\tiny{\texttt{Conv2d(20,15,1,bias=F)}}};
        \node[relu,rotate=90,minimum width=2.8cm] (relu3) at (7.,0) {\tiny{\texttt{ReLU}}};

        \node[conv,rotate=90,minimum width=2.8cm] (conv4) at (7.5,0) {\tiny{\texttt{Conv2d(15,1,1,bias=F)}}};

        \node (d) at (8.75, 0) {\tiny{$\iter{d}{k}$}};
        
        \draw[->] (d1) to [out=0, in=180] (conv1.north);
        \draw[->] (d2) to [out=0, in=180] (conv1.north);
        \draw[->] (d3) to [out=0, in=180] (conv1.north);
        \draw[->] (d4) to [out=0, in=180] (conv1.north);
        \draw[->] (d5) to [out=0, in=180] (conv1.north);
        \draw[->] (d6) to [out=0, in=180] (conv1.north);
        \draw[->] (d7) to [out=0, in=180] (conv1.north);

        \draw[-] (conv1) -- (relu1) -- (conv2) -- (relu2) -- (conv3) -- (relu3) -- (conv4);
        \draw[->] (conv4) -- (d);

        \node (xk) at (10, 1.) {\tiny{$\iter{x}{k}$}};
        \node (new) at (10, 0) {\tiny{$\iter{x}{k+1}$}};
        \draw[->] (xk) to [out=270, in=90] (new.north);
        \draw[->] (d) -- (new);
        
    \end{tikzpicture}
    \caption{Algorithmic update for the MNIST experiment: Based on the two norms $\iter{n}{k}_1$ and $\iter{n}{k}_2$, and the current and previous (logarithmically scaled) loss $\iter{\ell}{k}$, $\iter{\ell}{k-1}$, we compute seven scalars $\notation{w}_1, ..., \notation{w}_7$, which are used for weighting $\iter{d}{k}_1$, $\iter{d}{k}_2$, their corresponding preconditioned versions $g_i \odot \iter{d}{k}_1$, $m_i \odot \iter{d}{k}_2$, and their coordinate-wise product $\iter{d}{k}_1 \odot \iter{d}{k}_2$. Then, they get fed (as separate channels) into the first layer of a 1x1-convolutional block, which computes an update direction $\iter{d}{k}$. Finally, we update $\iter{x}{k+1} := \iter{x}{k} + \iter{d}{k}$.}
    \label{architecture:mnist}
\end{figure}
\begin{figure}[t!]
    \centering
    \includegraphics[width=\textwidth]{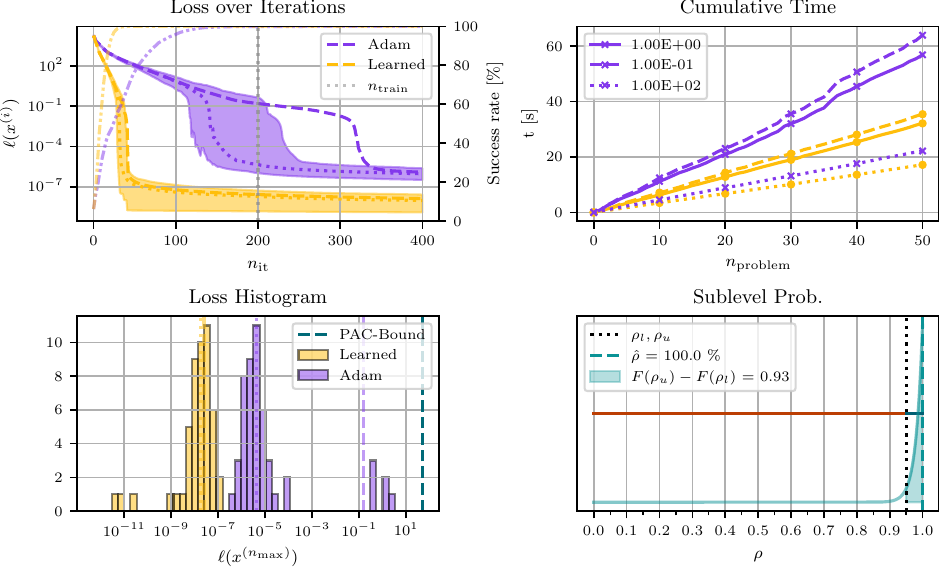}
    \caption{\textbf{Upper left:} Dashed lines represent the mean losses, dotted lines represent the median losses, and the shaded regions indicate the 10th to 90th percentile. Further, dashdotted-lines represent the classification-accuracy, which is shown on the right y-axis. Here, Adam is shown in purple and the learned algorithm in yellow. 
    \textbf{Upper right:} The different lines show the cumulative computation time of the algorithms to solve all the test problems up to a certain accuracy (in function values) measured by $\ell(\beta^{(i)}, \notation{\realPar}) - c(X_i, Y_i) < \varepsilon$. However, note that both algorithms are run for maximally $n_{\rm{max}} = 5000$ iterations. 
    \textbf{Lower left:} Loss histogram (after $n_{\rm{train}}$ iterations) and PAC-bound. 
    \textbf{Lower right:} The teal dashed line shows the point estimate for the sublevel probability, while the teal solid line shows the $\mathrm{Beta}$-posterior. Here, the black dotted lines indicate the constraints $\sublevelprob_l, \sublevelprob_u$ and show the feasible region as dark teal line.}
    \label{fig:mnist_summary}
\end{figure}

This experiment considers the problem of training a neural network to do classification on the MNIST data set, that is, $\mathcal{A}$ is trained to predict the parameters $\beta \in \R^{\notation{m}}$ of a neural network $\mathtt{N}_\beta$, which then is used to predict a class-label $k \in \{0, ..., 9\}$ based on an input image. Hence, the optimization variable is given by $\beta \in \R^{\notation{m}}$. Here, the model consists of two convolutional layers with ReLU-activation functions and Max-Pooling, followed by three linear layers with ReLU-activation functions. Through this, the optimization variable $\beta$ has dimension $m = 13090$. 
\begin{Rem}
    Note that, as the theory does not apply to stochastic algorithms, we have to compute full gradients. This limits the amount of images (20x20) per data set to 250.
\end{Rem}
As loss-function, we use a penalized cross-entropy loss to enforce higher classification-accuracy, as parameters $\realPar$ of the loss-function we use the data sets consisting of input images and ground-truth labels, that is, $p \in \R^{250 \times (20 \times 20) \times 1} = \parSpace$, and as baseline we use Adam. The architecture of the learnt algorithm is shown in Figure~\ref{architecture:mnist} and consists of two blocks: The first block uses linear layers with ReLU-activation functions and computes seven weights based on the gradient-norm, the norm of the momentum-term, and the incurred losses. Then, these weights are used to weigh the input-directions of the second block. It consists of $1\times 1$-convolutional layers with ReLU-activation functions, and computes an update-direction based on the gradient, the momentum, their coordinate-wise product, and four additional directions, which are computed from the gradient and the momentum-term by coordinate-wise preconditioning. For more details we refer to the GitHub repository. Figure~\ref{fig:mnist_summary} shows the results of this experiment: The upper left plot shows that the learned algorithm outperforms Adam, classifying all images correctly after about 50 iterations, while Adam needs about 200 iterations to reach the same classification-accuracy. The upper right plot confirms that, also in terms of computation time, $\mathcal{A}$ is faster in training the neural network than Adam. However, based on the higher computational cost per iteration, the gap is not as large as for the function values. The lower left plot shows that the predicted PAC-bound is not tight here. This can be attributed to the fact that we had to use a smaller amount of data, due to the high computational cost in each iteration. Finally, the lower right plot indicates that the algorithm did reach the sublevel set in all test cases.

\newpage
\bibliography{bibliography.bib}

\end{document}